\documentclass{article}
\usepackage{geometry}
\usepackage{tikz}
\usepackage{mathrsfs} 
\newcommand{\mathbcal}[1]{{\mathscr{#1}}}
\usetikzlibrary{calc}
\usepackage[cmintegrals,cmbraces]{newtxmath}
\usepackage[T1]{fontenc}
\usepackage{mathtools}
\usepackage{subcaption}
\usepackage{algorithm}
\usepackage[noend]{algpseudocode}

\usetikzlibrary{decorations.pathreplacing,calligraphy, decorations.markings}
\usetikzlibrary{arrows}
\usetikzlibrary{arrows.meta}
\usepackage{amsmath}
\usetikzlibrary{decorations.pathmorphing}
\tikzset{snake it/.style={decorate, decoration=snake}}
\usepackage{empheq}
\usepackage{xcolor}
\usepackage[most]{tcolorbox}
\usepackage[makeroom]{cancel}
\usepackage{subcaption}
\usepackage{graphicx}
\graphicspath{{images/}}
\usepackage{blindtext}
\usepackage{authblk}

\usepackage[
pdfstartview=XYZ,
bookmarks=true,
colorlinks=true,
linkcolor=blue,
urlcolor=blue,
citecolor=blue,
pdftex,
bookmarks=true,
linktocpage=true,   
hyperindex=true
]{hyperref}

\graphicspath{{./figures/}}

\usepackage{natbib}
\setcitestyle{authoryear}

\usepackage{xcolor}

\newcommand{\aref}[1]{App.\,~\ref{#1}}
\newcommand{\fref}[1]{Fig.\,~\ref{#1}}
\newcommand{\tref}[1]{Table\,~\ref{#1}}
\newcommand{\eref}[1]{Eq.\,~(\ref{#1})}

\newcommand{\sref}[1]{Sec.\!~\ref{#1}}

\newcommand{\cref}[1]{Ref.\,~\cite{#1}}

\newcommand{\pdf}{{\pi}}
\newcommand{\policy}{\boldsymbol{p}}
\newcommand{\action}{\boldsymbol{a}}
\newcommand{\state} {\boldsymbol{s}}

\newcommand{\ds}{\mathsf{d}}
\newcommand{\rs}{\mathsf{r}}

\newcommand{\ys}{\mathsf{y}}

\newcommand{\As}{\mathsf{A}}

\newcommand{\Fs}{\mathsf{F}}
\newcommand{\Qs}{\mathsf{Q}}

\newcommand{\Rs}{\mathsf{R}}

\newcommand{\Ms}{\mathsf{M}}

\newcommand{\Ks}{\mathsf{K}}

\newcommand{\Ss}{\mathsf{S}}

\newcommand{\fb}{\mathbf{f}}
\newcommand{\gb}{\mathbf{g}}

\newcommand{\mb}{\mathbf{m}}
\newcommand{\xb}{\mathbf{x}}

\newcommand{\yb}{\mathbf{y}}
\newcommand{\zb}{\mathbf{z}}

\newcommand{\Ib}{\mathbf{I}}

\newcommand{\Cbb}{\mathbb{C}}

\newcommand{\mub}{\boldsymbol{\mu}}

\newcommand{\epsilonb}{{\boldsymbol{\epsilon}}}
\newcommand{\varepsilonb}{\boldsymbol{\varepsilon}}
\newcommand{\etab}{{\boldsymbol{\eta}}}
\newcommand{\deltab}{{\boldsymbol{\delta}}}
\newcommand{\sigmab}{{\boldsymbol{\sigma}}}
\newcommand{\Sigmab}{\boldsymbol{\Sigma}}
\newcommand{\thetab}{{\boldsymbol{\theta}}}

\newcommand{\partialb}{{\boldsymbol{\partial}}}

\newcommand{\tr}{\operatorname{tr}}

\usepackage{amsthm}
\theoremstyle{remark}
\newtheorem*{remark}{Remark}

\title{\bf Design of experiments for the calibration of history-dependent models via deep reinforcement learning and an enhanced Kalman filter}

\author[1]{Ruben Villarreal}
\affil[1]{Sandia National Laboratories, Livermore, California}
\author[2]{Nikolaos N. Vlassis\footnote{Corresponding author: Nikolaos N. Vlassis, email address: nnv2102@columbia.edu}}
\author[2]{Nhon N. Phan}
\affil[2]{Department of Civil Engineering and Engineering Mechanics, Columbia University, New York}
\author[1]{Tommie A. Catanach}
\author[1]{Reese E. Jones}
\author[3]{Nathaniel A. Trask}
\affil[3]{Sandia National Laboratories, Albuquerque, New Mexico}
\author[3]{Sharlotte L.B. Kramer}
\author[2]{WaiChing Sun}
\date{}
\begin{document}
\maketitle

\begin{abstract}
Experimental data is costly to obtain, which makes it difficult to calibrate complex models. 
For many models an experimental design that produces the best calibration given a limited experimental budget is not obvious.
This paper introduces a deep reinforcement learning (RL) algorithm for design of experiments that maximizes the information gain measured by Kullback-Leibler (KL) divergence obtained via the Kalman filter (KF). 
This combination enables experimental design for rapid online experiments where traditional methods are too costly.
We formulate possible configurations of experiments as a decision tree and a Markov decision process (MDP), where a finite choice of actions is available at each incremental step. 
Once an action is taken, a variety of measurements are used to update the state of the experiment. 
This new data leads to a Bayesian update of the parameters by the KF, which is used to enhance the state representation. 
In contrast to the Nash-Sutcliffe efficiency (NSE) index, which requires additional sampling to test hypotheses for forward predictions, the KF can lower the cost of experiments by directly estimating the values of new data acquired through additional actions.
In this work our applications focus on mechanical testing of materials.
Numerical experiments with complex, history-dependent models are used to verify the implementation and benchmark the performance of the RL-designed experiments. 
\end{abstract}

\section{Introduction}

Finding an accurate representation for the physical response of a material with complex nonlinear behavior is a difficult endeavor that depends on the parametric complexity of the selected model, the experimental data available for calibration, and other factors such as indirect, noisy, or incomplete observations.
This has motivated experimental design as a longstanding research area \citep{fisher1937design} with a multitude of approaches.

The physical characteristics of the material of interest in large part guide optimal experimental design.
Material symmetry plays a role in model and response complexity, and hence in the complexity of the experiments used to characterize it.
For instance, the irreducible number of parameters for isotropic linear elasticity is just two, so two linearly independent observations of stress are sufficient to characterize the model; however, the number of parameters increases for lower symmetry materials such as transversely isotropic, orthotropic, monoclinic and fully anisotropic elasticity (which has 21 independent parameters).
This complexity requires more independent observations and poses a more challenging optimal experimental design problem.
Nonlinearity and path/history dependence of the material response also play a role in model complexity and the data needed to obtain a sufficiently accurate calibration. 
Given large enough loads, most materials exhibit both nonlinearity in the stress-strain response and dissipation which leads to path dependence.
For instance, the material model in \citet{ames2009thermo} requires 37 material parameters and complex forms to capture the thermo-mechanical response of amorphous polymers, and \citet{ma2020computational} requires 25 material parameters to capture the crystal plasticity and phase transition of salt under high pressure and high temperature.
While the use of machine learning to generate constitutive laws \citep{jones2022neural, vlassis2022component} may enable one to bypass the need to identify a particular model form, the data required to obtain a sufficiently accurate neural network or Gaussian process model may be considerable and complex/ambiguous, and hence costly to obtain \citep{fuchs2021dnn2, heider2021offline, wang2019cooperative, wang2021non}. 

\begin{figure}[ht!]
\centering
\includegraphics[width=0.59\textwidth]{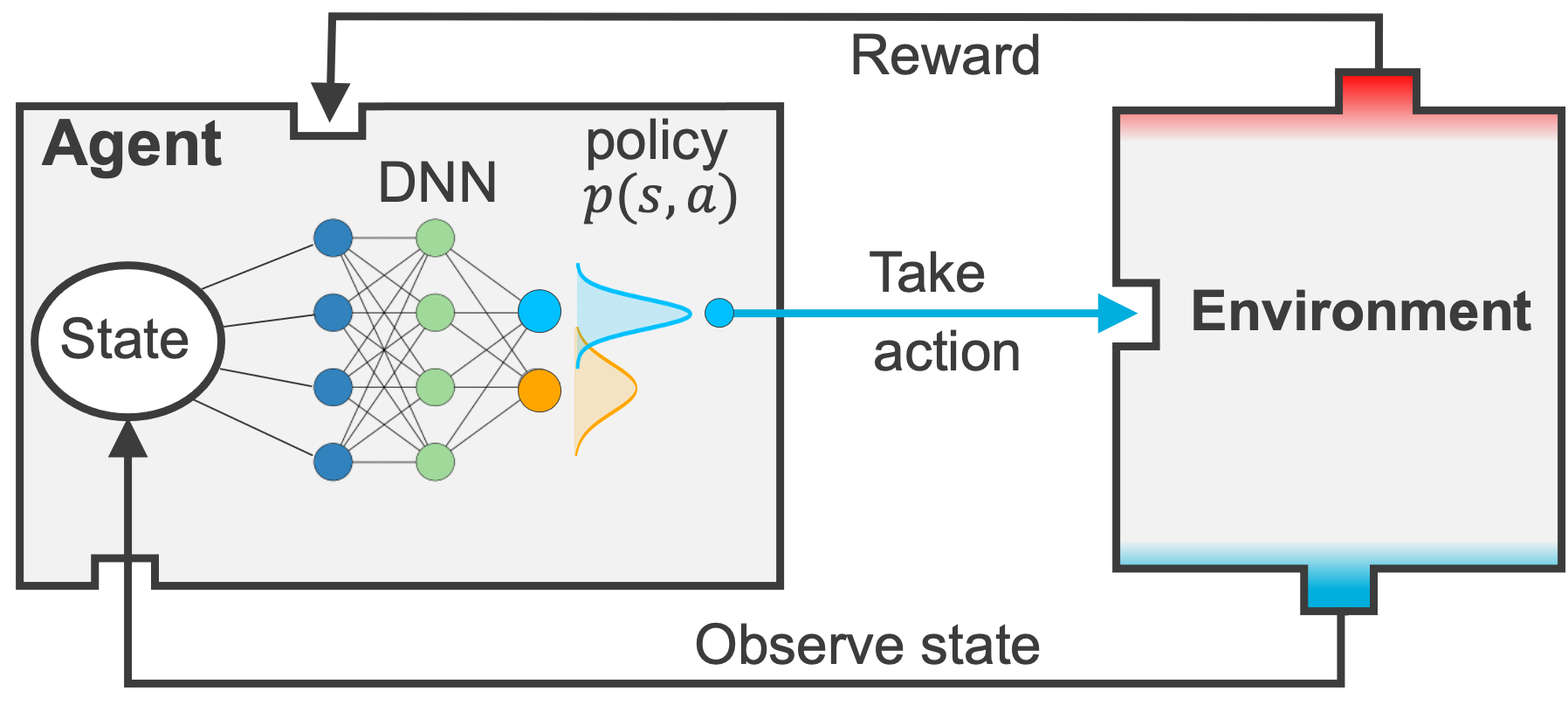}
\caption{Deep reinforcement learning.
The {\it environment} consists of a physical experiment, which reacts to {\it actions}, such as a prescribed strain, to produce a new observed {\it state}.
The reaction of the experiment, by way of the calibration method, produces a {\it reward} that drives the (external) {\it agent} to generate a {\it policy} that takes actions that maximizes rewards. 
In deep RL the action-value policy is represented with a deep neural network (DNN) that takes states and current rewards as inputs and produces favorable actions and values (which are  accumulated rewards) as outputs.
}
\label{fig:drl}
\end{figure}

This complexity of both traditional and emerging models, combined with the epistemic uncertainty that commonly occurs with experimental calibration, makes it difficult to use intuition alone to foresee the optimal experimental design for a fixed amount of resources (such as duration of the experiment).
While Bayesian design of experiments (DOE) \citep{chaloner1995bayesian,pukelsheim2006optimal} may efficiently estimate and optimize the information gain of an experiment with a limited number of design variables, calibrating a material model often requires hundreds or even thousands of loading steps, each with a set of available options/control actions. 
This decision-tree, which starts with the material in its reference state and expands from there, amounts to a combinatorial state space whose growth with loading steps precludes application of traditional Bayesian DOE.
The sequential process of decision-action-feedback of an experiment can be represented as a Markov decision process (MDP) in discrete time, where the optimal design of the experiment can be recast as a policy that optimizes a pre-selected set of rewards.
This is the foundation of reinforcement learning which is shown schematically in \fref{fig:drl} and centers around an {\it actor}  utilizing a {\it policy} to maximize {\it rewards}.
Please refer to \cite{graesser2019foundations} for an introduction to reinforcement learning.

In this work, we introduce a reinforcement learning (RL) \citep{kaelbling1996reinforcement, li2017deep, sutton2018reinforcement} approach to optimal experimental design that utilizes a deep neural network and an enhanced Kalman filter (KF) \citep{lee1994extended, williams1992training} for policy estimation to maximize the information gain for an experiment.  
We develop a framework which exploits our prior knowledge of the underlying physics by utilizing the structure of common history-dependent models, such as plasticity, and augments the RL state with the uncertainty associated with model parameters. 
This allows the agent to select actions which are guided by parameter sensitivities to reduce the estimated variance of the parameters.
The enhanced Kalman filter we present will provide a computationally efficient means of estimating parameter uncertainty for nonlinear systems.
This approach offers several salient features and improvements over previous efforts (cf. \citet{wang2019meta}) and is focused on providing optimal experimental designs for complex models.
First of all, the use of a KF to provide a parameter calibration and parametric uncertainty-based reward via the Kullback–Leibler (KL) divergence measuring information gain enables the method to bypass the costly bottleneck induced by the Nash-Sutcliffe efficiency (NSE) index \citep{mccuen2006evaluation} used to estimate the values of new experimental data. 
This bypass leads to significant cost savings for experimental data that are expensive to obtain. 
Second, the introduction of a deep neural network trained by Monte Carlo tree search (MCTS) balances the needs for exploration and exploitation. 
Third, the RL and KF are enhanced with additional state information and strategies to handle the history dependence of the process and model.
In essence, we seek to replace a traditional experimentalist who pre-conceives experiments with an RL actor who is guided by a pre-trained policy and reacts to real-time rewards to minimize uncertainty in the model calibration.

In \sref{sec:calibration} we develop a Kalman filter that can handle the non-linearity, non-smoothness, and history dependence of common material models.
It also exploits known behavior of the model, for example specific parameter sensitivities dominate in certain regimes.
Since the Kalman filter is essentially an incremental Bayesian calibration (and state estimation) method, it provides parameter covariances as well as estimates of the mean parameters.
The deep reinforcement learning method, described \sref{sec:RL}, utilizes this information in both the state and the reward that defines and guides the policy.
The policy for controlling an experiment with the chosen reward creates data along a path that maximizes information gain in the selected model's parameters.
We employ a policy-value scheme that represents the rewards for control actions over the decision tree with a neural network.
Since the possible paths of even simple experiments with a few allowable actions/decisions at every state have an enormous number of possible paths, we need to use Monte Carlo sampling on this tree to train the policy-value network.
In \sref{sec:results}, the proposed algorithm is demonstrated with widely-used plasticity models \citep{lubliner2008plasticity}.
Two numerical examples are specifically designed to validate the deep RL where a benchmark of optimal experimental design is available.  
A third example demonstrates that the algorithm is effective in obtaining an optimal experimental design given a pre-selected model where the best design is not known {\it a priori}. 
\tref{tab:notation} summarizes the notation used in the following sections.

\begin{table}[ht]
\centering
\begin{tabular}{|c|l|}
\hline
Symbol & Description \\
\hline
$\ds$ & observable output (data) \\
$\yb$ & model output \\
$\xb$ & controllable input \\
$\zb$ & hidden model state \\
$\mb$ & observation/response model \\
$\fb$ & hidden state dynamics\\
$\thetab$ & parameters \\
\hline
$\sigmab$ & stress (model output) \\
$\epsilon$ & strain (observable input) \\
$\epsilon^p$ & plastic strain (hidden model state) \\
\hline
$\mub$ & mean estimate of parameters \\
$\Sigmab$ & covariance estimate of parameters \\
$\As$ & parameter sensitivity of $\mb$ \\ 
$\Fs$ & hidden state transition \\
$\Rs$ & observation noise covariance \\
$\Ks$ & Kalman gain matrix \\
\hline
$\mathcal{S}$ & experiment states $\state$ \\
$\mathcal{A}$ & experiment actions $\action$ \\
$\mathcal{R}$ & reward \\
$v$ & value \\
$\policy$ & policy  \\
\hline
\end{tabular}
\caption{Notation for generic model, exemplar, Kalman filter and reinforcement learning method.} \label{tab:notation}
\end{table}

\section{Model calibration} \label{sec:calibration}

The calibration task is, given a model $\mb$:
\begin{equation}
\yb = \mb(\xb; \thetab)
\end{equation}
with parameters $\thetab$, find a path $\{ \xb_k \}$ that leads to the highest accuracy and lowest uncertainty in the calibrated parameters.
Here $\xb_k = \xb(t_k)$ is the input at discrete time $t_k$ and the sequence $\{ \xb_k, k=1,n\}$ represents an experimental protocol.
The total number of steps $n$ indicates the cost of the experiment.
The controls $\xb$ evoke an observable response $\yb$ from the physical system that $\mb$ models.
At each step, a finite number of actions are available to the machine controlling the experiment; choosing these actions is the subject of \sref{sec:RL}.

Our focus is on models where the current response $\yb_k$ depends on the current and previous inputs $\xb_i,\ i \le k$.
With a history-dependent model
\begin{equation}
\yb = \mb(\xb,\zb ; \thetab) \ ,
\end{equation}
latent variables $\zb$ are introduced that have their own evolution 
\begin{equation}
\dot{\zb} = \fb(\xb,\zb ; \thetab)
\end{equation}
and are typically hidden from observation.
In the realm of material physics the latent variables typically describe internal states that are linked to dissipation, such as plastic strain.
This irreversible behavior greatly complicates experimental design, since trial loadings can lead to permanent changes in the material.

\subsection{Exemplar: elastoplasticity} \label{sec:exemplar}
\begin{figure}[ht!]
    \centering
    \includegraphics[width=0.98\textwidth]{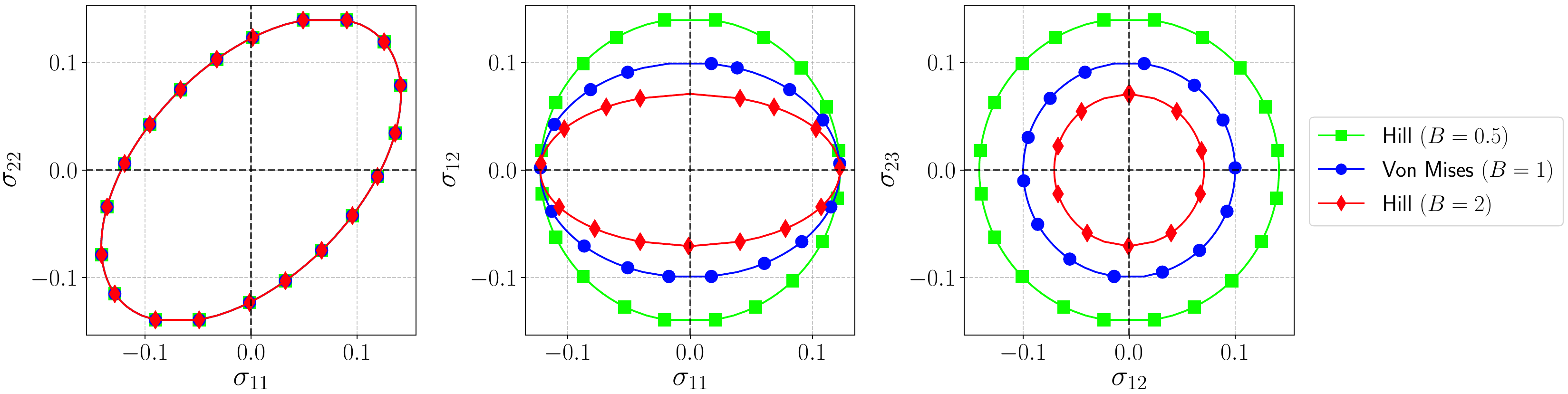}
    \caption{The modified Hill yield surface model under different stress axes for parameters $B=0.5$, $1$ and $2$.}
    \label{fig:hill_surface}
\end{figure}

One class of particularly technologically important examples of this problem type is the calibration of traditional elastoplasticity models \citep{lubliner2008plasticity, scherzinger2017return, simo2006computational}.
In these models the observable stress $\sigmab$ is a function of elastic strain $\epsilonb^e$.
Typically, a linear relationship between $\sigmab$ and $\epsilonb^e$ is assumed:
\begin{equation} \label{eq:stress}
\sigmab = \Cbb \epsilonb^e \ ,
\end{equation}
where $\Cbb$ is a fourth-order elastic-modulus tensor.
For instance, with transverse isotropy, 
\begin{eqnarray}
C_{1111} &=& E \frac{(1-\nu_\perp)}{(1-2 \nu^2 \nu_\perp)} \ , \\
C_{2222} = C_{3333} &=& E \frac{(1-\nu^2)}{(1-2 \nu^2 \nu_\perp)} \ , \nonumber \\
C_{1122} = C_{1133} &=& E \frac{\nu}{(1-2 \nu^2 \nu_\perp)} \ , \nonumber \\
C_{2233} &=& E \frac{(\nu^2+\nu_\perp)}{(1-2 \nu^2 \nu_\perp)(1+\nu_\perp)} \ , \nonumber \\
C_{1212} =  C_{1313} &=& E \frac{1}{(1-\nu)} \ ,  \nonumber \\
C_{2323} &=& E \frac{1}{(1-\nu_\perp)} \ ,  \nonumber
\end{eqnarray}
where $E$ is an effective Young's modulus, $\nu$ is an in-plane Poisson's ratio and $\nu_\perp$ is an out-of-plane Poisson's ratio.
History dependence is incorporated via plastic strain $\epsilonb^p$, which is a hidden material state variable that elicits dissipative behavior.
The elastic strain in \eref{eq:stress} is the difference between the controllable, observable total strain $\epsilonb$ and the irreversible plastic strain $\epsilonb^p$:
\begin{equation}
\epsilonb^e  = \epsilonb - \epsilonb^p \ .
\end{equation}
A closed, convex yield surface limits the elastic region and demarcates the elastic, reversible behavior in the interior of the surface from the irreversible plastic flow at the limit defined by the surface.
For instance, a modified/simplified Hill anisotropic yield surface 
\begin{eqnarray}
\label{eq:hill_surface}
Y = \phi(\sigmab) \equiv \biggl( \frac{1}{3} \left( 
(\sigma_{22} - \sigma_{33})^2 +
(\sigma_{11} - \sigma_{33})^2 +
(\sigma_{22} - \sigma_{11})^2 \right)  
+
\frac{B}{2} \left( \sigma_{23}^2 +
\sigma_{13}^2 +
\sigma_{21}^2  \right)
\biggr)^{1/2} \nonumber
\end{eqnarray}
generalizes the widely-used von Mises yield surface \citep{lubliner2008plasticity}; in fact, \fref{fig:hill_surface} shows that it reduces to von Mises when $B=1$.
The yield surface evolves with hardening of the material
\begin{equation}
Y = Y_0 + h(e^p) \ ,
\end{equation}
where $Y_0$ is the initial yield strength, $h$ is the hardening function and $e^p$ is the equivalent plastic strain.
For instance, $h = H e^p$ induces linear hardening.
The plastic strain evolves via the (associative) flow rule
\begin{equation} \label{eq:flow}
\dot{\epsilonb}^p = \dot{\lambda} \partialb_\sigmab \phi \ ,
\end{equation}
where the direction of evolution is given by the normal to the yield surface $\partialb_\sigmab \phi$.

The yield surface 
\begin{equation} \label{eq:yield}
g \equiv \phi(\sigmab) - Y(e^p) \le 0
\end{equation}
constrains the possible response of the material.
When $g < 0$, the material is in an elastic state, and the material state variables $\epsilonb^p$ and $\lambda = e^p$ are fixed so that the stress at the new state $k$ is
\begin{equation} \label{eq:elastic_step}
\sigmab_k = \Cbb (\epsilonb_k - \epsilonb^p)
= \sigmab_{k-1} + \Cbb ( \Delta \epsilonb) \ .
\end{equation}
Otherwise, the material is in a plastic state; the evolution equations and the constraint $g=0$ need to be solved through a Newton iteration with increments $\Delta\sigmab^{(i)}$
\begin{equation} \label{eq:plastic_step}
\sigmab_k = \sigmab_{k-1} + \sum_i \Delta\sigmab^{(i)} \ .
\end{equation}
This aspect complicates obtaining parameter sensitivities.
Further details can be found in \cite{simo2006computational}.

For this exemplar the parameters are $\thetab = \{ E, \nu, \nu_\perp, B, Y_0, H \}$. 
If $Y_0\to\infty$ the model reduces to elasticity, and if $\nu_\perp = \nu$ it reduces to isotropic elasticity, $\thetab = \{ E, \nu\}$.
If $Y_0$ is finite, $\nu_\perp = \nu$ and $B=1$ it reduces to the widely-used von Mises plasticity model,  $\thetab = \{ E, \nu, Y_0, H \}$.

\subsection{Kalman filter for calibration}

For simplicity we will assume we have an experiment where we can control all components of strain and observe the stress of the material sample we would like to model; hence the calibration data consists of a sequence of strain-stress input-output pairs.
There are many applicable methods to obtain parameter estimates given calibration data \citep{sun2015model}, such as nonlinear least squares regression.
Here we use an online method, the Kalman filter (KF), that provides concurrent uncertainty quantification.
The KF estimates a parameter covariance as well as a parameter mean since it is essentially an incremental Bayesian update of the parameters.
It also provides a probabilistic estimate of the response.
The current context presents a few complications that require an enhanced KF: (a) the model is nonlinear with respect to the inputs and parameters, (b) it is not smooth with respect to the parameters and (c) it is history-dependent.

\begin{remark}
In our design of experiments, application the KF is primarily tasked with guiding the experiments; after the data is collected, the model could be re-calibrated using other methods e.g. standard Bayesian calibration.
\end{remark}

\subsubsection{Extended Kalman filter} \label{sec:EKF}

The complication of calibrating a nonlinear model with the Kalman filter, which was developed \citep{kalman1960new} for models that are linear in their parameters, can be handled with linearization.
In the extended Kalman filter (EKF) \citep{jazwinski2007stochastic}, the state transition and observation models are linearized to maintain the usual Kalman update formula.
Since we assume the stress is the only observable variable, the observation model is provided by the material model $\mb$, and the appropriate parameter sensitivities are
\begin{equation}
\As_{k}=\partialb_\thetab \mb \big\rvert_{\thetab_{k-1}, \xb,\zb} \ ,
\end{equation}
where $k$ is the step, so that the linearization
\begin{equation}
\mb(\xb,\zb;\thetab_k) \approx \mb(\xb,\zb;\thetab_{k-1}) +  \As_k [\thetab_k - \thetab_{k-1}]
\end{equation}
is sufficiently accurate.
This linearization provides a mapping from the parameter covariance $\Sigmab$ to the observable output covariance $\As \Sigmab \As^T$.
\begin{remark}
Note that for a linear model $\yb = \thetab \xb$, like the elastic response described in \sref{sec:exemplar}, the sensitivities $\As$ increase with $\xb$.
This relationship has ramifications on the KL divergence reward discussed in \sref{sec:RL}.
\end{remark}

We assume the (observable) data $\ds$ corresponds to the model plus uncorrelated (measurement) noise
\begin{equation}
\ds = \mb(\xb,\thetab) + \varepsilonb \ ,
\end{equation}
where parameters $\thetab \sim \mathcal{N}(\mub, \Sigmab)$ and noise $\varepsilonb \sim \mathcal{N}(\mathbf{0}, \Rs)$ follow normal distributions.
The (continuous time, noiseless) hidden state transition model is 
\begin{eqnarray}
\dot{\thetab} &=& \mathbf{0} \\
\dot{\zb}     &=& \fb(\xb,\zb ; \thetab) \ ,
\end{eqnarray}
where states comprised of model parameters $\thetab$ and hidden (material) state $\zb = \{ \epsilon^p, \lambda \}$.
Since the parameters $\thetab$ are fixed, their state transition is the identity.
In discrete time, the state transition model become difference equations.
Due to the yield surface constraint \eref{eq:yield}, the  plasticity model exemplar is actually a system of differential algebraic equations (DAEs).
Handling DAEs in a KF requires additional care;
\aref{app:DAE_KF} describes a rigorous treatment based on the work of \cite{catanach2017computational}.
For the results given in \sref{sec:results}, we merely 
perform the chain rule 
\begin{equation}
\As = \partialb_\theta \mb(\xb,\zb; \thetab) = \partialb_\thetab \mb + \partialb_\zb \mb \, \partialb_\thetab \zb \ ,
\end{equation}
which accounts for the change in $\zb$ over the step, but not the entire history.

The residual, residual covariance and so-called Kalman gain are
\begin{eqnarray}
\rs_{k} & = & \ds_{k}-\mb(\xb_k,\zb_k,\mub_{k-1})\\
\Ss_{k} & = & \As_{k}\Sigmab_{k-1}\As_{k}^{T}+\Rs\\
\Ks_{k} & = & \Sigmab_{k-1}\As_{k}^{T} \Ss_{k}^{-1} \ ,
\end{eqnarray}
respectively.
They are used to update the parameter ($\thetab$) mean $\mub$ and covariance $\Sigmab$:
\begin{eqnarray}
\mub_{k}       & = & \mub_{k-1}+\Ks_{k} \rs_k\\
\Sigmab_{k} & = & \Sigmab_{k-1}-\Ks_{k}\Ss_{k}\Ks_{k}^{T}
\end{eqnarray}
given the data $\ds_k$ provided at step $k$.

The iterative KF method requires initialization.
Initial values of the mean $\mub_0$ and covariance $\Sigmab_0$ represent the prior information of the parameters $\thetab$.
In the present context the noise variance $\Rs$ is fixed and a scaling of the identity matrix with the scale chosen {\it a priori} to be small; however, it can be calibrated as well.
Moreover, in this case, the diagonal of $\Rs$ primarily regularizes the inversion of $\Ss$.
Predictions can be made using the current values of the mean parameters $\mub$ and their covariance $\Sigmab$:
\begin{equation}
\ys^* \sim  \mathcal{N}(\mb(\xb^*,\mub),\As^*\Sigmab\As^{*T}) \ .
\end{equation}

\subsubsection{Switching Kalman filter} \label{sec:SKF}

Plasticity models consist of two response modes, (a) an elastic one where only some of the parameters are influential, and (b) a plastic one where all parameters affect the response. 
A customization of the EKF is therefore necessary to adapt to these physical regimes and avoid errors in accumulation of latent variables in the plastic mode.
With the exemplar in \sref{sec:exemplar}, no information on the plastic response is available from the material in its reference/starting state.
If the exact state at which the material changes to a plastic response was known, it would be trivial to select the appropriate KF; unfortunately, uncertainty in the yield stress causes erroneous Kalman updates of the parameters and hidden state.

Two solutions are implemented to handle the discontinuity in response and the switching between elastic and plastic modes.
The first sets a convergence tolerance criteria on the parameters. 
This {\it ad hoc} version of the EKF sets parameter sensitivities to zero when convergence is detected by masking the sensitivity matrix $\As$ with a diagonal matrix $\Ms$ that has unit entries for parameters that have not converged and zeros for ones that have.
This masking has the effect of fixing the converged parameters at their current mean. 
In particular it allows the elastic parameters to be calibrated first while the material state is still within the yield surface, and then the calibration of the plastic parameters ensues, with fixed elastic parameters, once yield is encountered.
For this simple method, convergence is assessed with
a Cauchy convergence criterion on the mean $\mub$ for the particular parameter and a check that the current diagonal entry of the covariance $\Sigmab$ has decreased from its previous value. 

Alternatively, an extended {\it switching} Kalman filter (SKF) is a more rigorous approach and can deal with discontinuous parameters, such as elastic vs. plastic response, by simultaneously competing multiple models (in this case, material modes).
This competition allows for a more accurate prediction of material behavior when there is a latent variable with abrupt or discontinuous changes. 
We implement a Generalized Pseudo Bayesian (GPB) algorithm \citep{Murphy98switchingkalman}, which takes Gaussian distributions associated with each material mode and collapses them into a Gaussian mixture of (at most) two previous history steps.
The order 2 GPB is a good balance between accuracy and the complexity of longer views of the history.

In SKF, the material modes can be assigned to a discrete switching variable $\Psi_k \in \{0,1\}$, where $\Psi_k$ can take values of 0 (elastic) or 1 (plastic) at every step $k$ depending on the likelihood of the active mode. 
We partition our exemplar into two submodels $\mb^e$ and $\mb^p$, which represent elastic, \eref{eq:elastic_step}, and plastic, \eref{eq:plastic_step}, modes, respectively. 

Since we do not know {\it a priori} when switching occurs, we assign each mode a probability; $\pdf(\Psi_0| \thetab)$ is the prior probability before we collect any data, and  $\pdf(\Psi|\ds_{1:k}; \thetab)$  is the probability after we collect data, where $\ds_{1:k}$ is data observed up to step $k$ and $\thetab$ are the model parameters shared between both modes. 
The mode probability $\pdf(\Psi)$ update equation is derived from Bayes rule for the joint probability $\pdf(\Psi_{k-1}=\alpha,\Psi_{k}=\beta)$ given the data $\ds$ up to the current step $k$:
\begin{align}
\pdf(\Psi_{k-1}=\alpha,\Psi_{k}=\beta  | \ds_{k},\ds_{1:k-1}) & \propto \pdf(\Psi_{k-1}=\alpha,\Psi_{k}=\beta,\ds_{k} | \ds_{1:k-1})\\
 & =\pdf(\ds_{k}|\Psi_{k-1}=\alpha,\Psi_{k}=\beta,\ds_{1:k-1})
 \pdf(\Psi_{k-1}=\alpha,\Psi_{k}=\beta | \ds_{1:k-1}) \nonumber\\
 & =\underbrace{\pdf(\ds_{k} | \Psi_{k-1}=\alpha, \Psi_{k}=\beta, \ds_{1:k-1})}_{L_{k}(\alpha,\beta)} \nonumber \\
 & \times \underbrace{\pdf(\Psi_{k}=\beta | \Psi_{k-1}=\alpha,\ds_{1:k-1}) }_{Z(\alpha,\beta)} \nonumber 
 \underbrace{\pdf(\Psi_{k-1}=\alpha|\ds_{1:k-1})}_{M_{k-1|k-1}(\alpha)} \nonumber \ ,
\end{align}
where $\alpha, \beta \in \{0,1\}$, with $0$ corresponding to $\mb^e$ and $1$ to $\mb^p$.
The likelihood function $L_k(\alpha,\beta)$ is a multivariate Gaussian distribution $ L = \mathcal{N}(\mathbf{0}, \As \Sigmab \As^T + \Rs) $  based on the parameter distributions.
The function $Z(\alpha,\beta)$ is the transition matrix for modes $\Psi$, and $M_{k|k}(\beta)= \sum_{\alpha}M_{k-1,k|k}(\alpha,\beta)$ is the posterior distribution for $\Psi$ at step $k-1$.
If the transition is partitioned between modes, the switching becomes soft, and the Kalman filter is effectively a mixture of the two discrete filter modes.
Since the material response is either elastic or plastic, we manipulate the transition matrix to be binary.
When the material is more likely to deform elastically with a probability of $\pdf(\Psi_k=0|\ds_{1:k},\thetab_k) > 1/2$, the prior parameters $\thetab_k$ are updated according to $\mb^e$;  else, if the material begins to deform plastically with probability $\pdf(\Psi_k=1|\ds_{1:k}, \thetab_k) > 1/2$, then the model parameters are updated according to $\mb^p$. 
Treating the calibration as separate modes allows for better estimation of both elastic and plastic parameters considering that new data can be partitioned appropriately by the mode assignment.

\section{Deep reinforcement learning for experimental design with extended Kalman filter} \label{sec:RL}

\begin{figure}[ht!]
\centering
\includegraphics[width=0.65\textwidth]{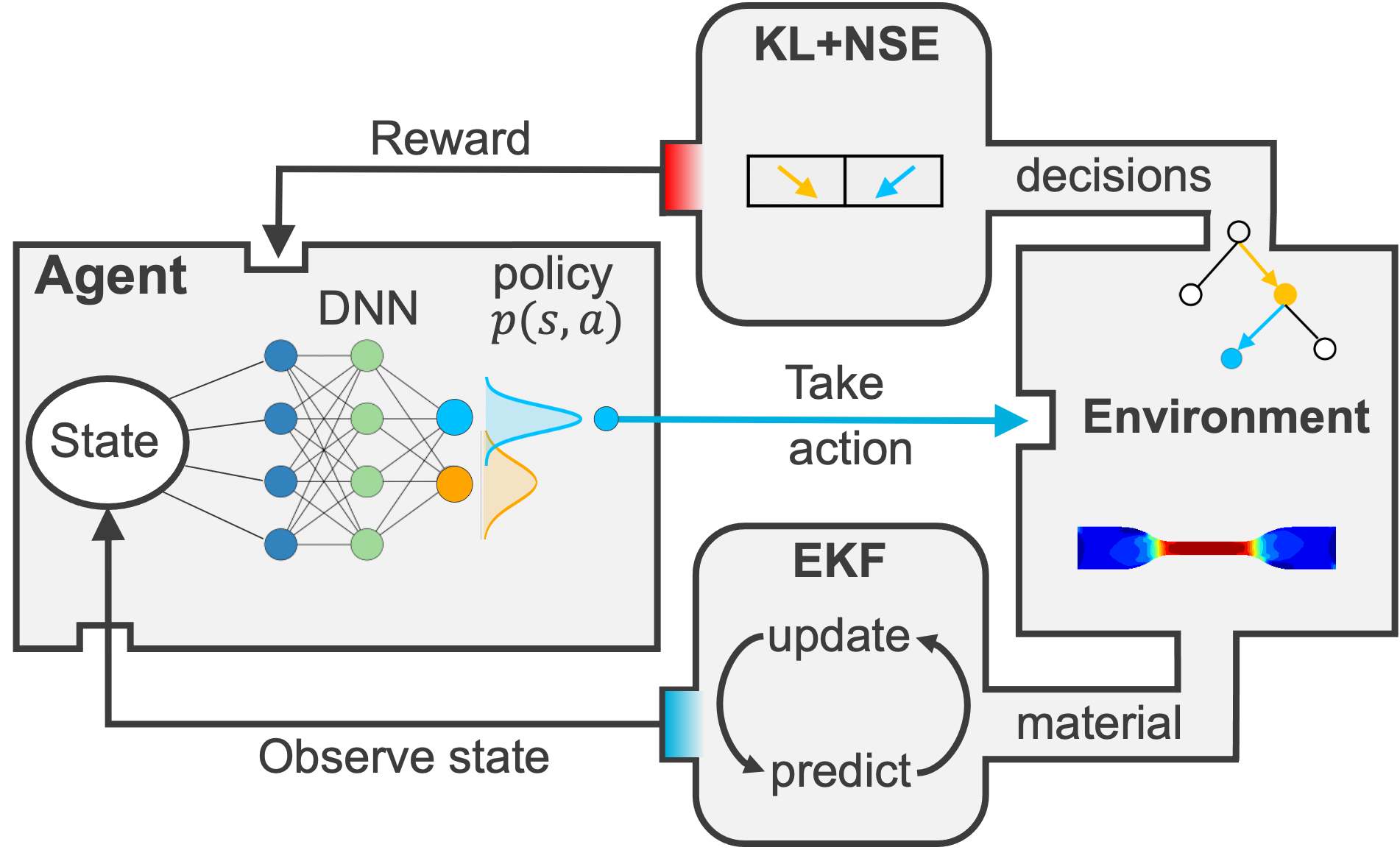}
\caption{Workflow of the KF-based deep reinforcement learning.}
\label{fig:workflow}
\end{figure}

This section describes the incorporation of the extended Kalman filter (EKF) into deep reinforcement learning (DRL) for experimental designs that maximize information gain.
Based on the reward hypothesis of reinforcement learning  \citep{niv2009reinforcement}, we assume that the goal of a calibration experiment can be formalized as the outcome of maximizing a cumulative reward defined by the information gain estimated from the EKF.
As such, the design of the experiment that generates a model can be viewed as a game where an agent seeks feasible actions in an experiment to maximize the total information gain, and hence the best-informed model parameters, as shown in \fref{fig:workflow} which illustrates the workflow of the proposed RL approach.

In this section, we have two objectives: (1) to describe how an experiment is run as a MDP in a RL framework enhanced by the EKF, and (2) to provide some key highlights on how  the DRL agent that runs the experiment is trained.
\sref{sec:MDP} first formulates a family of mechanical tests as a MDP.
Then, the policy that gives the action selection is described in \sref{sec:policy} and the action of the experiments represented by a decision tree is described in \sref{sec:action}.
The action-state-reward relationship quantified by the EKF is described in \sref{sec:state}.
With these ingredients properly defined, the Monte Carlo tree search (MCTS) used to update the policy and improve the DRL agent's decision-making process is provided in \sref{sec:MCTS}.

\subsection{Experiments as a Markov decision process} \label{sec:MDP}

Here, we consider an experiment conducted by an agent as a single-player game formulated as MDP where the agent interacts with the environment based on its state in a sequential manner.
This MDP can be a tuple $(\mathcal{S}, \mathcal{A}, \state, \gamma)$ consisting of state set $\mathcal{S}$, action set $\mathcal{A}$, and joint probability of reward $\mathcal{R}$ for a given state $\state$ and discount factor $\gamma$ that balances the relative importance of earlier and later rewards.

For the experiment we set the RL {\it actions} with all the allowed experimental control actions that affect the strain $\epsilonb$, and the RL {\it state} set as all the results of possible actions over a sequence of steps.
Even for a discrete set of $n$ actions the state space increases exponentially with time starting from the reference/initial state and growing $n$-fold with every step.

In an {\it episode}, the agent makes a sequence of decisions of (discrete) actions $\Delta\epsilonb_k$ to take.
An episode is a complete traversal of the decision tree from the root node to a leaf node that corresponds to the end state of an experiment after a fixed number of steps.
In effect an episode is a particular experiment defined by selected control actions $\{\xb_i,i=1,k\}$.
For each decision within the same episode, an agent takes an action based on the policy $\policy(\state,\action) \in [0,1]$ that suggests the probability of the preferred choice. 
Each action taken by the agent must lead to an update of the (observable) state $\yb_k$. 
This updated state is then used as a new input to generate the next policy value for the available actions. 
This feedback loop continues until the particular experiment/episode concludes.

The RL {\it environment} is defined as any source of observations that provides sufficient information to update the state.  
In this work, our goal is to introduce the KF as a component of the environment where the estimation of the KF state is used to: (1) provide an enhanced state representation, as well as (2) constitute the reward of the experiments to redefine the objective of the game, which is now formulated to maximize the information gain of the experiments.
Finally, learning occurs whenever the deep neural network that predicts the policy is retrained in an RL {\it iteration} \citep{silver2017masteringchess, wang2019cooperative}. 
An iteration can be called upon whenever a sufficient amount of new state-action pair labels is collected during the experiments. 
The ratio between the number of iterations and episodes as well as the architecture of the neural network itself are both hyperparameters that can be fine-tuned for optimal performance.

\subsection{Policy represented by deep neural network} \label{sec:policy}

In principle, the RL {\it policy} that guides the agent to select rewarding actions can be determined by directly sampling the states, actions and rewards among all the existing options. 
However, such an approach is not feasible when the number of possible paths to conduct experiments becomes too large.
A classical example includes the game of Go \citep{silver2017masteringb} where exhausting all the possible moves is intractable.
This large number of paths is also common in experiments where a sequence of decisions has to be made both before and during the experiments.
More discussion of this point is given in \sref{sec:results}. 

\begin{figure}[ht!]
    \centering
    \includegraphics[width=0.75\textwidth]{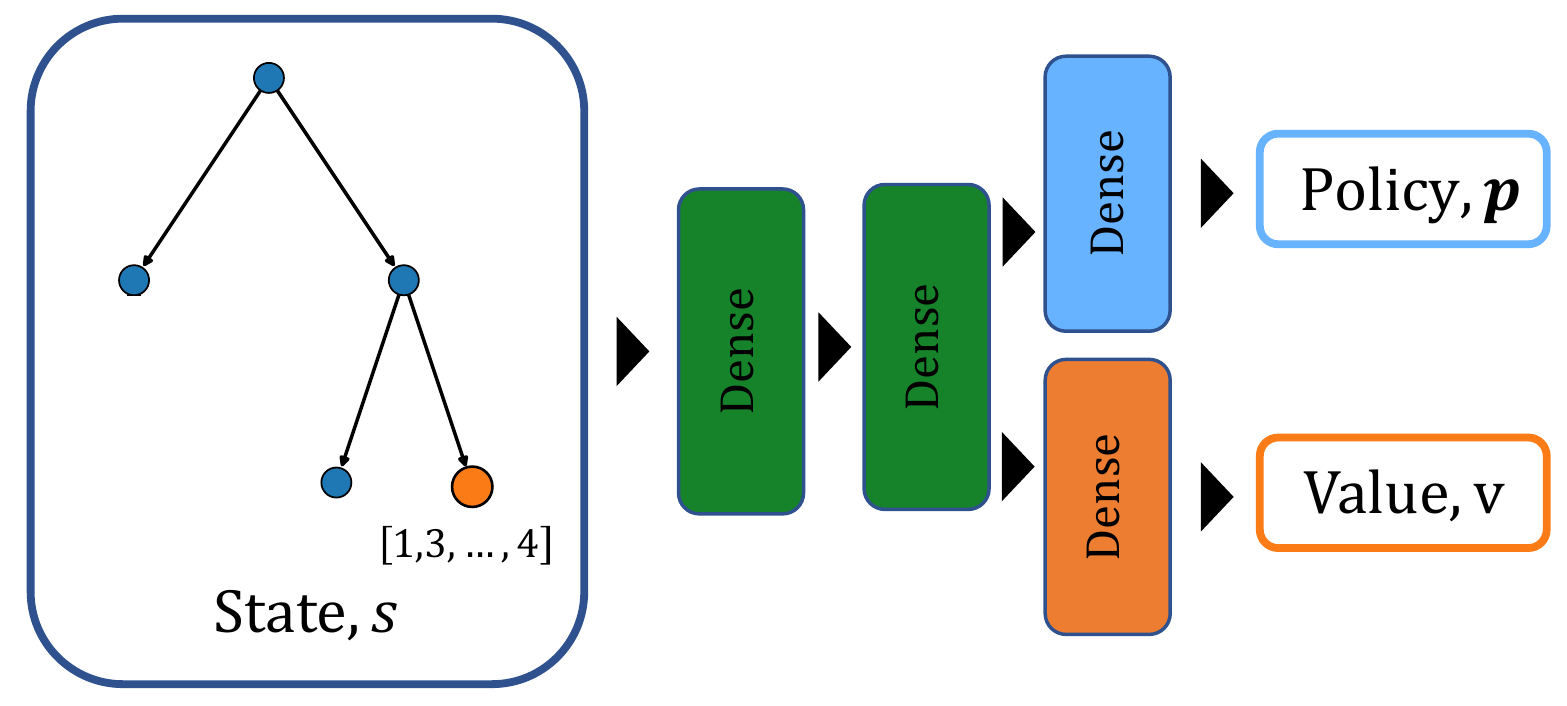}
    \caption{A schematic of the reinforcement learning policy-value neural network architecture. The network inputs a state $s$ corresponding to a node of the decision tree and outputs a policy $\policy$ and value $v$.}
    \label{fig:drl_net_schematic}
\end{figure}

As shown in, for instance \citet{silver2017mastering, silver2017masteringb, silver2017masteringchess}, an option to manage this curse of high dimensionality is to approximate the policy (and potentially other sets in the tuple) via a trained neural network and use MCTS to improve the efficiency of the sampling.
In our work, the deep neural network is solely designed for the purpose of generating the policy value when given a particular combination of state $\state \in \mathcal{S}$ and action $\action \in \mathcal{A}$. 
Note that policy function can also be approximated by other techniques, including mathematical expressions obtained from symbolic regression \citep{landajuela2021discovering}  and the choice of the approximation may affect the difficulty of the representation problem.
In this paper, a standard overparameterized deep neural network is adopted throughout the entire training process. 
The neural network follows a standard multilayer perceptron feed-forward architecture. 
The network takes RL state $\state$ that corresponds to a node of the decision tree as inputs and is subsequently fed into a series of dense hidden layers. 
The network has two outputs: a policy vector $\policy$ representing the probability of taking the action $\action$ from the current state $\state$ and a predicted scalar value $v$ estimating the reward from the state (see \fref{fig:drl_net_schematic}). 
A policy is a probabilistic function in part to allow for exploration as well as exploitation of previous information such as the rewards for previous actions.
Since the policy represents a probability, a $\operatorname{softmax}$ layer is used for this output, while a $\operatorname{tanh}$ layer is used for the continuous value output.
Additional specifics of the network's architecture and training hyperparameters are provided in \sref{sec:results}.

The mapping between the input state $\state$ and the output policy $\policy$ and value $v$ is, thus, represented by a neural network approximator $\widehat{f}$. 
The approximation is defined such that $\left(\widehat{\policy},\widehat{v}\right)=\widehat{f}\left( \state \mid \boldsymbol{W}, \boldsymbol{b}\right)$, where $\widehat{\policy}$ and $\widehat{v}$ are the approximated values of the policy vector and value, respectively, and $\boldsymbol{W}$ and $\boldsymbol{b}$ are the weight matrices and bias vectors of the architecture, respectively, to be optimized with stochastic gradient descent during the network training.
The training objective for the training samples $i \in [1, ..., N]$ is to reduce the mean squared loss:
\begin{equation}
    \boldsymbol{W}^{\prime}, \boldsymbol{b}^{\prime}=\underset{\boldsymbol{W}, \boldsymbol{b}}{\operatorname{argmin}}\left(\frac{1}{N} \sum_{i=1}^N\left(\left\|\policy_i-\widehat{\policy}_i\right\|_2^2+ \left\|v_i-\widehat{v}_i\right\|_2^2\right)\right) \ .
\end{equation}

\begin{remark}
Note that this policy corresponds to the reward but is not the reward itself, as the policy must be a trade-off between exploration and exploitation \citep{sutton2018reinforcement}. This balance will be further discussed in \sref{sec:MCTS}.
\end{remark}

\subsection{Action representation: decision tree for experiments} \label{sec:action}

In a Markov decision process, a state from an earlier decision is connected to all of the corresponding possible child states through an action. 
Furthermore, an earlier action may affect the latter states but a latter decision has no effect on the prior state. 
Hence, all the possible actions and states together are connected in a directed and acyclic manner, which, in graph theory \citep{west2001introduction} is referred as a {\it poly-tree}. 
We will employ the terminology of graph theory and refer the initial state of an experiment as the {\it root} (the vertex with only outgoing edged in the tree) and the end of the experiment as the {\it leaf} (a vertex with only an incoming edge). 
As such, the decision tree of an experiment is a specific poly-tree that represent all the possible states at vertices, each connected by edges representing the corresponding actions available during an experiment. 
The role of the policy of the RL agent (the actor) is to determine the action when a state is given as input such that an RL agent may create a path that started from the root and end at one of the leaves of the decision tree. 

\subsection{Environment: states and rewards of the design-of-experiment problem}
\label{sec:state}

As pointed out by \citet{reda2020learning}, a key ingredient to generate the effective learned policy and value of the states that often get overlooked is the parameterization of the environment, in which the DRL agent/algorithm interacts with.
In the design-of-experiment problem, the environment provides feedback caused by actions selected by the DRL agent.
This feedback can be in the form of a state or reward.  

We formulate the design-of-experiment as a multi-objective problem in which we want to (1) minimize the expected value of the discrepancy between the predictions made by the calibrated models and the ground truth (by maximizing the Nash-Sutcliffe efficiency (NSE) index of the \textbf{calibrated model}, see \eref{eq:NSE}) and (2) improve the efficiency of the \textbf{experiments} by maximizing the Kullback-Leiber (KL) divergence, and hence maximizing the information gain between states. 
While both measurements provide a valuation of the actions through measuring the improvement of the model due to additional data gained from new actions in the experiment, the KL divergence does not require additional sampling, and hence is more cost efficient. 
However, the extended Kalman filter (EKF) used to calculate the KL divergence also has well-known limitations, such as the need to make a sufficiently close initial guess to avoid divergence triggered by the linearization process and the consistency issue due to the underestimation of the true covariance matrix. 

As such, we employ a mixed strategy in which we only approximate the NSE index by under-sampling a few states outside of the training data region. 
This cheaper approximated NSE index is augmented with the KL divergence as the combined reward to circumvent the inconsistency and divergence issue of the EKF, whereas the  KF-predicted mean and variance of the calibrated parameters are used, in addition to the loading history of the experiments, to represent the enhanced state of the experiment.

For brevity and to avoid confusion with the estimated added reward within each state update, we would use the term game score to refer to the total reward accumulated within an experiment game, while the history of the performances is measured by monitoring the distribution of the game score against the policy neural network iteration at which the policy neural network is re-trained. 

\subsubsection{Information-gain reward: Kullback-Leibler divergence}
\label{sec:kl}

The KL divergence has a lower-bound value of 0 when the model perfectly describes the data, meaning that no information was gained.
Similarly, a perfect NSE score has an upper-bound value of 1 when the model perfectly replicates the data.
There is no direct relationship between the NSE score and the KL divergence, but both measure aspects of model accuracy and parameter uncertainty.

Generally speaking, the KL divergence is a measure of the statistical change between two probability distributions. 
Denoted as $D(\pdf_1||\pdf_0)$ where $\pdf_0$ is a reference prior probability distribution and $\pdf_1$ is the updated posterior probability distribution.
The difference is calculated as the expectation of the logarithmic difference between distributions with respect to the posterior probabilities $\pdf_1(x)$:
\begin{equation}
D_\text{KL}(\pdf_1||\pdf_0)=\int_{\mathcal{X}}\pdf_1(x)\log\Bigg(\frac{\pdf_1(x)}{\pdf_0(x)}\Bigg) \ ,
\end{equation}
where $x\in X$ represents a shared probability space.

\begin{figure}[ht!]
\centering
\includegraphics[width=0.65\textwidth]{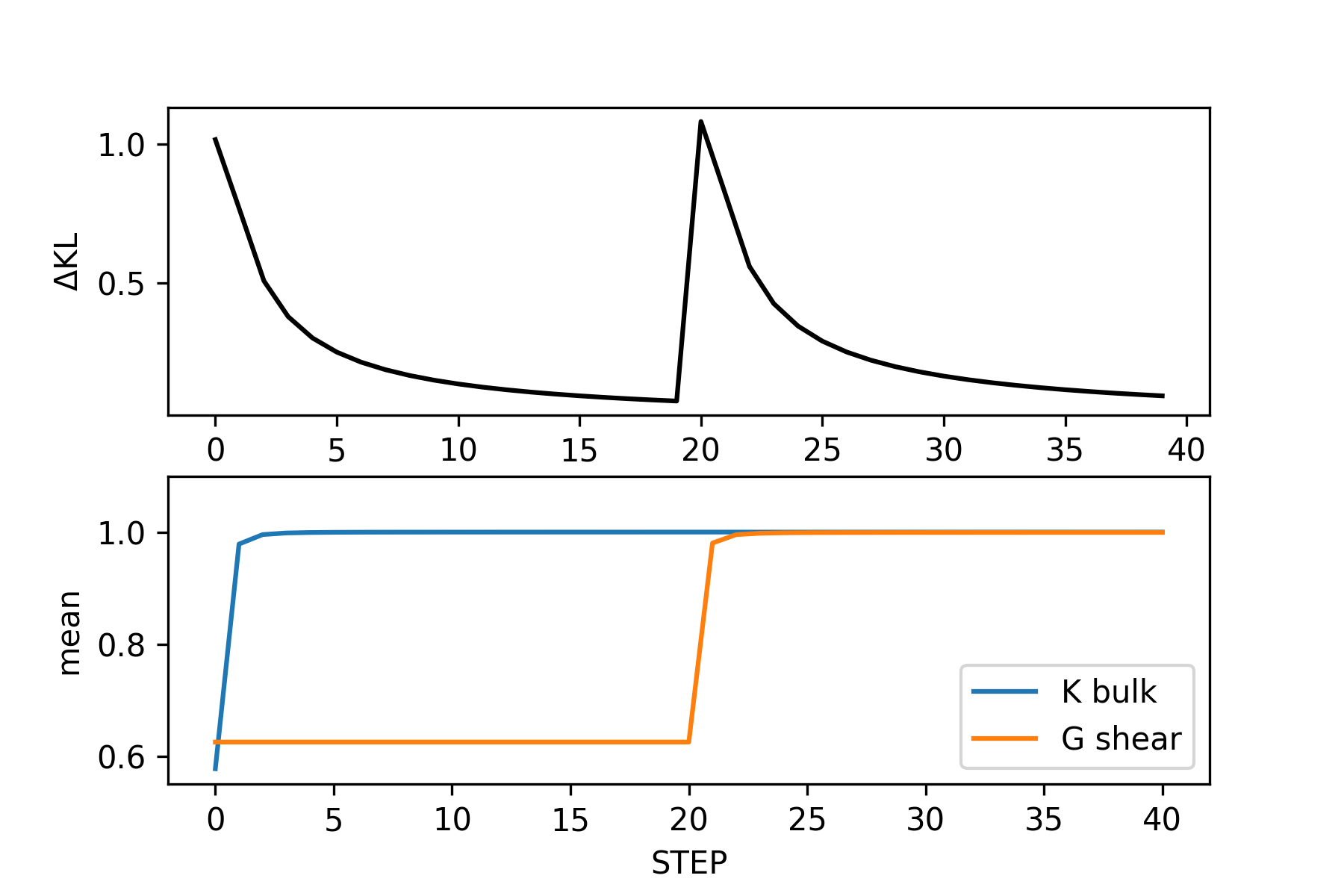}
\caption{Calibration of the isotropic elastic material model (bottom), and the incremental KL-based reward along the calibration path (top) that starts with a volumetric deformation and then switches to shear.
}
\label{fig:elastic_reward}
\end{figure}

By formulating a reward based on the KL divergence, we can get a measure of how informative new data $\ds_{n+1}$ is in updating the distribution of the estimated parameters $\pdf(\thetab | D)$ (see \fref{fig:elastic_reward}):
\begin{equation}
\Delta\text{KL}=\int \pdf(\thetab|D_{n+1})\log\frac{\pdf(\thetab|D_{n+1})}{\pdf(\thetab)} \mathrm{d}\thetab
-\int \pdf(\thetab|D_{n})\log\frac{\pdf(\thetab|D_{n})}{\pdf(\thetab)} \mathrm{d}\thetab \ ,
\end{equation}
where $D_n = \{ \ds_k, k \le n \}$.
For a multivariate Gaussian distribution, like the one we have with the KF, the KL divergence is analytic:
\begin{equation}
\text{KL}_k=
\frac{1}{2} \left(
\log \frac{\det \Sigmab_{0}}{\det \Sigmab_{k}} + \tr\left( \Sigmab_{0}^{-1} \Sigmab_{k} \right) + (\mub_k - \mub_{0}) \Sigmab_0^{-1} (\mub_k - \mub_{0})  - n_\thetab
\right) \ ,
\end{equation}
where $n_\thetab$ is the number of parameters.
A reward for the experiment over $n$ steps can be formulated as
\begin{equation}
\label{eq:knowledge_gain_reward}
\mathcal{R}_\text{KL} = \sum_{k=1}^n \Delta\text{KL}_k \ ,
\end{equation}
where $\Delta\text{KL}_k \equiv \text{KL}_k - \text{KL}_{k-1}$.

For example, the reward for a single step in a linear elastic material with bulk modulus $K$ and shear modulus $G$ has covariance
\begin{equation}
\Sigmab_{k} =    \begin{bmatrix}
\operatorname{var}(K)_{k} & \operatorname{cov}(K,G)_{k}\\
\operatorname{cov}(G,K)_{k} & \operatorname{var}(G)_{k}
\end{bmatrix}
\label{eq:covariance}
\end{equation}
and mean vector of the estimated bulk and shear moduli
\begin{equation}
    \mub_k = \begin{bmatrix}
K_{k}\\
G_{k}
\label{eq:meanparameter}
\end{bmatrix} \ .
\end{equation}
As illustrated in \fref{fig:elastic_reward},  the distribution of the parameters begins with a pre-selected {\it prior} value for the mean $\mub_0$ and covariance $\Sigmab_{0}$ before measuring any data.
The KL divergence initially increases since the new data is more informative, but then the rewards tail off as less information is gained from subsequent samples and the mean parameter values converge.
When the material response exhibits a discontinuity, as in the onset of plastic deformation, the Kalman switching filter \sref{sec:SKF} can be used to select the deformation mode (elastic or plastic) that best captures the data at the current step. 
\sref{sec:von_mises} provides a detailed illustration of this switching mechanism.

\subsubsection{State represented by action history and Kalman filter prediction}

The RL {\it state} represents the complete information necessary to describe the consequences of actions taken by the agent from the beginning of the game to the current step.

In this work, the state of the experiment contains two components: (1) the entire loading history of the experiments since the beginning of the test and (2) the calibrated mean and covariance of the material parameters (see, for instance,  Eqs. \eqref{eq:covariance} and \eqref{eq:meanparameter}). 
Depending on the types of experimental tests, the loading history can be represented via different parametrizations \citep{heider2020so}.
For a strain-controlled test, the available action choices are the increments of individual components of the strain tensor, and hence the loading history can be represented by a stack of these strain increments stored in the Voigt notation, which leads to a matrix of dimension $6 \times N_{\text{step}}$ where 6 is the number of independent components of the symmetric strain tensor and $N_{\text{step}}$ is the total number of strain increments. 
For future time steps that are not yet executed, the corresponding columns of the matrix are set to zero. 
For the case where the loading combination is more limited, such as a shear box apparatus, loading history can be sufficiently represented by a vector. 

This state is augmented by the mean and covariance of the material parameters predicted by the EKF. 
For instance, in the case of linear elasticity, the state can be represented by the mean of two elastic material parameters and the components of the symmetric covariance matrix, which contains three independent components.
In other words, the state representation of the experiment employs both a representation of the loading path and the parameter distribution provided by the EKF. 
Together, they form the RL state, which is used as the input for the policy neural network shown in \fref{fig:drl_net_schematic}. 

\begin{remark}
The decision points for the MDP can be on a larger step size than the KF, i.e., the KF can be sub-cycled for additional stability and accuracy.
\end{remark}

\begin{remark}
Note that while incorporating more sensory information may provide more information to learn the policies and values, in practice adding more information may also increase 
the dimensionality of the state representation, and hence 
significantly increase the difficulty of the DRL. 
The exploration of more efficient state representation methods and the implications of more efficient state representations are active research areas \citep{schrittwieser2021online}, but are out of the scope of this study.
\end{remark}

\subsubsection{Forecast prediction reward via an under-sampled Nash-Sutcliffe efficiency index}
\label{sec:efficiency_index}

The Nash-Sutcliffe efficiency index is a simple normalized measure of the discrepancy between model predictions and ground truth data. 
The NSE index is:
\begin{equation}
\mathcal{R}_\text{NSE} =
1-\frac{\sum_{k=1}^{N_{\text{data}}}|\ds_{k}-\mb(\thetab)|}{\sum_{k=1}^{N_{\text{data}}}|\ds_{k}-\operatorname{mean}(\ds)|}\in(-\infty,1] \ ,
\label{eq:NSE}
\end{equation}
where $\ds_k$ is the data point at step $k$, $\mb(\thetab)$ is the calibrated model at the end of an episode and $\operatorname{mean}(\ds)$ is the mean value of the dataset.
To measure the mean of the forecast accuracy via the NSE index, the number of data points $N_{\text{data}}$ must be sufficiently large such that empirical loss and population loss lead to a sufficiently small difference (refer to \citet{gnecco2008approximation}).
Hence, this sampling requirement can be costly, especially for physical experiments that require labor, time and material costs. 
As such, we propose an under-sampling and static strategy where the data points $d_{k}$ collected to calculate the NSE reward are sampled prior to the training of the DRL agent \citep{wang2019cooperative, wang2021non}. 
This reduction on the sampling size may, in principle, leads to increasing bias by missing data of statistical significance. 
However, since the sub-reward $\mathcal{R}_{\text{NSE}}$ introduced here merely functions as a regularization term for the KL reward in \eref{eq:knowledge_gain_reward} to guide the early exploration of the DRL agent, no significant issues manifested in the numerical experiments showcased in \sref{sec:results}. 

\subsubsection{Combined reward for multi-objective experiments}
\label{sec:mixed_reward}
As mentioned earlier, approximating the parameter distributions of highly nonlinear function via the standard EFK may lead to inaccurate state estimation. 
This inaccuracy, in turn, can inflate the KL divergence and leads to an overoptimistic reward for the DRL agent \citep{laviola2003comparison}. 
While there are alternatives, such as unscented Kalman filter \citep{julier1997new} and the ensemble Kalman filter \cite{evensen2003ensemble} that can circumvent the loss of accuracy due to the linearization of the nonlinear function, a simpler implementation approach is to add the NSE reward into the DRL framework during the training process to modify the exploitation behavior of the DRL agent. 

As such, we introduce a weighted average reward that augments the KL divergence with the under-sampled NSE index: 
\begin{equation}
\label{eq:mixed_reward}
    \mathcal{R}_{\text{total}} = w_\text{NSE}\mathcal{R}_\text{NSE} + w_\text{KL}\mathcal{R}_\text{KL} \ ,
\end{equation}
where $w_\text{NSE}$ and $w_\text{KL}$ are weighting factors for the two rewards. 
To promote the neural network representation training, the rewards in the following applications are rescaled to be order 1 based on an estimate of the range of expected values.
When using the mixture of the two rewards, $\mathcal{R}_{\text{NSE}}$ and $\mathcal{R}_{\text{KL}}$, we rescaled each to the range of $[0,1]$ and choose the weights such that $w_{\text{NSE}} + w_{\text{KL}} = 1$ so that the total reward is in the same range.
Unless stated otherwise, all the numerical experiments are conducted with equally weighted sub-reward, i.e., $w_{\text{NSE}} = w_{\text{KL}} = 1/2$. 

\begin{remark}
Note the assumption of a fixed step budget is not overly constraining. 
If the incremental reward tails off, and the parameter uncertainties are acceptable, the experiment can be truncated early, potentially with a significant cost savings.
\end{remark}

\begin{remark}
There is distinction between (a) the reward calculated from the KL divergence, NSE index or a combination of the two, and (b) the state value and expected cumulative reward predicted by the policy. 
As in the RL literature, we use {\it reward} to refer to the former and {\it value} to reference the latter.
\end{remark}

\subsection{A Monte Carlo tree search with Kalman reward estimator} \label{sec:MCTS}

Here, we consider the case where planning and learning are both needed to maximize the information gain within a limited number of actions. The learning objective is the optimal calibration of a material model.
For costly physical experiments, the goal of designing an experiment is not just to finish a task (e.g., calibration, discovery and uncertainty quantification), but to finish the task within the allocated resources.
The multitude of decisions complicate this goal.
For instance, in a biaxial compression/extension test where the specimen can be compressed/extended in two directions, there will be $4^{50} \approx 10^{30}$ possible ways to run an experiment with 50 incremental time steps.

Sampling all the available paths and selecting actions that maximize the optimal reward can be a feasible strategy only if all paths can be visited.
The classical Monte Carlo simulation is not feasible as random sampling is not sufficiently efficient to discover the optimal policy given limited opportunities to visit only a small fraction of the possible paths.
As such, a tactic to balance exploitation and exploration is necessary \citep{moskovitz2021tactical}. 

In this work, Monte Carlo tree search (MCTS) is used to enable estimation of the policy value $\policy(\state, \action)$ by visiting state-action pairs according to an optimality equation, \eref{eq:action}, that balances exploitation and exploration of the decision tree. 
It is necessary to setup a material simulator and model calibrator before beginning the policy search in algorithm \ref{alg:cap}. 
The output of these components constitute the observable variables in the RL environment that help the agent learn and improve its policy. 
Code implementations of the simulator and calibrator are not covered here in detail, but the material exemplar is found in \sref{sec:exemplar}. 
An KF is used for both evaluating the reward and calibrating material parameters. 
The EKF and the SKF methods we empolyed are described in \sref{sec:EKF} and \sref{sec:SKF}. 

Once the policy DNN is initialized (see \sref{sec:linear_elastic_results}), the iterator $i$ starts the outer loop. 
Each iteration executes a policy update after an inner loop (iterator $j$) over a number of episodes, where the outcome of each episode represents the result of the current policies for the design of the experiment. 
Update of the policy DNN is accomplished with a stochastic gradient descent optimizer, see \sref{sec:policy}.
During an episode, the decision tree is populated with state and action pairs until the end of the tree is reached. 
Each action is selected according to a probability of moving to a state of maximum value. 
This value is estimated by a policy $\policy_{i}(\state, \action)$ at the $i^\text{th}$ iteration.

At the end of each episode, the history of control actions taken are fed as input into the material simulator and calibrator. 
The reward $\mathcal{R}_{KL}$ assigned to an episode is calculated using the KL divergence \eref{eq:knowledge_gain_reward} which is based on the mean and covariance posterior updates. 
The KL divergence is a measure of the expected amount of information gained about the system state. 
A higher reward indicates that the agent is learning more about the system as it informs the model calibration. 
The reward can also be calculated according to \eref{eq:NSE} or a mixture, \eref{eq:mixed_reward}.
At the conclusion of each iteration, the policy is updated using the training examples generated during the episode simulations (step 26, algorithm \ref{alg:cap}).
After several iterations, the policy becomes a good estimator of action value and the exploitation of high value actions will be balanced by \eref{eq:action} which diminishes the probability of selecting repetitive pathways. 

\begin{algorithm}
\caption{Reinforcement learning for Design of Experiments}\label{alg:cap}
\begin{algorithmic}[1]
\Require The definitions of the experiment game: environment, states, actions, rewards.
\State Initialize the experimentalist policy/value network DNN. For fresh learning, the network is randomly initialized. For transfer learning, load pre-trained network instead.
\For{i in iterations} 
    \State Initialize empty sets of the training examples $trainExamples \leftarrow \{\}$.
    \For{j in episodes} 
        \State Initialize the starting game state vector $\state$ (container for experiment control history).
        \State Initialize empty tree of the Monte Carlo Tree search (MCTS), by setting containers for edge visits $N(\state, \action)$, and mean action values $Q(\state, \action)$  
        \While{True} 
            \State Check for all allowed actions at the current state $\state$ according to the games rules.
            \State Get the action probabilities $\policy(\state,.)$ for all allowed actions by performing repeated MCTS simulations.
            \State Sample action $\action$ from the probabilities $\policy(\state,.)$
            \State Modify the current game state to a new state $\state$ by taking the action $\action$.
            \If{$\state$ is the end state of the game of the experimentalist}
                \State \textbf{break}
                \EndIf
                \EndWhile
        \State Calibrate material model using EKF with the selected paths in the decision tree.
        \If{the information-gain reward (\sref{sec:kl}) is used}
                \State Evaluate $\mathcal{R}_\text{KL}$ from the model calibration. 
                \State Evaluate the total reward $\mathcal{R}_{\text{total}} = \mathcal{R}_\text{KL}$ of this gameplay.
                \EndIf
        \If{the Nash-Sutcliffe efficiency index reward (\sref{sec:efficiency_index}) is used}
                \State Test calibrated model against set of blind experiments and evaluate $\mathcal{R}_\text{NSE}$. 
                \State Evaluate the total reward $\mathcal{R}_{\text{total}} = \mathcal{R}_\text{NSE}$ .
                \EndIf
        \If{the combined reward (\sref{sec:mixed_reward}) is used}
                \State Evaluate $\mathcal{R}_{KL}$ from the model calibration. 
                \State Test calibrated model against set of blind experiments and evaluate $\mathcal{R}_\text{NSE}$. 
                \State Evaluate the total reward $\mathcal{R}_{\text{total}} = w_\text{NSE}\mathcal{R}_\text{NSE} + w_\text{KL}\mathcal{R}_\text{KL}$ of this gameplay.
                \EndIf
        \State Append the gameplay history $[\state,\action,\policy(\state,.),\mathcal{R}_{\text{total}}]$ to $trainExamples$
    \EndFor
    \State Train the policy/value network DNN with $trainExamples$
    \EndFor
\State Use the trained network DNN of the last iteration to select the optimal experiments for model calibration.
\State Exit
\end{algorithmic}
\end{algorithm}

\begin{figure}[ht!]
    \centering
    \includegraphics[width=0.75\textwidth]{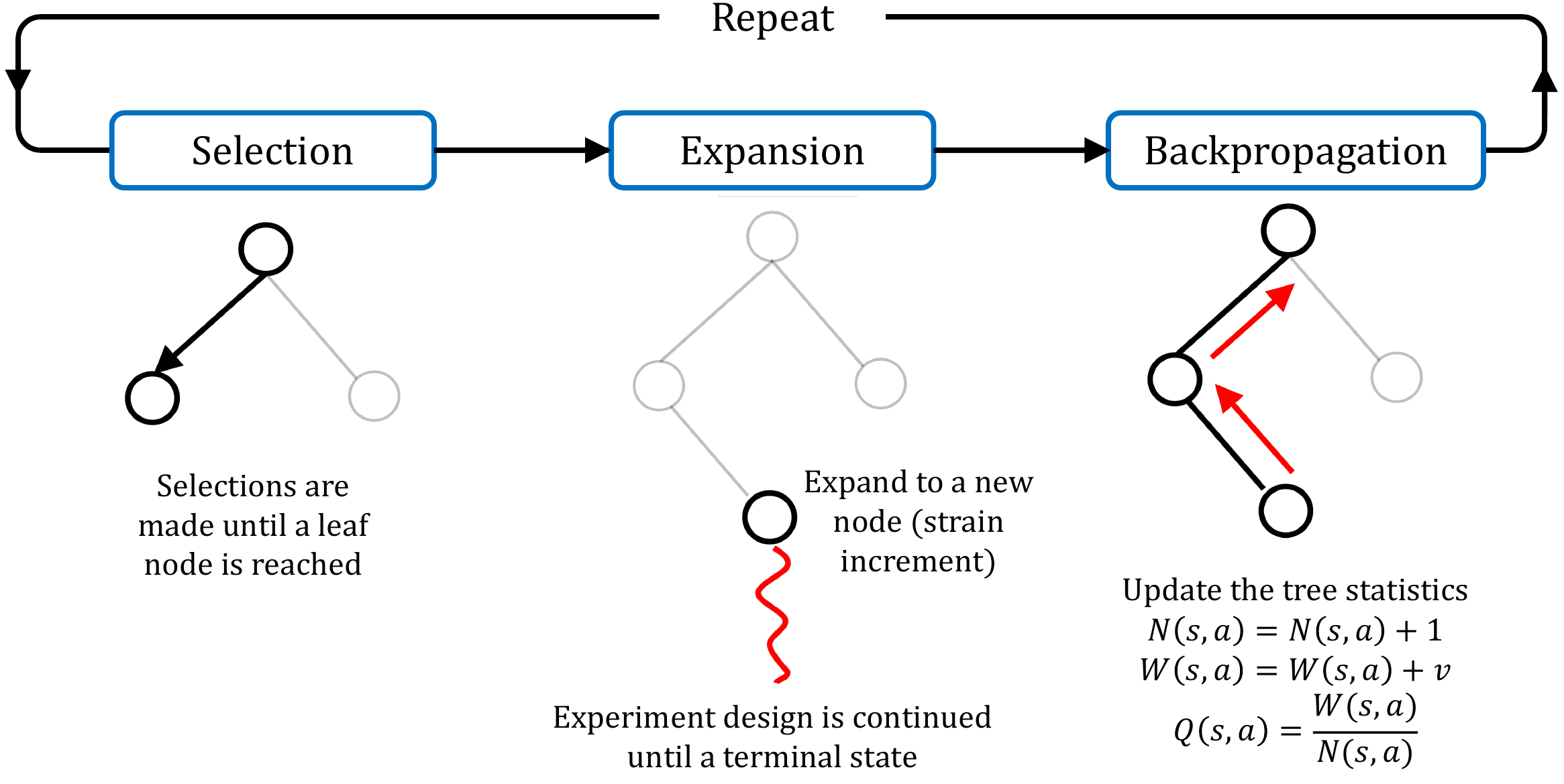}
    \caption{Monte Carlo tree search steps.
    }
    \label{fig:mcts}
\end{figure}

Algorithm \ref{alg:mcts} details and  the MCTS in the inner loop in algorithm \ref{alg:cap} (7-13). 
Within one episode, the MCTS we employed repeatedly performs the three steps illustrated in \fref{fig:mcts}) \citep{silver2017mastering}:

\begin{algorithm}
\caption{Monte Carlo tree search}\label{alg:mcts}
\begin{enumerate}
    \item Selection. A path is determined by picking action according to the estimated policy $\policy(\state,\action)$ (until a leaf of the tree (the state node with no child) is reached), i.e.,  
    \begin{equation}
\action_{t} = \underset{\action \in \mathcal{A}}{\operatorname{argmin}} \:\Bigg(Q(\state, \action)+c_{p u c t} \policy(\state, \action) \frac{\sqrt{\sum_{\action'} N(\state, \action')}}{1+N(\state, \action)}\Bigg) \ ,
\label{eq:action}
\end{equation}
where $\action_{t}$ is the chosen action, $c_{p u c t}$ is a parameter that controls the degree of exploration. $N(\state, \action)$ denotes the number of visit/time action $a$ is taken at state $\state$. 
In \eref{eq:action}, the first term is the $Q$ value in our case, which is the expected value of the reward, i.e.,
\begin{equation}
Q(\state, \action) = \mathbb{E}[\mathcal{R}_{\text{total}} | \action_{t} = \action, \state_{t} = \state] \ ,
\end{equation}
whereas the second term is the upper confidence bound, which can be derived from Hoeffding's inequality \citep{sutton2018reinforcement}.
Here we simply average the action value we have collected from $N(\state,\action)$ sampling as the $Q(\state,\action)$ value. 
Meanwhile, the policy values $\policy(\state,\action)$ are estimated from the deep neural network trained in the last iteration.

The key feature of this action selection model is that it reduces the value of the second term for a given action when it is visited more frequently. 
Consequently, this reduction triggers a mechanism for the agent to explore actions with high uncertainty if given the same expected return. 
    \item Expansion. The available options of the experiment for a given state is added to the tree. Generally speaking, options may vary for different states. In this work, the available options of actions are identical for each state. The allowed actions from every state are the same increments of the strain tensor component. The number of allowed actions/strain component increment options depends on the complexity of the respective material and respective decision tree.
    
    \item Back-propagation. At the terminated state, the KL divergence and other feasible indices that yield the reward are calculated (see \fref{fig:workflow}). All the policies between the root node and leaf nodes will be updated. The visit count $N(s,a)$ at every node traversed is increased by $1$ and the action value $Q(\state,\action)$ is updated to the mean value.
\end{enumerate}
\end{algorithm}

Once the search is complete, the search probabilities/policies $\pi$ are evaluated based on how often a state was traversed:
\begin{equation}
    \pi(a \mid s)=\frac{N(s, a)^{1 / \tau}}{ \sum_b N(s, b)^{1 / \tau}} \ ,
\end{equation}
where $N$ is the visit count of each move from the root node and $\tau$ is a temperature parameter, which is another parameter controlling the exploration.
Thus, the discovered policies are proportional to the number of visits in each state.

After a fixed number of episodes, there will be enough labeled data of state, action, policy and reward to update the policy network. At this point, an {\it iteration} of the policy network is conducted, and all the collected data within the episode will be used to retrain the policy neural network.
In this context, iteration refers to a policy update which is conducted after a certain number of episodes have collected sufficient new reward data.

\section{Numerical experiments} \label{sec:results}
In this section, we introduce three design-of-experiments examples to (1) validate the implementation of the EKF-DRL algorithm, (2) provide benchmarks against the classical DRL approach that employs the Nash-Sutcliffe sampling to estimate the rewards, and (3) showcase the potential applications of the EKF-DRL algorithm for designing mechanical experiments for models of high-dimensional spaces that require a significantly larger decision tree for long-term planning due to history dependence and other complications.

\subsection{Implementation verification 1: experiment for linear isotropic elastic materials} \label{sec:linear_elastic_results}

For a linear isotropic elasticity model, there are two independent elastic moduli.
Hence two linearly independent observations are sufficient to provide the necessary information to determine the model parameters \citep{bower2009applied}.
For example, a single step in a uniaxial test should be sufficient to identify any pairs of independent linear elastic parameters if the stress is observed in independent directions (e.g., the Poisson effect is observed as a supplement to the surface traction and uniaxial stress along the loading direction).

\begin{figure}[ht!]
    \centering
    \includegraphics[width=0.35\textwidth]{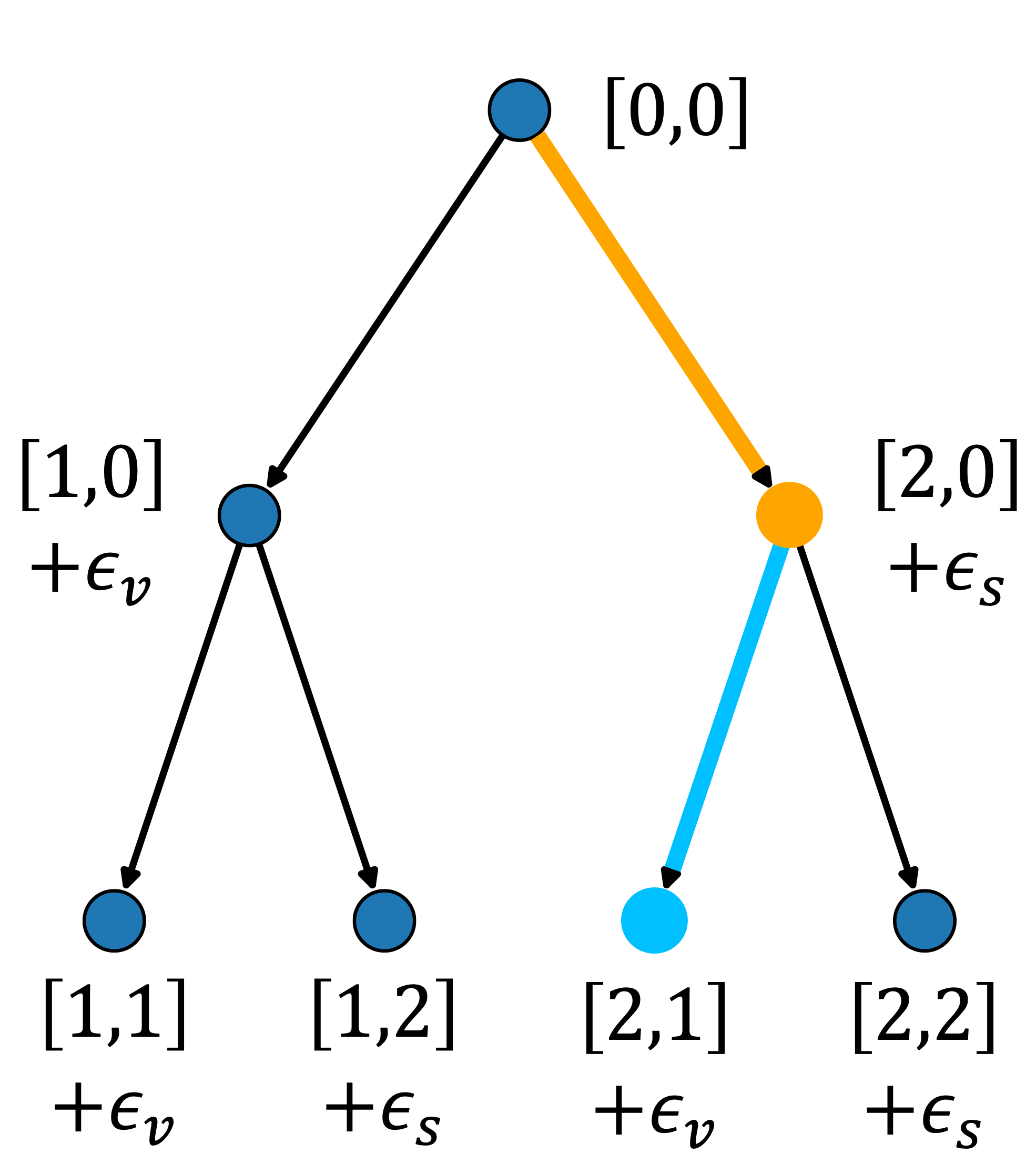}
    \includegraphics[width=0.35\textwidth]{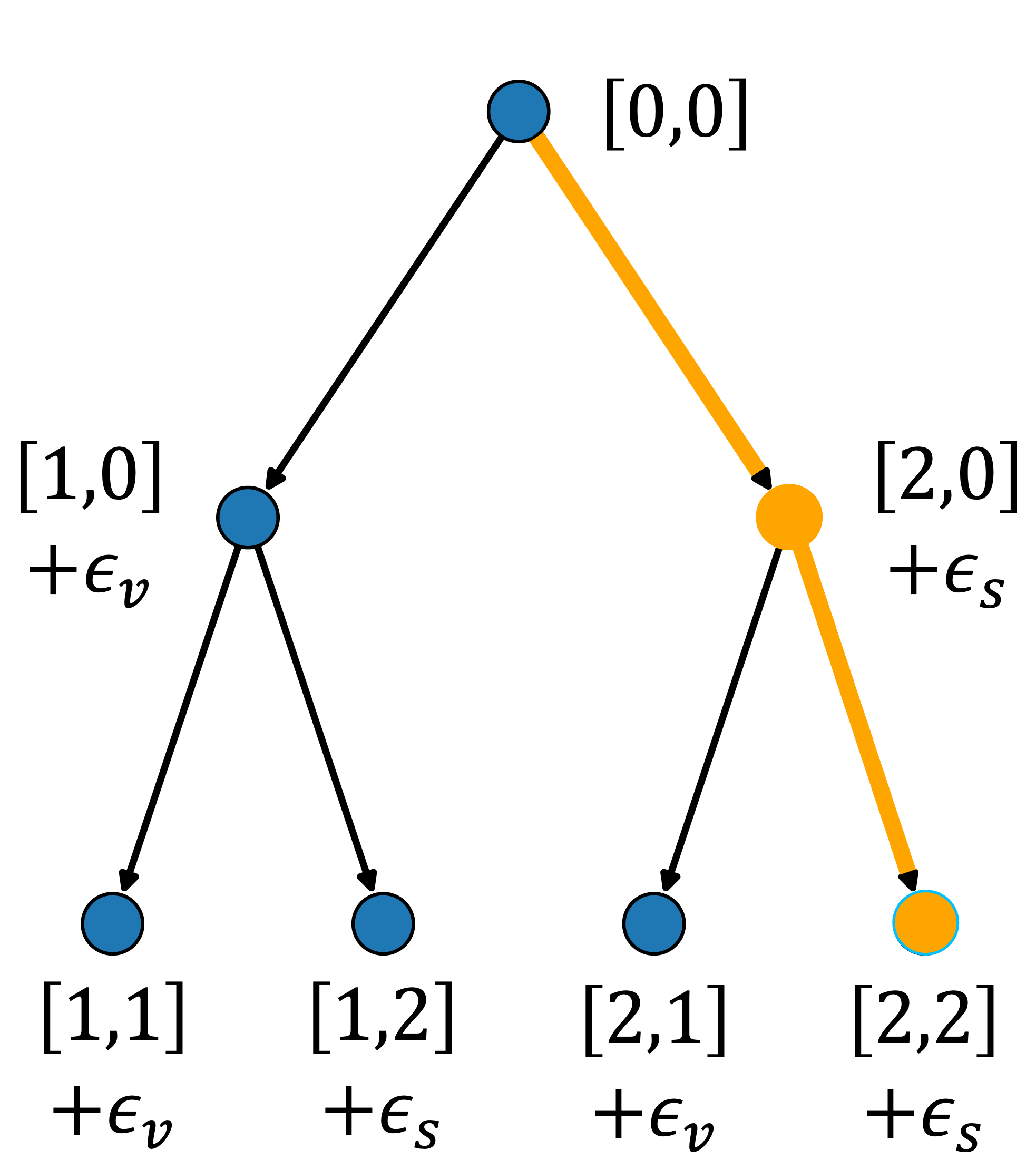}
    \includegraphics[width=0.35\textwidth]{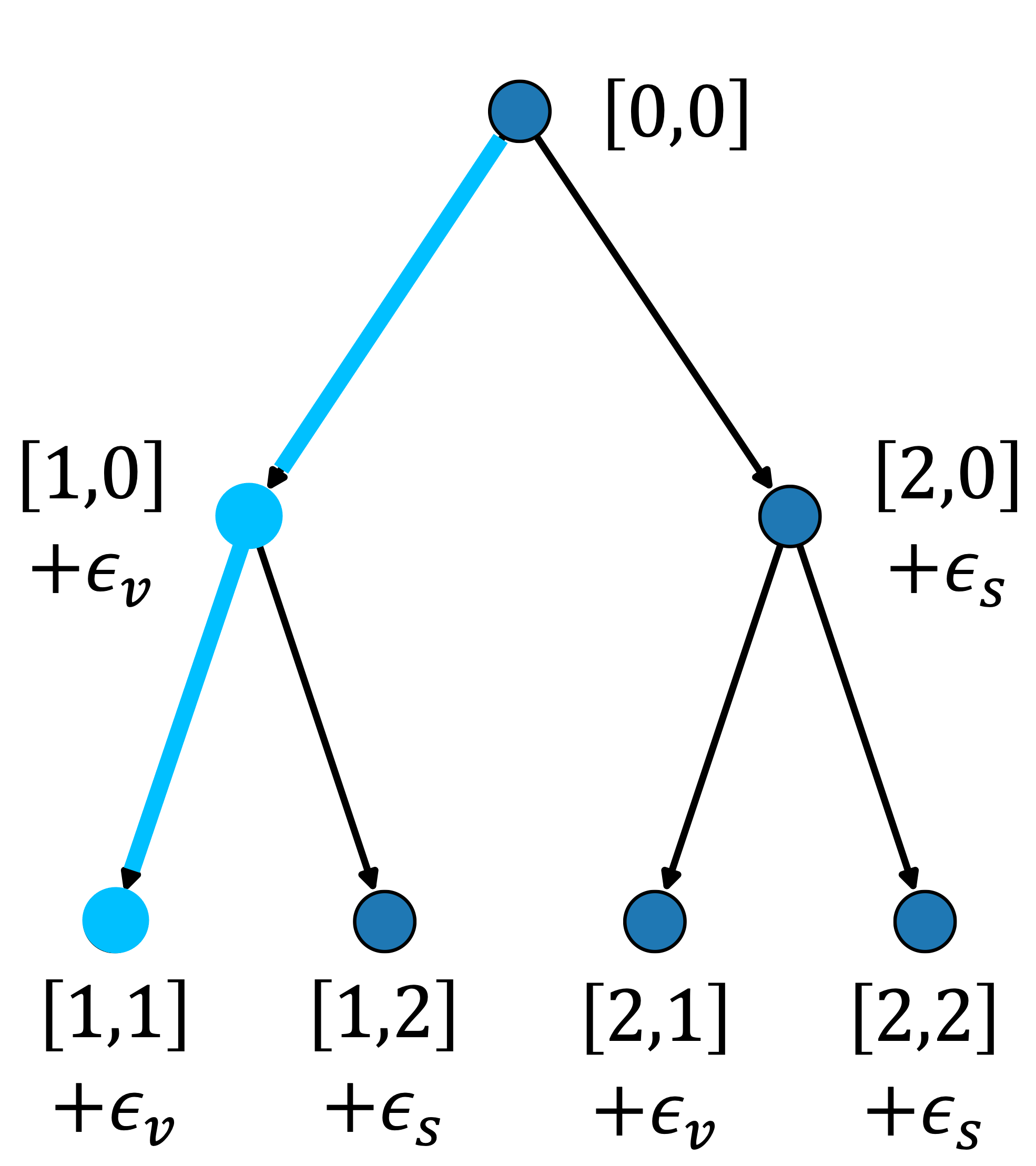}
    \includegraphics[width=0.35\textwidth]{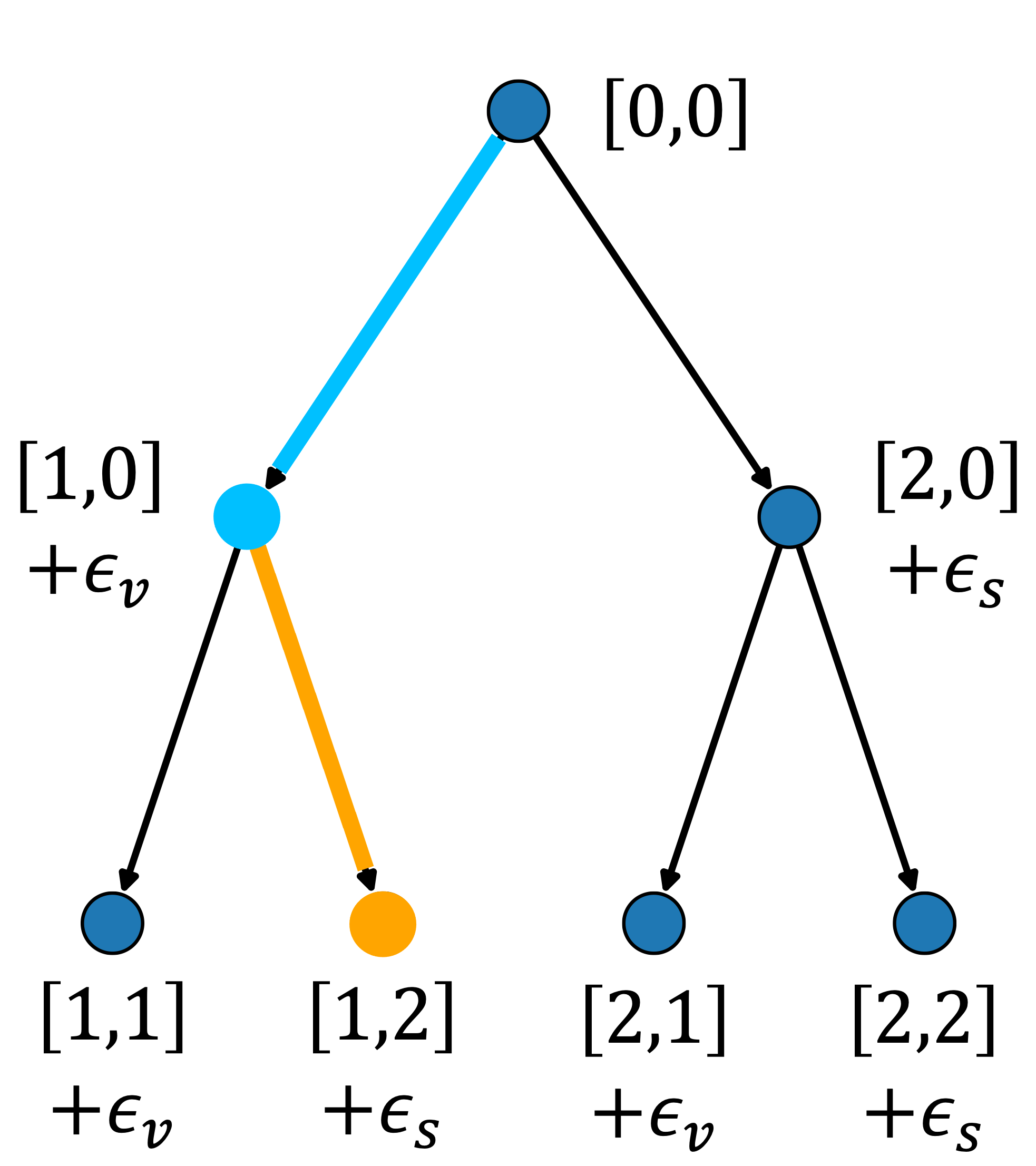}
    \caption{All possible paths (orange/blue lines) of the decision tree for the volumetric/deviatoric test within two steps. The weight of the nodes stores the state, which represents the current strain of the specimen, and the edge weight is the strain increment. The integers in the state vector indicate 0 for no action, 1 for compression and 2 for shear, where the first and second components of the state vector record the choice made by the agent.}
    \label{fig:linearelastictree}
\end{figure}

In this numerical experiment, we introduce a virtual test where the EKF-DRL agent observes the volumetric and deviatoric stress of a linear isotropic elastic material whenever the corresponding volumetric and shear strain are prescribed.
This agent is then tasked with designing a strain-controlled mechanical test of a specimen to identify the bulk $K$ and shear $G$ moduli.
The decision tree that includes the possible paths for two incremental steps is shown in \fref{fig:linearelastictree}.

Since the specimen is strain-controlled in two directions, there are two optimal strategies: (1) first shear then compress the specimen or (2) first compress then shear the specimen. 
As with model-free Q learning \citep{gu2016continuous}, one can simply visit all the possible states and the optimal strategy will be learned. 
As such, the goal of this numerical example is to verify the implementation where the optimal design is trivial in this sense.

A synthetic measurement model was used to verify convergence to the correct policy as a benchmark for the RL algorithm. 
A training experiment was performed for 10 policy training iterations.
Every iteration has 10 game episodes. 
As mentioned, each episode corresponds to a complete traversal of the decision tree from the root node to a leaf node to design the experiment strain path.
This includes gathering the linear elasticity data, calibrating the Kalman filter (KF) model on that data and calculating the information gain reward.
During each episode, we gather information for the states $\state$ traversed as well as the corresponding policies $\policy$ and values $v$. 
At the end of every iteration, the RL neural network is trained on the $(\state,\policy,v)$ data collected as described in \sref{sec:policy}. 
The exploration parameter \eref{eq:action} is set to linearly reduce every iteration, starting at $c_{p u c t}=10$ and being equal to $c_{p u c t}=1$ at iteration 10. 
This parameter was chosen to encourage exploration more in earlier iterations, sample the tree choices more uniformly, and avoid accumulating bias towards one decision tree path.

The policy-value neural network utilized for this example had two hidden dense layers with a width of 50 neurons each and Rectified Linear Unit ($\operatorname{ReLU}$) activation functions.
The policy vector output layer had two neurons (equal to the maximum number of allowed actions on the tree) and a $\operatorname{softmax}$ activation function.
The (scalar) value output layer had one neuron and a $\operatorname{tanh}$ activation function. 
The kernel weight matrix was initialized with a Glorot uniform distribution and the bias vector with a zero distribution for every layer.
At the end of every iteration, the model architecture and optimized weights from the previous iteration are reloaded and trained for 100 epochs with a batch sample size of 32 using an Adam optimizer \cite{kingma2014adam}.
 
\begin{figure}[ht!]
    \centering
    \includegraphics[width=0.45\textwidth]{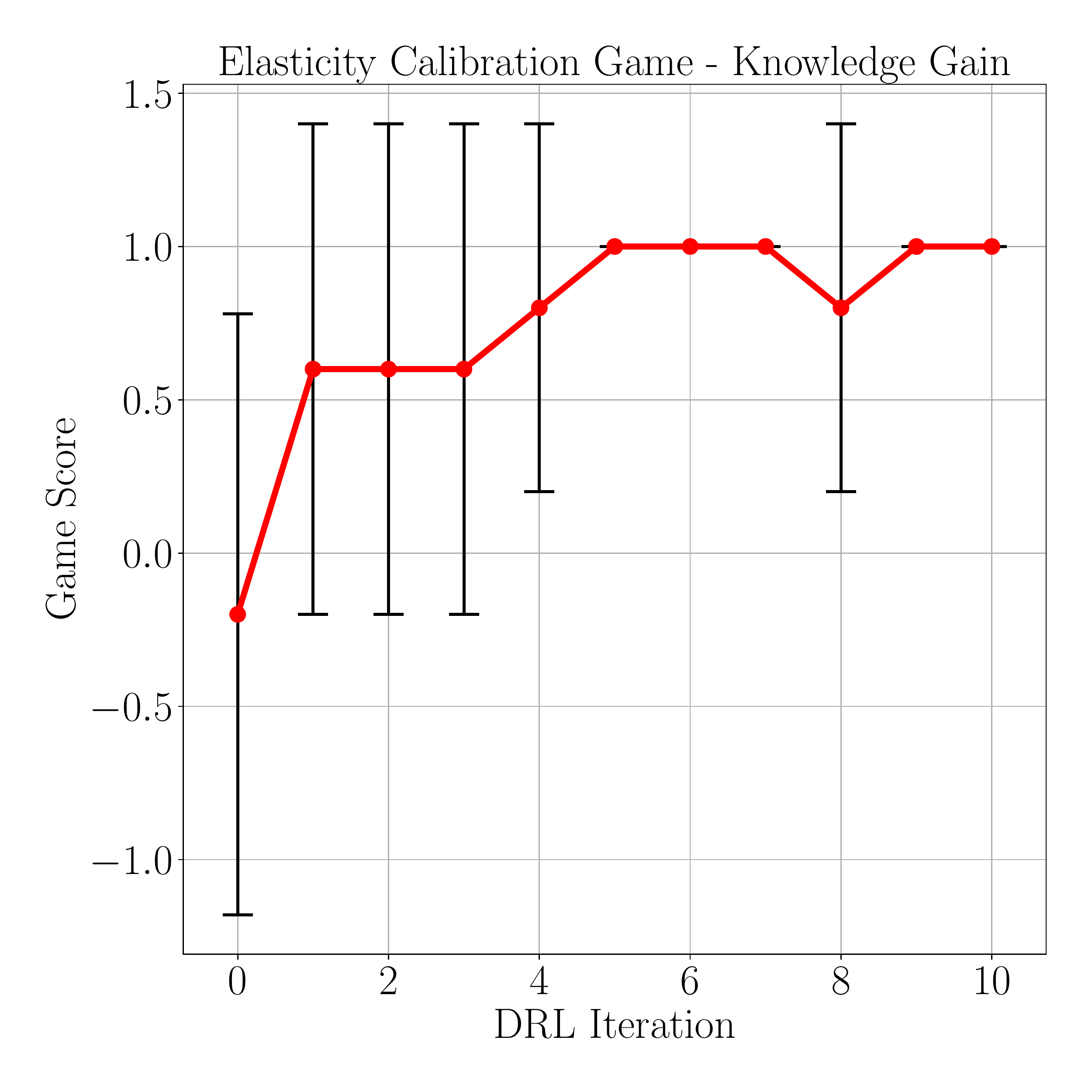}
    \caption{Convergence of a RL training experiment for the linear elastic calibration game with the information gain reward. The error bars indicate the standard deviation of the rewards over all episodes.}
    \label{fig:elasticity_convergence}
\end{figure}

The convergence of a training experiment is demonstrated in \fref{fig:elasticity_convergence}. 
The RL reward for the experiments is based on the information gain described in \eref{eq:knowledge_gain_reward}. 
The mean episode reward converges to a maximum value, with essentially zero standard deviation, of zero after 10 training iterations. 
Note that for this small decision tree, the rewards are scaled to be exactly 0 for the cases where the model is not calibrated successfully and 1 for the cases it is. 

\begin{figure}[ht!]
    \centering
    \includegraphics[width=0.95\textwidth]{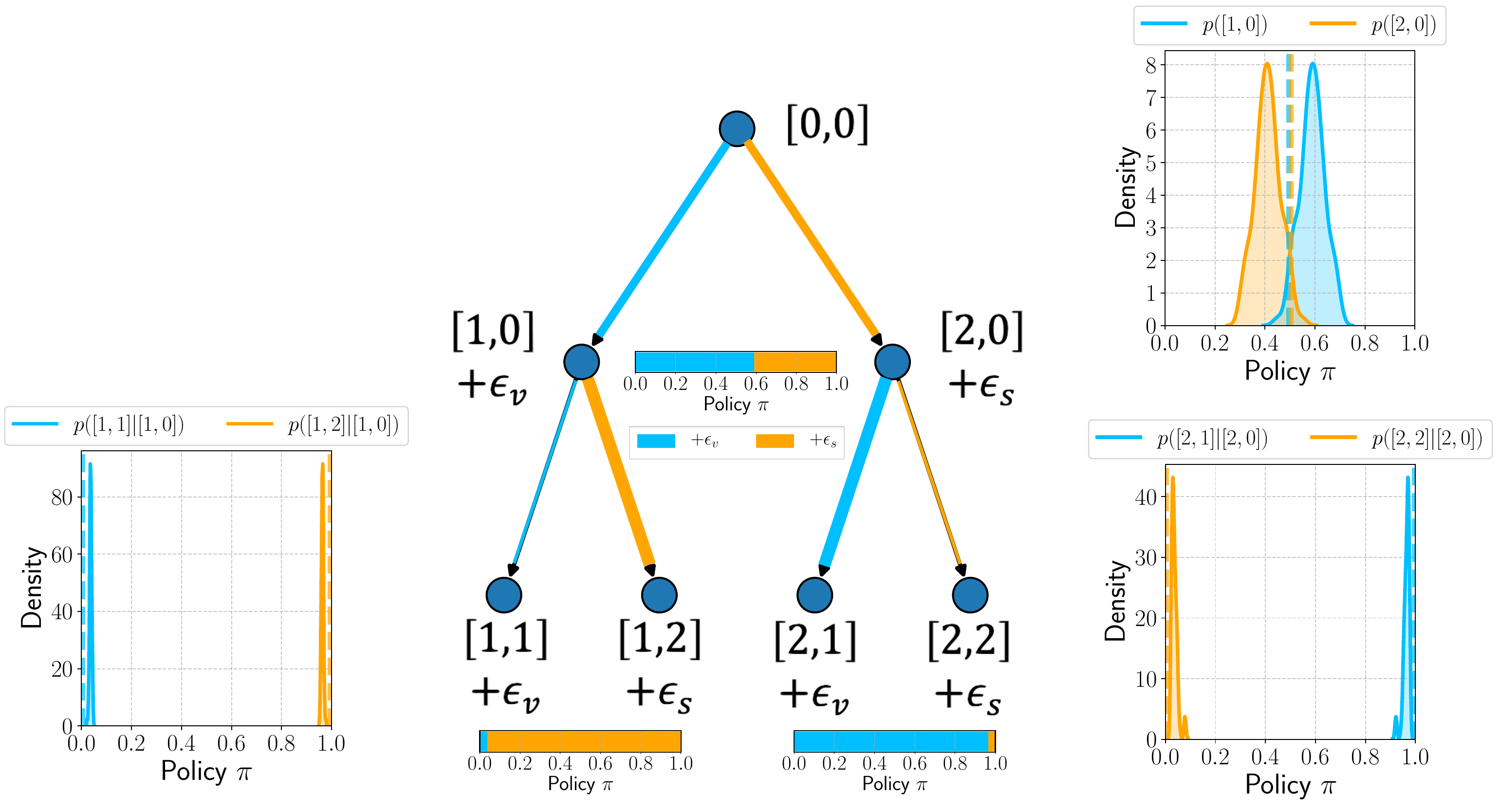}
    \caption{Linear elastic policy tree (two actions and two steps) for the information gain reward.
    }
    \label{fig:linearelastictree_policy}
\end{figure}

Since the RL algorithm applied to this problem is fast to converge and test, we also set up a numerical experiment to test the algorithm's repeatability and the effect the random initialization of the MCTS and neural network has on convergence.
Recall that this simple decision tree is composed of two actions; volumetric strain and shear strain were employed to generate the training data. 
\fref{fig:linearelastictree_policy} shows the distribution of converged policies $\policy$ for 100 trials to the expected policy, which should favor paths that take one of each of the independent actions. 
Upon the first step, there is little to distinguish between the value of taking a shear or compression step, but after the second step, the policy is essentially binary.
The final policy assigns the highest value to the shear-compression ([2,1][2,0]) and compression-shear ([1,2][1,0]) experiments; effectively zero value is assigned to experiments that take repeated steps because they cannot calibrate the linear elastic model. 
Note that the observed symmetry of tree policies was ensured by selecting a high value for the exploration variable in earlier iterations. 
For smaller values of the exploration parameter $c_{p u c t}$, the algorithm was still observed to converge but randomly biasing towards one winning path over the other.

\subsection{Implementation verification 2: experiment for identifying von Mises yield function and hardening}\label{sec:von_mises}

In this example, we introduce an experimental design problem with a history-dependent model and significantly larger total number of possible paths but also a known optimal design. 
In this numerical experiment, the EKF-DRL agent is tasked with determining elastic and plastic parameters given the prior knowledge that the yield surface is of von Mises type and isotropic, which is embedded in the chosen model.

\begin{figure}[htb!]
\centering
\begin{subfigure}[b]{0.35\textwidth} \centering
\includegraphics[width=1.0\textwidth]{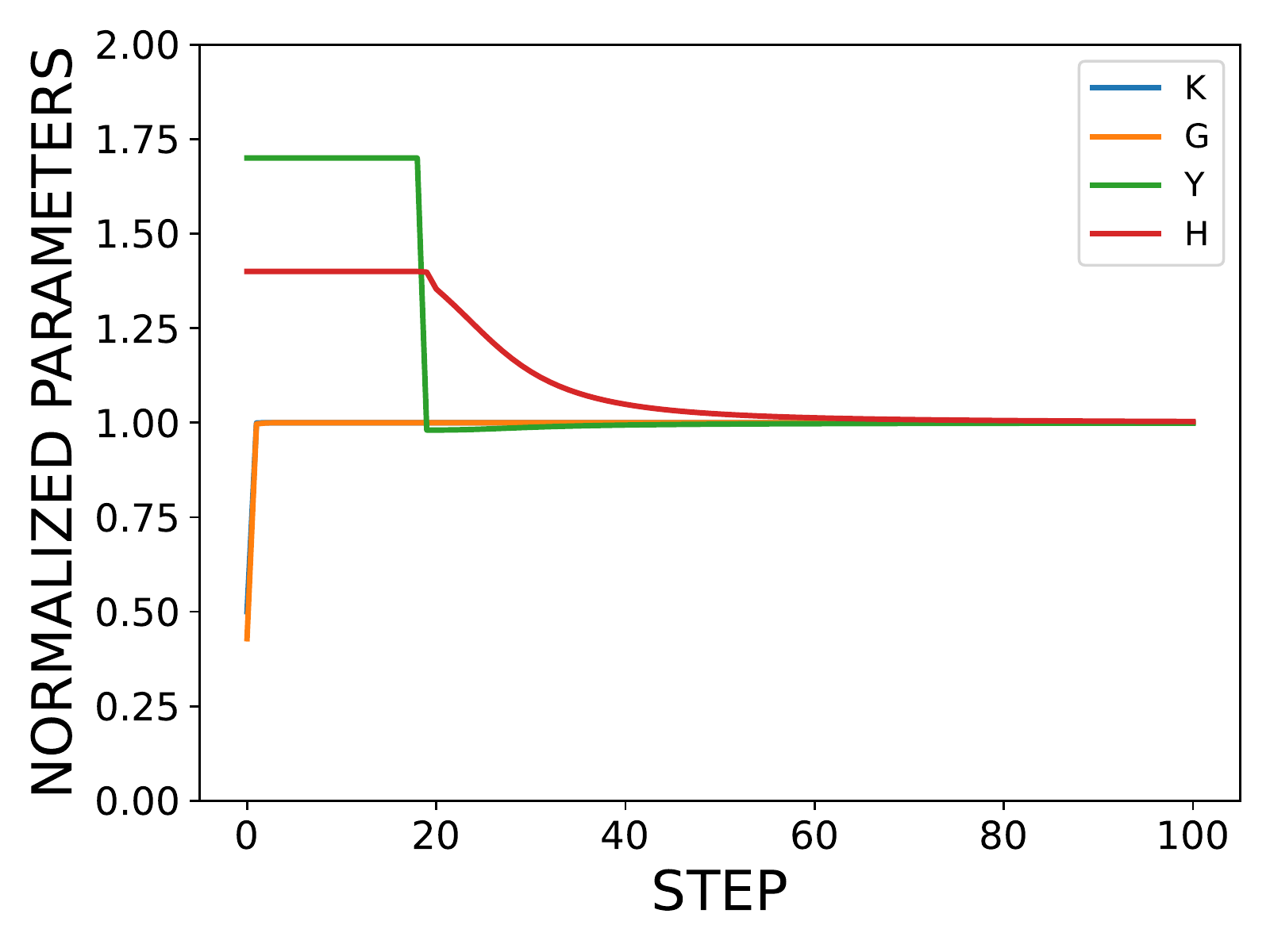}
\caption{masked calibration}
\end{subfigure}
\begin{subfigure}[b]{0.35\textwidth} \centering
\includegraphics[width=1.0\textwidth]{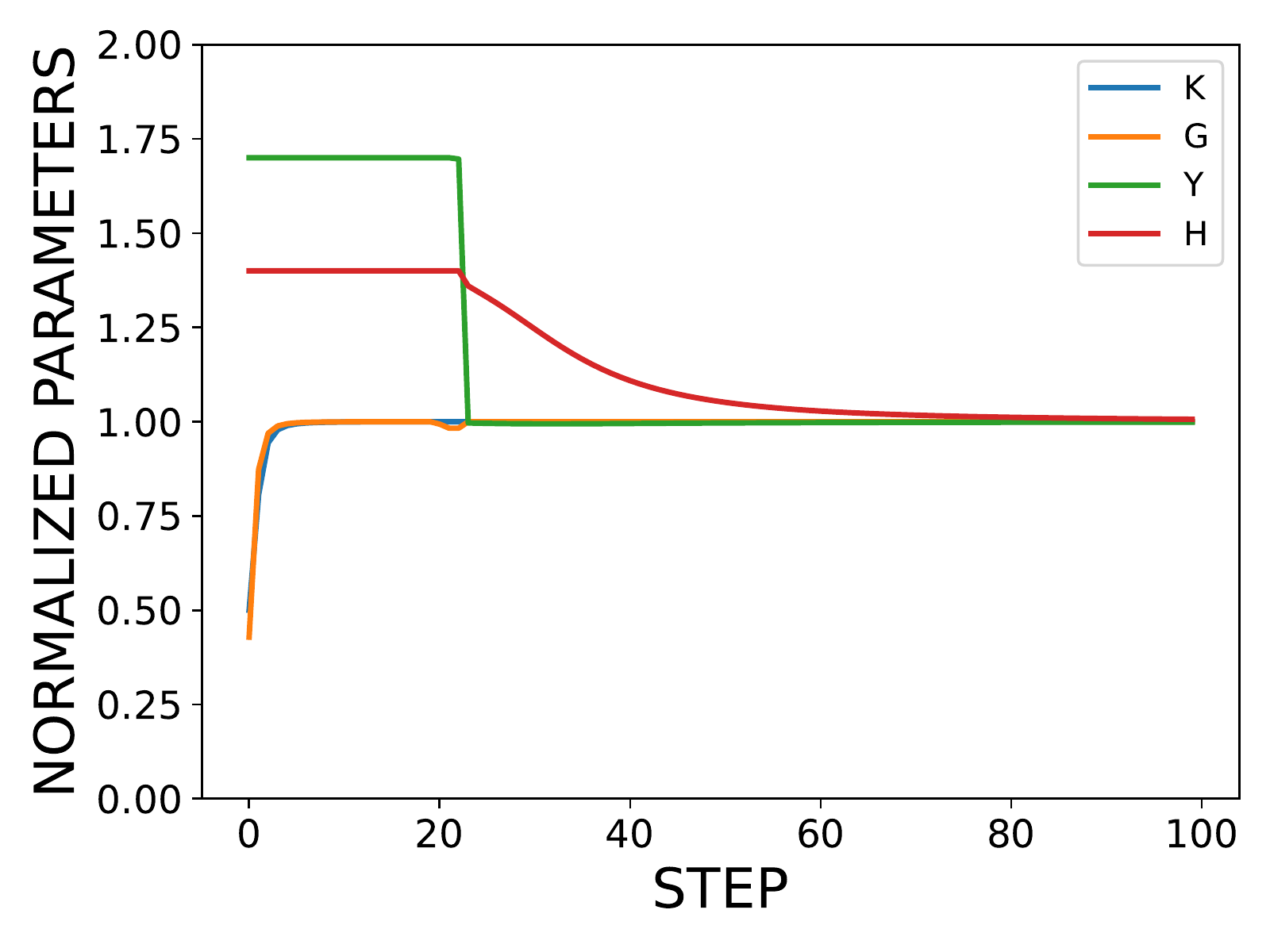}
\caption{switching calibration}
\end{subfigure}

\begin{subfigure}[b]{0.35\textwidth} \centering
\includegraphics[width=1.0\textwidth]{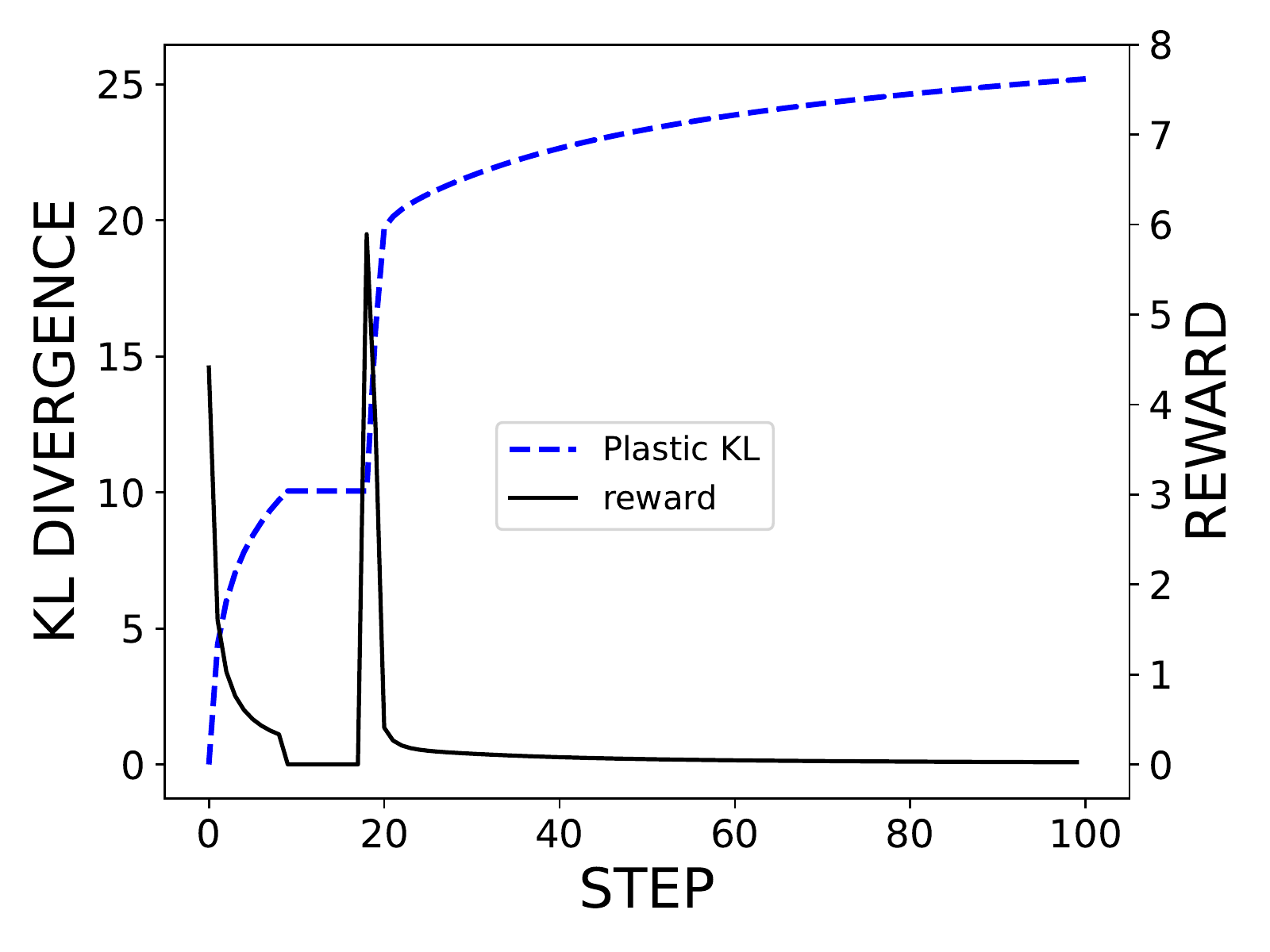}
\caption{masked reward}
\end{subfigure}
\begin{subfigure}[b]{0.35\textwidth} \centering
\includegraphics[width=1.0\textwidth]{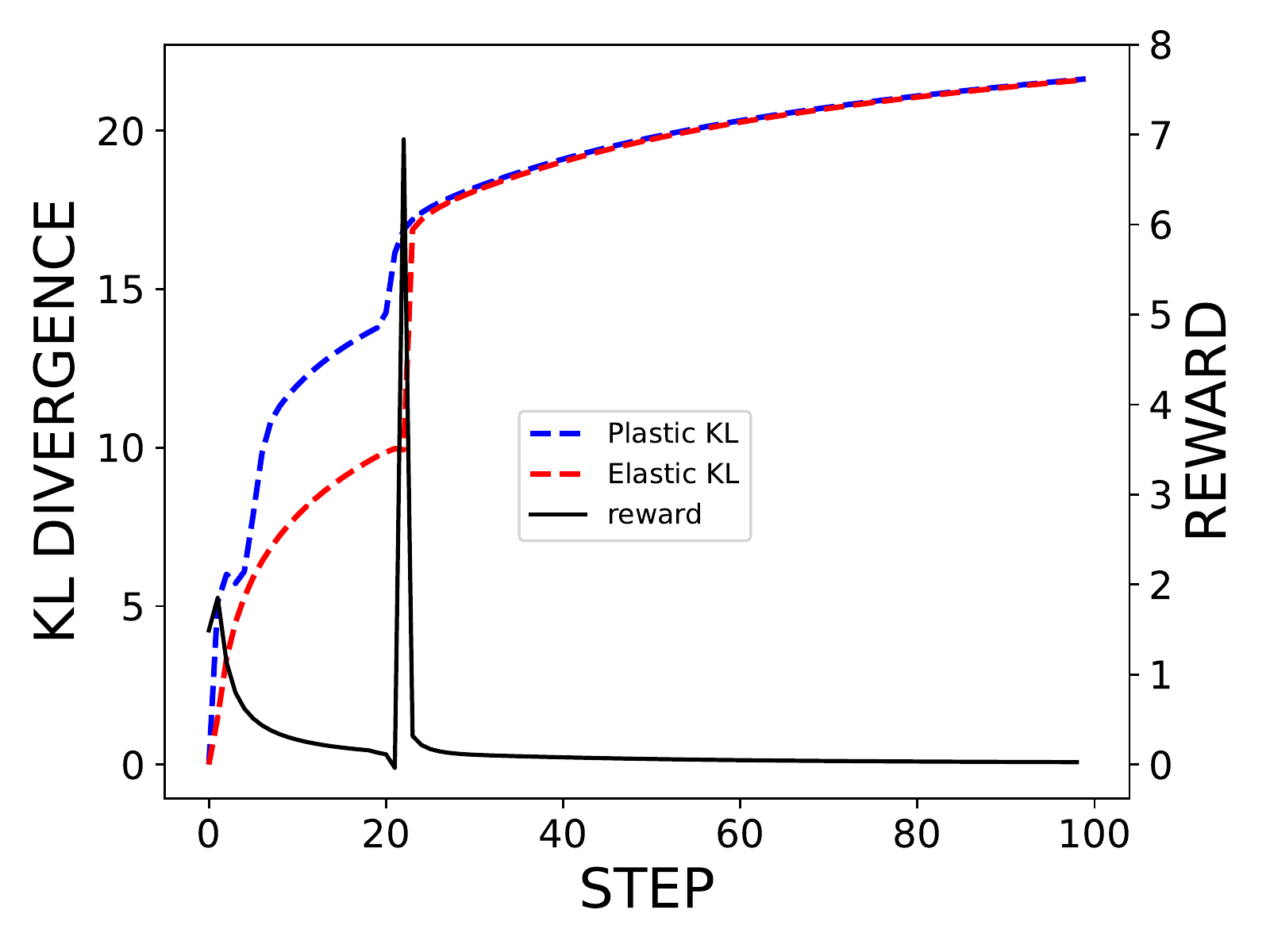}
\caption{switching reward}
\end{subfigure}

\begin{subfigure}[b]{0.35\textwidth} \centering
\includegraphics[width=1.0\textwidth]{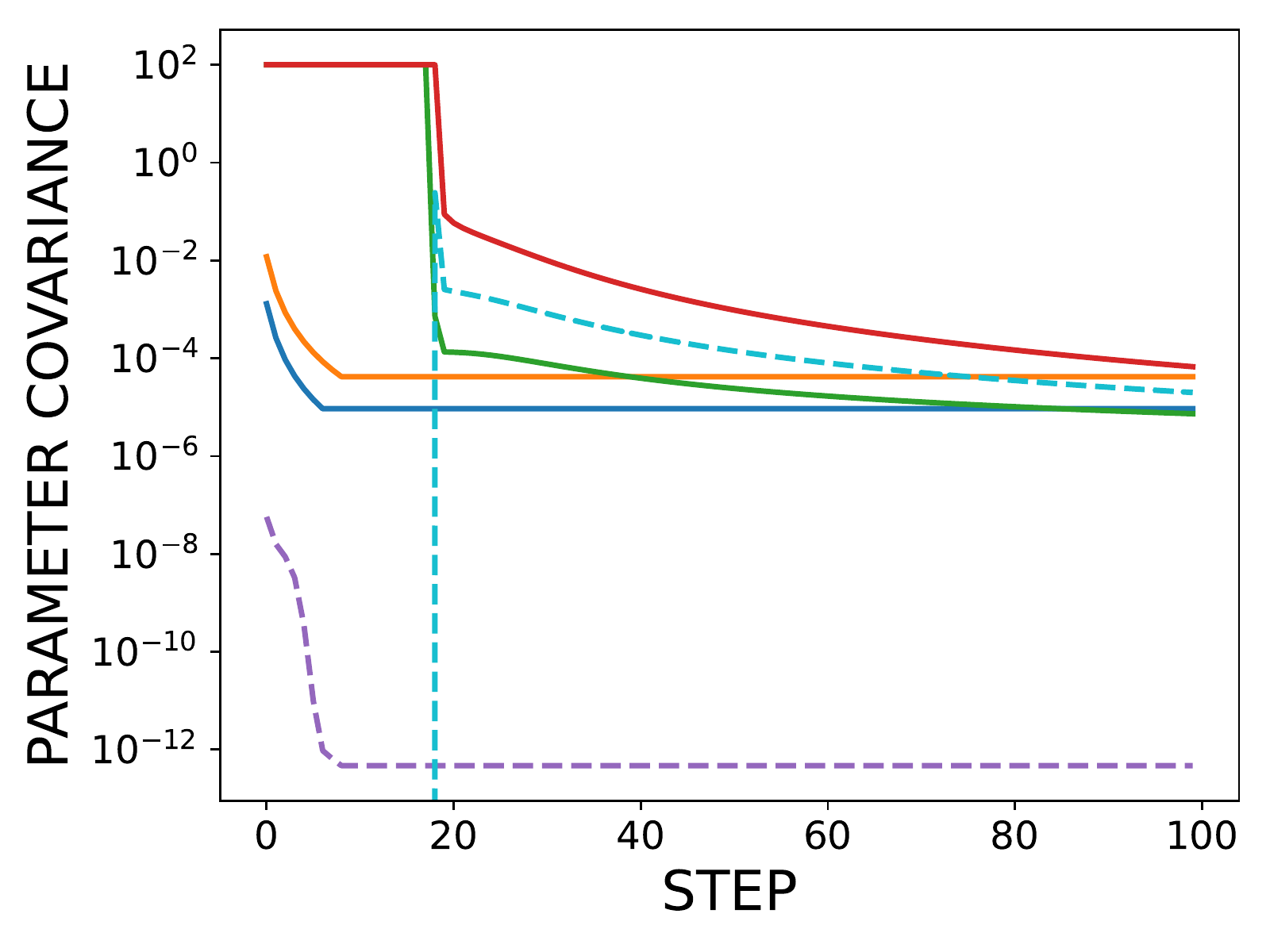}
\caption{masked covariance}
\end{subfigure}
\begin{subfigure}[b]{0.35\textwidth} \centering
\includegraphics[width=1.0\textwidth]{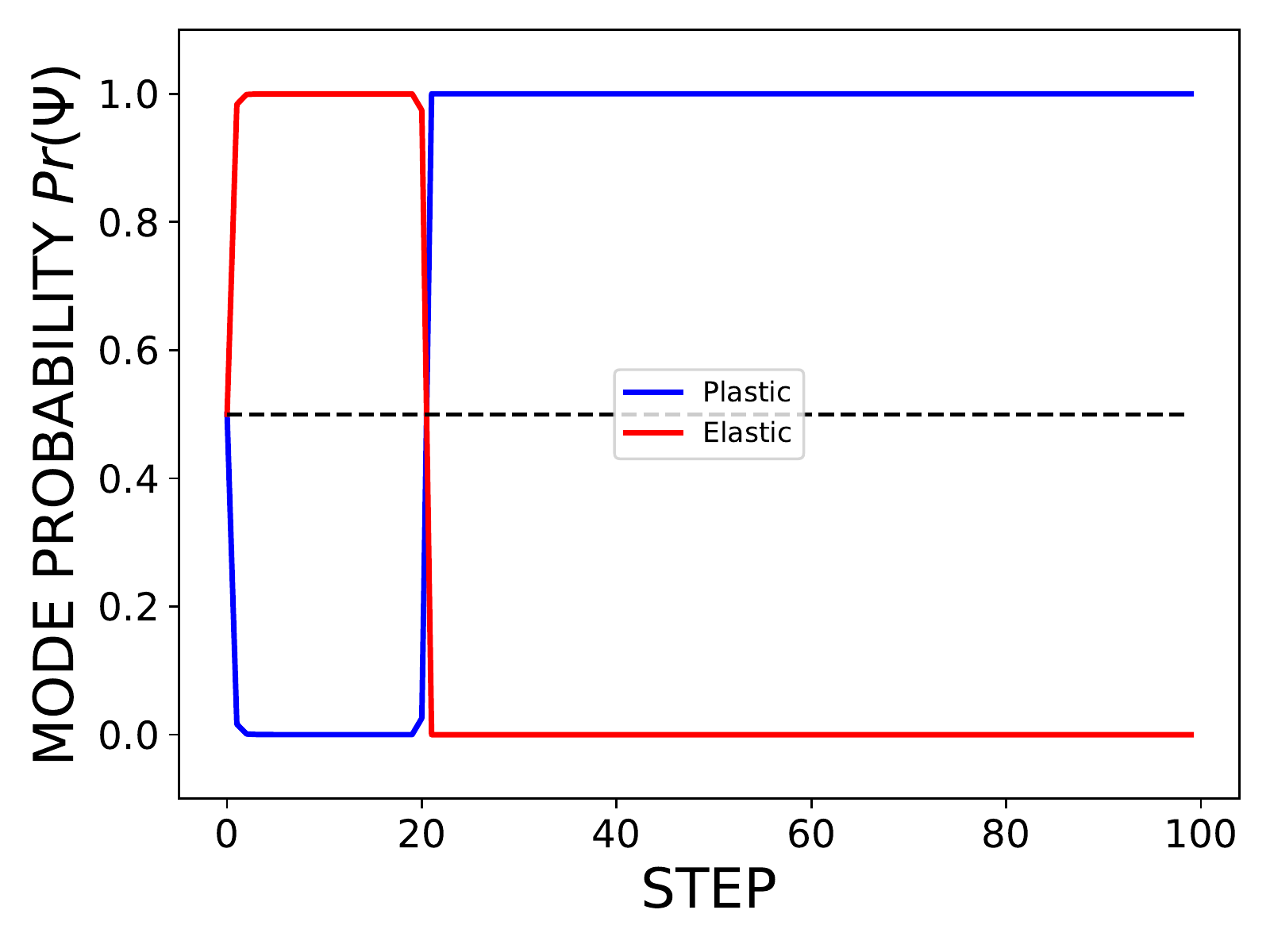}
\caption{switching probability}
\end{subfigure}

\caption{Comparison of the switching Kalman filter to the masking approach for a von Mises calibration. For (c) and (d), the dashed Kullback-Liebler Divergence lines represent the integral of the reward curve. In (d), the reward curve corresponds to the Plastic KL only since that is the relevant material model for experimental design.
}
\label{fig:J2_calibration}
\end{figure}

First we illustrate the performance of the EKF and compare the two variants of the switching algorithm.
\fref{fig:J2_calibration} illustrates the efficacy of the {\it ad hoc} masked Kalman filter and the more formally rigorous switching Kalman filter (SKF) described in \sref{sec:SKF}.
\fref{fig:J2_calibration}a,b show that the two methods both converge on the true parameters ($K$ bulk modulus, $G$ shear modulus, $Y$ yield strength, $H$ hardening modulus) with 100 steps; however, the convergence has different characteristics.
With the masked method, prior to yield the elastic parameters ($K$ bulk modulus, $G$ shear modulus) have converged and are fixed throughout the loading steps.
Furthermore, the plastic parameters ($Y$ yield strength, $H$ hardening modulus) are effectively fixed prior to encountering yield.
The abrupt change also affects the reward shown in \fref{fig:J2_calibration}c, and in particular, leads to the reward jumping when yield is encountered before leveling off again.
On the other hand, the SKF displays much smoother behavior in both the mean parameter convergence (\fref{fig:J2_calibration}b) and the reward (\fref{fig:J2_calibration}d). 
\fref{fig:J2_calibration}d shows the KL divergence for both elastic and plastic modes, but only the reward from the plastic mode is selected as the most probable reference of information gain. 
\fref{fig:J2_calibration}f demonstrates that the method switches effectively between the two models and chooses the elastic model in the elastic region and the plastic model post yield.
It should be noted that the convergence behavior is sensitive to the step size, i.e., a certain number of elastic samples are needed for convergence of the elastic parameters; yield is best detected over a moderate interval, and the estimate of the hardening improves at larger strains.
The covariance shown in \fref{fig:J2_calibration}e drives these changes. 
The relatively slow convergence of the hardening parameter $H$ for both methods is likely due to a lower sensitivity of the output to this parameter.

Next, we design the action-state space based on physical considerations.
Since the von Mises model is pressure-independent, we can constrain the experimental design search to the $\pi$-plane, i.e., the plane perpendicular to the pressure and passing through the origin of the three principal stress axes.
The underlying elasticity model is still linear elastic. 
Thus, we can constrain the exploration of strain and stress on the $\pi$-plane by designing strain increment decisions in the two principal directions $\pm\Delta\epsilon_1$ and $\pm\Delta\epsilon_2$ and the increment in the third direction being calculated to have a volumetric strain increment equal to 0.

\begin{figure}[ht!]
    \centering
    \includegraphics[width=0.99\textwidth]{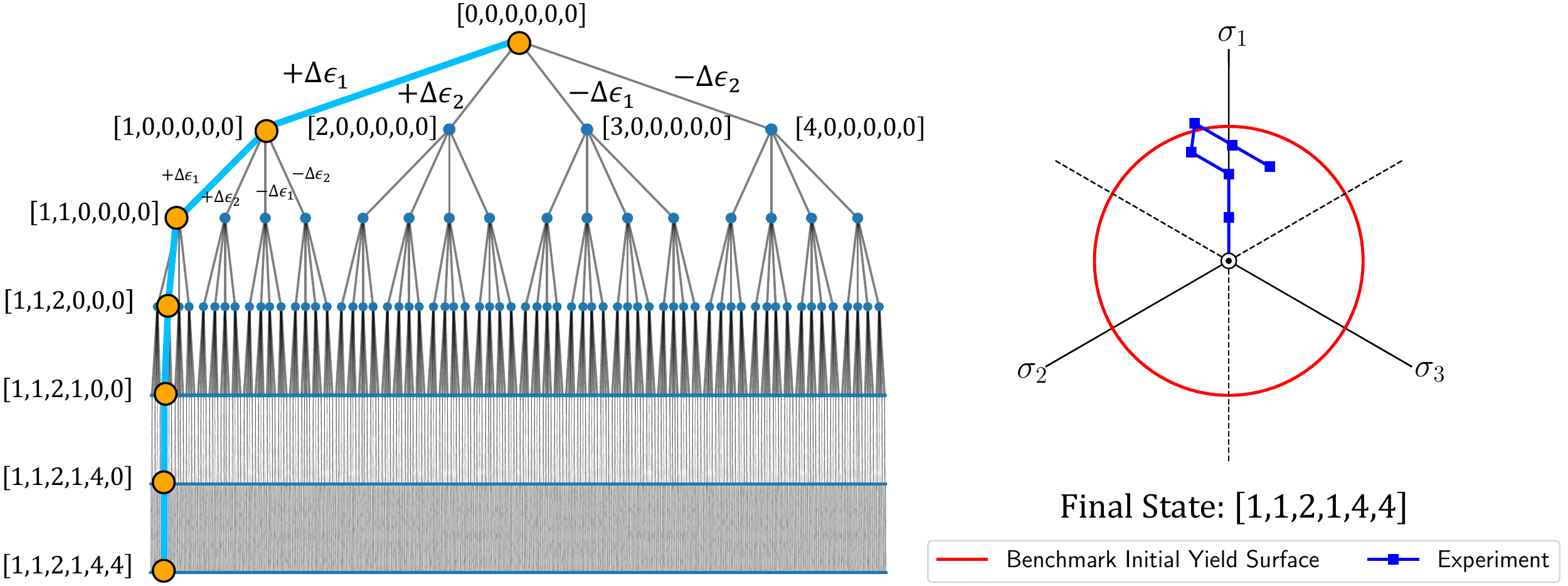}
    \caption{Decision tree for the exploration of the principal stress space on the $\pi$-plane. A complete path from the root note to a leaf node is shown along with the corresponding experiment stress path on the $\pi$-plane.}
    \label{fig:plasticity_tree}
\end{figure}

The decision tree that describes this process is shown in \fref{fig:plasticity_tree}. 
From every state in the decision tree, there can be four allowed actions: increase $+\Delta\epsilon_1$, increase $+\Delta\epsilon_2$, decrease $-\Delta\epsilon_1$ or decrease $-\Delta\epsilon_2$. 
The magnitude of the strain increment is chosen to be $\Delta\epsilon = 0.04$ in each direction. 
The number of options/layers in the decision tree is $n_\text{opt} = 6$ which was deemed enough to explore strain-stress cases that exceed the yielding point and demonstrate hardening behavior. 
The state vector has a length of $n_\text{opt} = 6$ as each component corresponds to a selected action. 
In the root state of the tree, all the components are equal to 0. 
An enumeration is used for all the actions.
For example, selecting the action to increase $+\Delta\epsilon_1$, the component corresponding to this action would be 1, selecting $+\Delta\epsilon_2$ it would be 2, and so on. 
In \fref{fig:plasticity_tree}, a final state that corresponds to a leaf node in the decision tree is also shown along with the corresponding stress path on the $\pi$-plane. 
The number of all possible configurations/states in the tree is 5461, while the number of final states/experiments is 4096.

Thus, we can define the RL algorithm environment to make experimental decisions (decision tree), generate experimental data (a von Mises benchmark model in this synthetic experiment) and calibrate the plasticity model (another von Mises model). 
In this RL benchmark experiment, we train a RL neural network to explore the $\pi$-plane stress space to design experiments that optimize the Kalman filter's discovery of the plastic parameters: yield stress $Y$ and hardening modulus $H$. 
Here, the underlying elastic model parameters are considered known ($K=1.0$ , $G= 0.7$). The RL algorithm was performed for 20 training iterations. Each iteration has 10 game episodes. 
Each game episode involves traversing through the decision tree, root to a leaf node, to design an experiment, collect the training data for this experiment, calibrate the KF model and calculate a reward for this episode. 
The exploration parameter was again set to a high value ($c_{p u c t}=10$) in the first iteration and linearly reduce to 1 in the final iteration to encourage exploration of the decision tree and to avoid converging to a local maximum of the reward. 

\begin{figure}[ht!]
    \centering
    \includegraphics[width=0.85\textwidth]{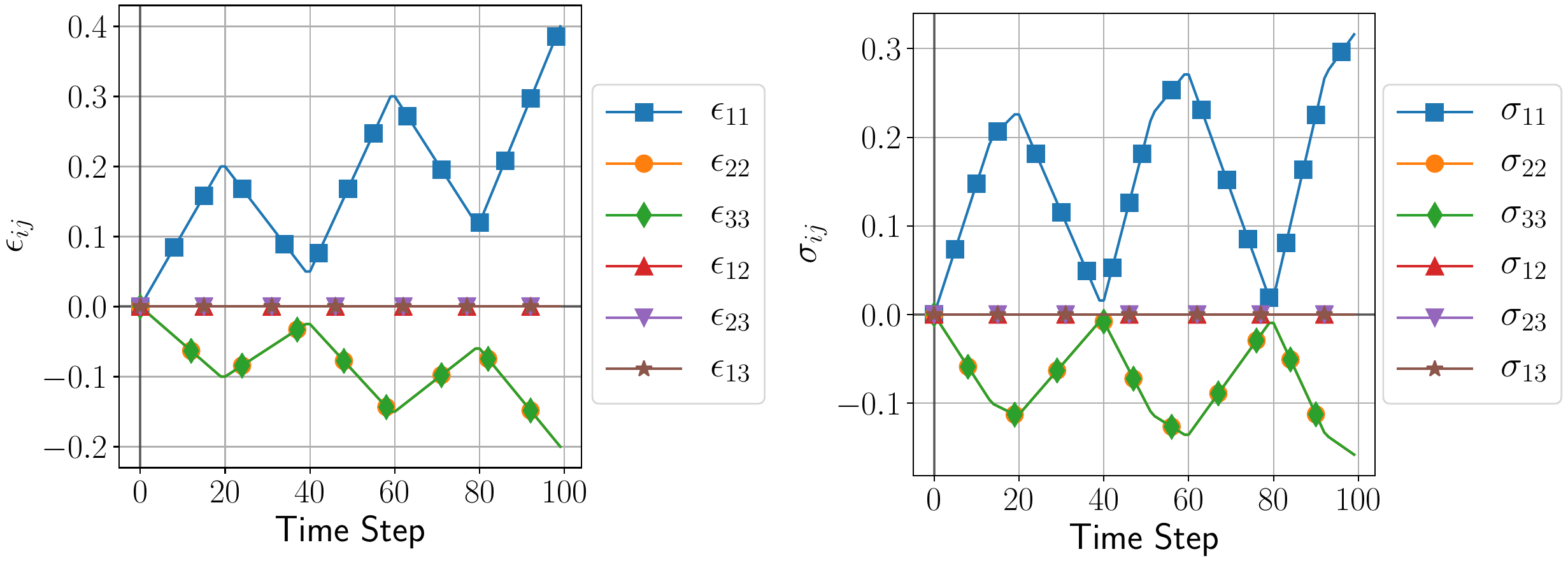}
    \caption{The strain and stress path of the blind test experiment used to calculate the efficiency index reward for the von Mises dataset.}
    \label{fig:j2_blind_test}
\end{figure}

\begin{figure}[ht!]
    \centering
    \includegraphics[width=0.95\textwidth]{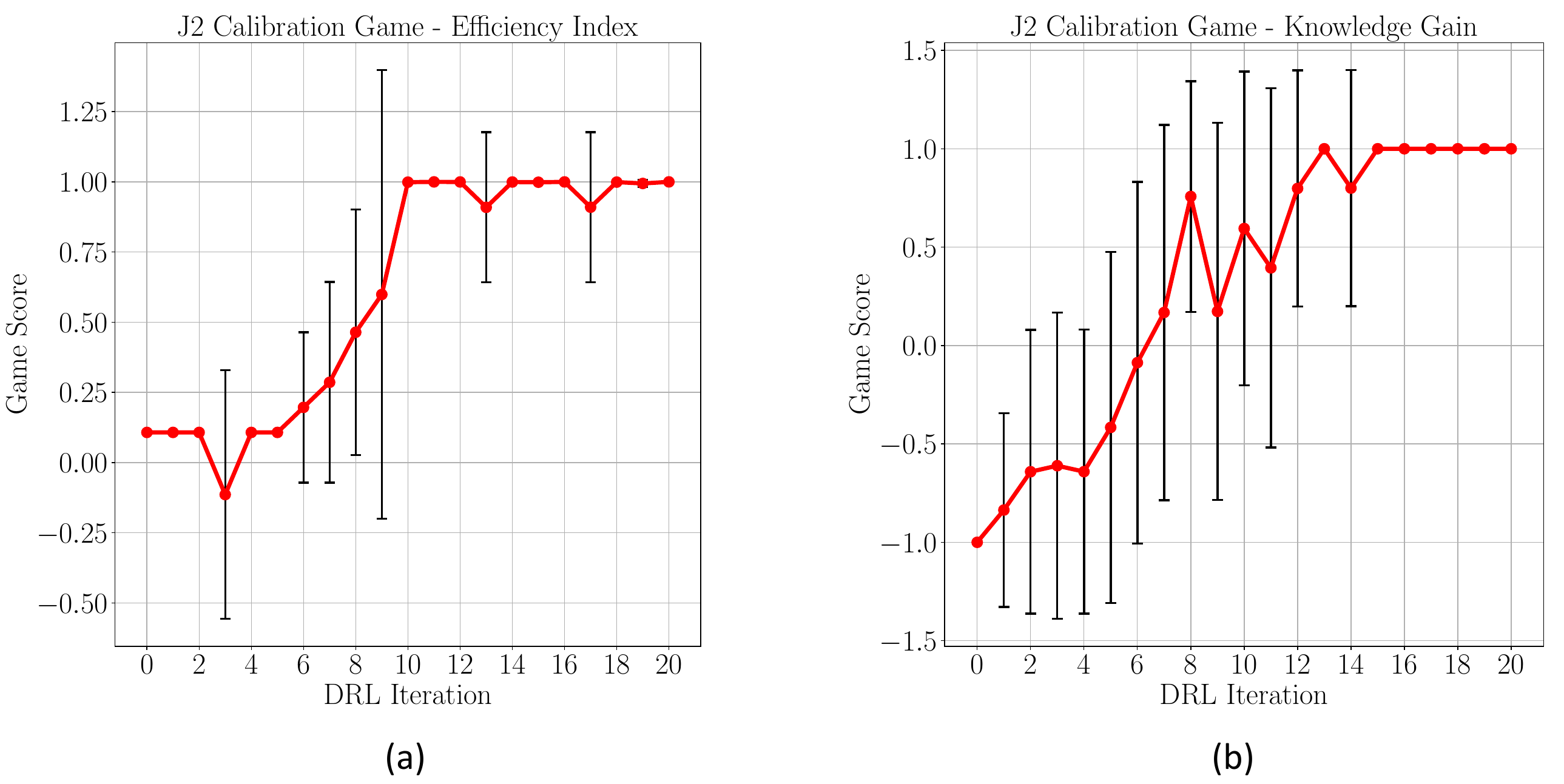}
    \caption{Game score vs. iteration for the von Mises plasticity calibration game calibrating with the Kalman filter model and using (a) an efficiency index reward (NSE) and (b) an information gain reward (KF/KL). The red dots and the error bars are the mean and $\pm 1$ standard derivation over the episodes.}
    \label{fig:j2_game_convergence}
\end{figure}

\begin{figure}[ht!]
    \centering
    \includegraphics[width=0.85\textwidth]{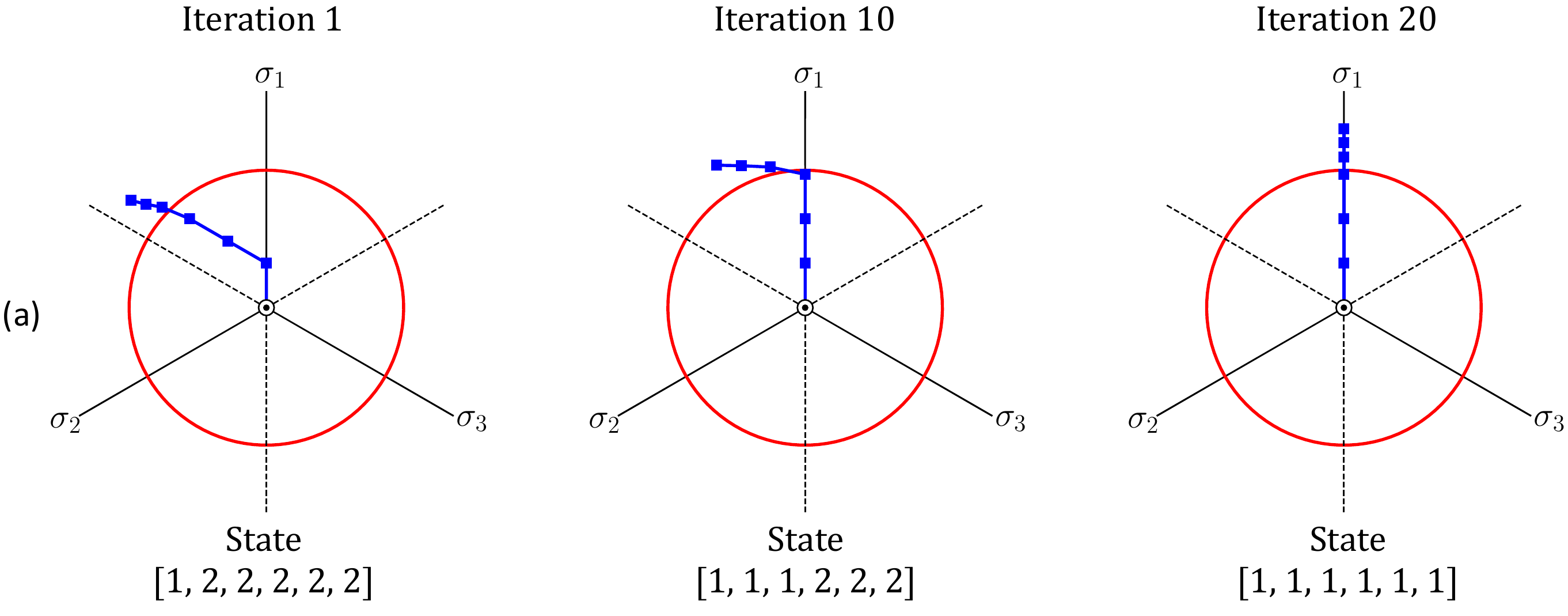}
    \includegraphics[width=0.85\textwidth]{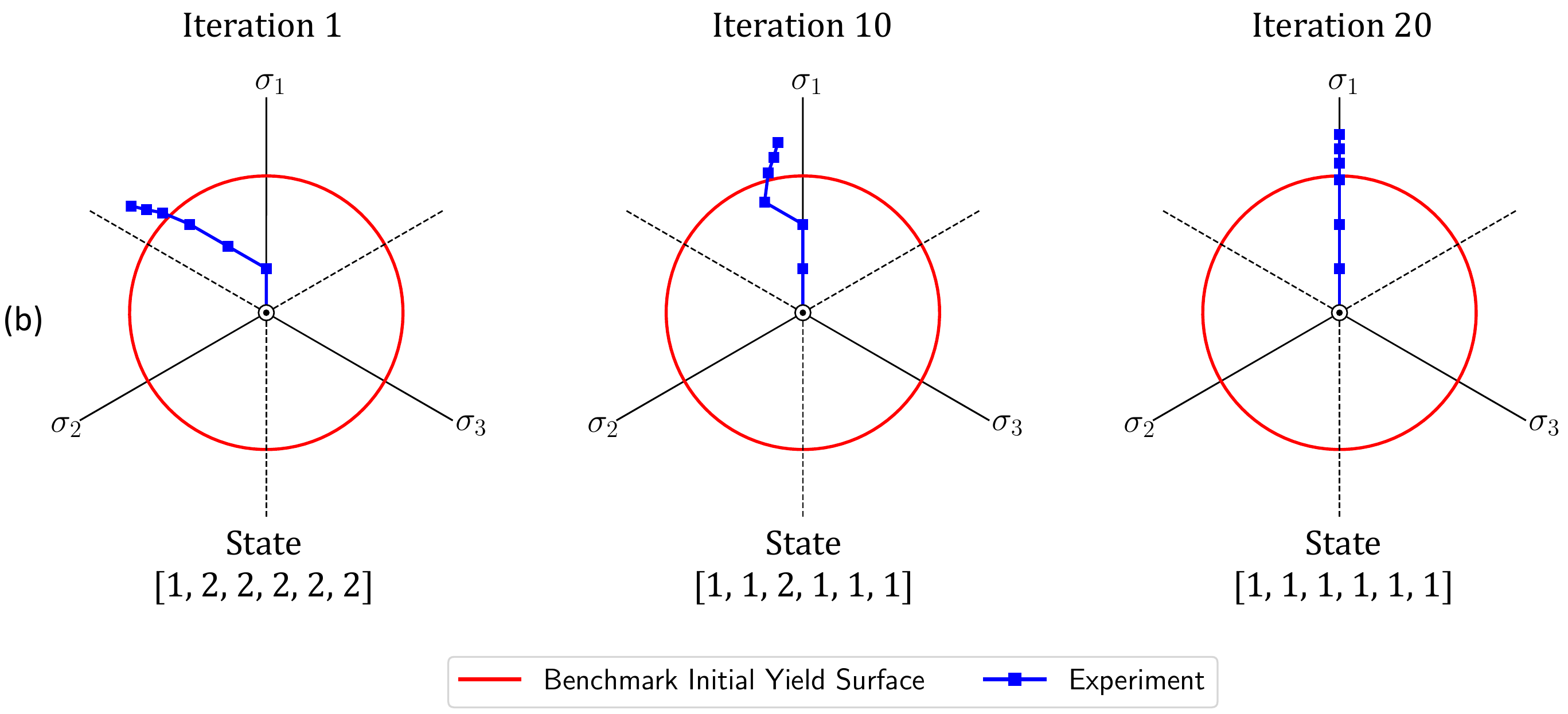}
    \caption{Designed experiments for the von Mises plasticity calibration game at the end of iterations 1, 10 and 20, calibrating with the Kalman filter model and using (a) an efficiency index reward and (b) an information gain reward.}
    \label{fig:j2_game_iterations}
\end{figure}

The neural network architecture used in this experiment is based on the architecture described in \sref{sec:policy}. Similar to elasticity example, the network inputs a six-component state vector, has two hidden layers (100 neurons each and $\operatorname{ReLU}$ activations), and two output layers: the policy output (six neurons and $\operatorname{softmax}$ activation) and value output (one neuron and $\operatorname{tanh}$ activation). 
As with the previous example, the kernel weight matrix of every layer was initialized with a Glorot uniform distribution and the bias vector with a zero distribution.
The model architecture and optimized weights from the previous iteration are reloaded and trained for 500 epochs at the end of every iteration with a batch sample size of 32 using the Adam optimizer, set with default values.

The RL experiment was performed once with an efficiency index reward based on \eref{eq:NSE} as described in \sref{sec:efficiency_index} and once with an information gain reward based on \eref{eq:knowledge_gain_reward} as described in \sref{sec:kl}.
The efficiency index reward requires an additional sampling of test data points to be calculated for every episode. The additional test curve selected is shown in \fref{fig:j2_blind_test}. 
We note that the information gain reward does not require additional sampling of the material response space, and it is calculated on the collected experimental data after calibration is complete. 
The convergence of these two numerical experiments is showcased in \fref{fig:j2_game_convergence}, showing the mean episode reward and the episode reward standard deviation for every game iteration. 
The effect of a higher exploration parameter and the random initialization of the RL neural network in the first game iterations are visible as the strain paths are sampled randomly leading to correspondingly poor mean rewards and high standard deviations. 
The RL network begins to encourage the sampling of experiments that provide increasingly higher rewards for both the efficiency index and information gain reward before converging to a maximum score at the last few iterations of playing.

The behavior of this convergence is reflected in the optimal experiment predicted by the RL neural network at the end of every game iteration.
In \fref{fig:j2_game_iterations}, we demonstrate the designed experiments at the end of iterations 1, 10 and 20. 
\fref{fig:j2_game_iterations}(a) and \fref{fig:j2_game_iterations}(b) show the designed experiments for the efficiency index and information gain rewards, respectively.
These paths are selected by making a forward prediction with the trained RL neural network at the end of each training iteration. 
We traverse through the decision tree starting from the root node $[0,0,0,0,0,0]$ and selecting the action with the highest policy/probability as predicted by the RL network. 
By the last iteration, the networks have converged to predict the final state $[1,1,1,1,1,1]$ that corresponds to a monotonically increasing strain-stress curve in the direction of the first principal stress axis.
This radial loading choice is expected as the yield surface model is isotropic.
Furthermore, as previously mentioned, the model sensitivities to moduli, such as $H$, increase with increased strain, so the radial directions tend to be the most informative.
Thus, the KF optimizes the calibration/maximizes the information gain when the data density in one direction is maximized. 
The converged final states and calibrated KF parameters are shown in Table~\ref{tab:J2_params}.

\begin{table}[ht!]
\centering
\begin{tabular}{|c|c|}
\hline
\textbf{Final State}     & \textbf{Experimental Design}                                                                                  \\ \hline
$[1,1,1,1,1,1]$               & $+\Delta\epsilon_1,+\Delta\epsilon_1,+\Delta\epsilon_1,+\Delta\epsilon_1,+\Delta\epsilon_1,+\Delta\epsilon_1$ \\ \hline
\textbf{Benchmark Parameters} & \textbf{Calibrated Parameters}                                                                                \\ \hline
$Y_0 = 0.3$                   & $Y_0 =0.30261$                                                                                                \\ \hline
$H = 1.0$                     & $H = 0.9578$                                                                                                  \\ \hline
\end{tabular}
    \caption{The RL-discovered final state and corresponding experimental design for the von Mises calibration game along with the benchmark and KF-calibrated plasticity parameters.}
    \label{tab:J2_params}
\end{table}

\subsection{Anisotropic plasticity}

In the last numerical experiment, we demonstrate the capacity of the KF model and RL algorithm to calibrate the anisotropic modified Hill model described in \sref{sec:exemplar}. 
In this numerical experiment, we are setting up the RL environment to calibrate both the elastic parameters ($E, \nu, \nu_\perp$) and plastic parameters ($B, Y_0, H$).
Given these complexities the ideal path is not known {\it a priori}.

\begin{figure}[ht!]
    \centering
    \includegraphics[width=0.95\textwidth]{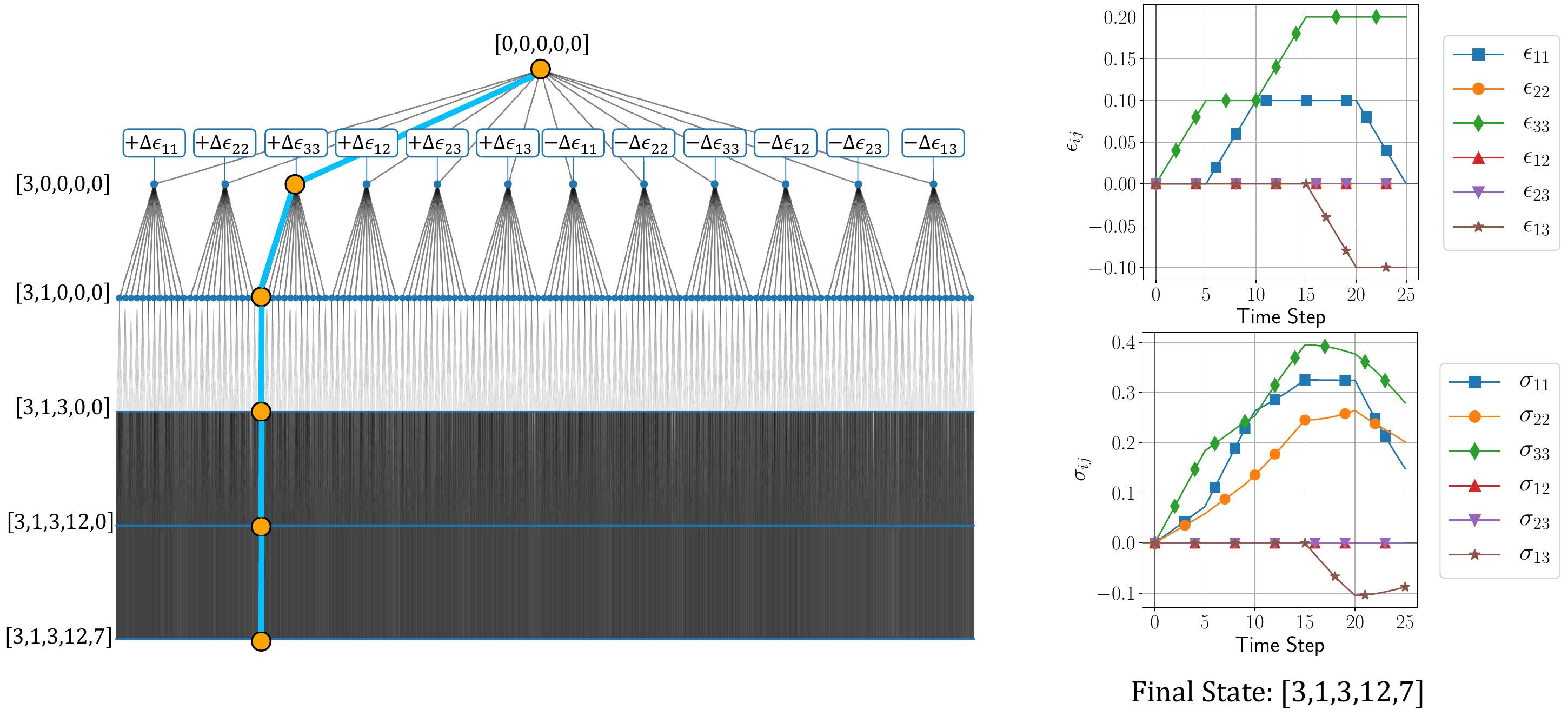}
    \caption{Decision tree for the exploration of the strain space. A complete path from the root note to a leaf node is shown along with the corresponding experiment strain-stress paths for the Hill model ($B= 0.5$).}
    \label{fig:hill_tree}
\end{figure}

For this material model, the $\pi$-plane is not enough to adequately describe the anisotropy of the data; thus, we opted for full control of the general strain-stress space. 
In \fref{fig:hill_tree}, the decision tree for the exploration of this parametric space is illustrated.
From every state in the decision tree, there are 12 allowable actions to select from.
Every choice corresponds to either increasing or decreasing of one of the six symmetric strain tensor components $\epsilon_{ij}$.
The number of options in the decision tree is set to be $n_\text{opt} = 5$, which is also the length of the state and policy vectors.
As a result, the initial state/root node of the decision tree corresponds to 
the zero vector of $[0,0,0,0,0]$. Selecting to increase $+\Delta\epsilon_{33}$ would correspond to state $[3,0,0,0,0]$, then increasing $+\Delta\epsilon_{11}$ would correspond to state $[3,1,0,0,0]$, and so on. 
In \fref{fig:hill_tree}, we also illustrate a final leaf state and the corresponding training strain-stress paths that will be used to calibrate the KF model. 
The number of all possible configurations/states in the tree is 271,453, while the number of final states/experiments is 248,832.

\begin{table}[ht!]
\centering
\begin{tabular}{cc}
\hline
\multicolumn{2}{|c|}{\textbf{Modified Hill model (B = 0.5)}}                                                                                                                \\ \hline
\multicolumn{1}{|c|}{\textbf{Final State}}          & \multicolumn{1}{c|}{\textbf{Experimental Design}}                                                                     \\ \hline
\multicolumn{1}{|c|}{$[1,1,4,1,1]$}                 & \multicolumn{1}{c|}{$+\Delta\epsilon_{11},+\Delta\epsilon_{11},+\Delta\epsilon_{12},+\Delta\epsilon_{11},+\Delta\epsilon_{11}$} \\ \hline
\multicolumn{1}{|c|}{\textbf{Benchmark Parameters}} & \multicolumn{1}{c|}{\textbf{Calibrated Parameters}}                                                                   \\ \hline
\multicolumn{1}{|c|}{$E = 1.5$}                     & \multicolumn{1}{c|}{$E = 1.5013$}                                                                                     \\ \hline
\multicolumn{1}{|c|}{$\nu = 0.3$}                   & \multicolumn{1}{c|}{$\nu = 0.2992$}                                                                                   \\ \hline
\multicolumn{1}{|c|}{$\nu_\perp =0.2 $}             & \multicolumn{1}{c|}{$\nu_\perp = 0.2014 $}                                                                            \\ \hline
\multicolumn{1}{|c|}{$B=0.5$}                       & \multicolumn{1}{c|}{$B=0.4996$}                                                                                       \\ \hline
\multicolumn{1}{|c|}{$Y_0 = 0.1$}                   & \multicolumn{1}{c|}{$Y_0 = 0.9999$}                                                                                   \\ \hline
\multicolumn{1}{|c|}{$H = 0.1$}                     & \multicolumn{1}{c|}{$H = 0.1001$}                                                                                     \\ \hline
                                                    &                                                                                                                       \\ \hline
\multicolumn{2}{|c|}{\textbf{Modified Hill model (B = 2.0)}}                                                                                                                \\ \hline
\multicolumn{1}{|c|}{\textbf{Final State}}          & \multicolumn{1}{c|}{\textbf{Experimental Design}}                                                                     \\ \hline
\multicolumn{1}{|c|}{$[1,1,1,4,1]$}                 & \multicolumn{1}{c|}{$+\Delta\epsilon_{11},+\Delta\epsilon_{11},+\Delta\epsilon_{11},+\Delta\epsilon_{12},+\Delta\epsilon_{11}$} \\ \hline
\multicolumn{1}{|c|}{\textbf{Benchmark Parameters}} & \multicolumn{1}{c|}{\textbf{Calibrated Parameters}}                                                                   \\ \hline
\multicolumn{1}{|c|}{$E = 1.5$}                     & \multicolumn{1}{c|}{$E = 1.4893$}                                                                                     \\ \hline
\multicolumn{1}{|c|}{$\nu = 0.3$}                   & \multicolumn{1}{c|}{$\nu = 0.2984$}                                                                                   \\ \hline
\multicolumn{1}{|c|}{$\nu_\perp =0.2 $}             & \multicolumn{1}{c|}{$\nu_\perp = 0.2109 $}                                                                            \\ \hline
\multicolumn{1}{|c|}{$B=2.0$}                       & \multicolumn{1}{c|}{$B=1.9889$}                                                                                       \\ \hline
\multicolumn{1}{|c|}{$Y_0 = 0.15$}                  & \multicolumn{1}{c|}{$Y_0 = 0.1505$}                                                                                   \\ \hline
\multicolumn{1}{|c|}{$H = 0.2$}                     & \multicolumn{1}{c|}{$H = 0.1999$}                                                                                     \\ \hline
\end{tabular}
    \caption{The RL-discovered final state and corresponding experimental design for the Hill model calibration game along with the benchmark and KF-calibrated parameters for $B=0.5$ and $2$.}
    \label{tab:hill_params}
\end{table}

We can thus define the environment to design the experiments (decision tree), gather the experimental data and calibrate an anisotropic elastoplasticity model (KF). 
In this section, we conduct three numerical experiments to observe the capacity of the network to collect experimental data to fully calibrate the KF model for different anisotropic parameters in the dataset.

\begin{figure}[ht!]
    \centering
    \includegraphics[width=0.85\textwidth]{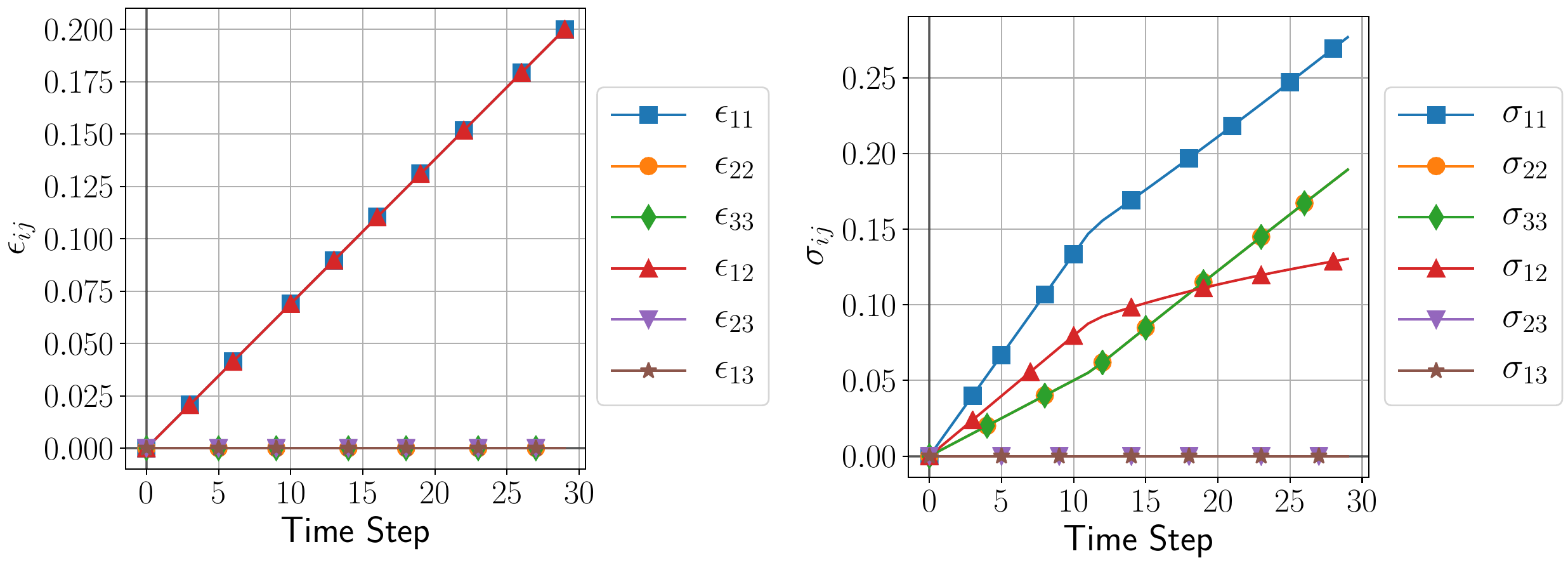}
    \caption{The strain and stress path of the blind test experiment used to calculate the efficiency index reward for the $B=0.5$ modified Hill model dataset.}
    \label{fig:hill_blind_test}
\end{figure}

For the first two experiments, the RL algorithm is tasked with designing an experiment and calibrating the model for yield surface data with different degrees of anisotropy, $B=0.5$ and $2$. 
A full description of the parameters of the dataset is shown in Table~\ref{tab:hill_params}. 
The algorithm was performed for 30 training iterations. 
In each iteration, there are 10 game episodes, each involves designing an experiment, calibrating the KF model and calculating the episode reward. 
In these experiments, the mixed reward is calculated through \eref{eq:mixed_reward} where the normalized efficiency index and information gain rewards of \eref{eq:NSE} and \eref{eq:knowledge_gain_reward}, respectively, are weighted by $w_\text{NSE}= w_\text{KL} = 0.5$. 
The efficiency index is calculated against a blind test experiment strain-stress curve illustrated in \fref{fig:hill_blind_test}.
The exploration parameter $c_{p u c t}$ was tuned to 5 in the first iteration and linearly reduce to 1 in the last iteration. 
The RL neural network architecture, hyperparameters and training procedures are identical to the ones described in \sref{sec:von_mises}.

\begin{figure}[ht!]
    \centering
    \includegraphics[width=0.85\textwidth]{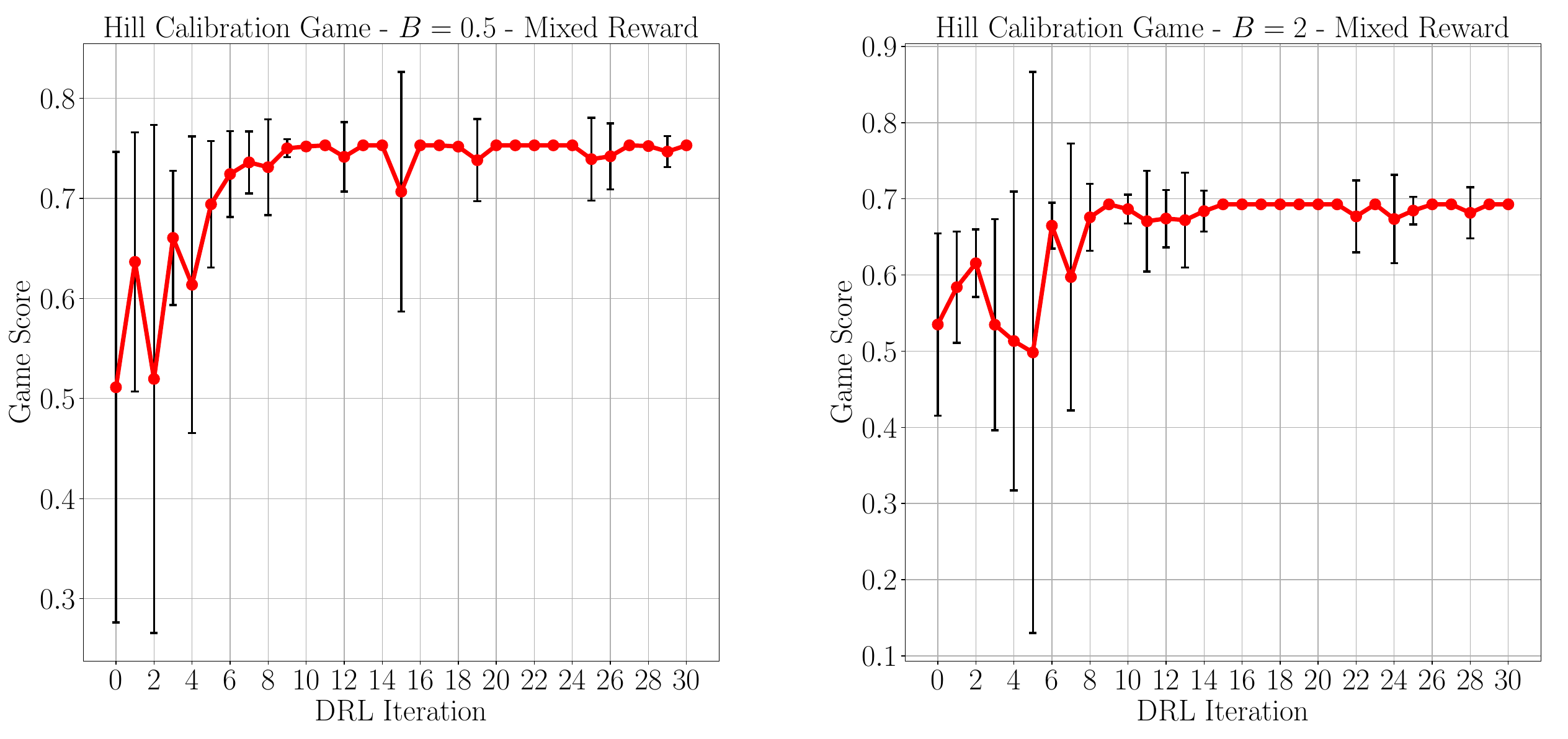}
    \caption{Game score vs. iteration for the Hill plasticity calibration game for $B=0.5$ and $2$ calibrating with the KF model and using a mixed reward combining the efficiency index and information gain reward. The red dots and the error bars are the mean and $\pm 1$ standard derivation of the rewards over the episodes.}
    \label{fig:hill_game_convergence}
\end{figure}

The convergence of the two experiments is demonstrated in \fref{fig:hill_game_convergence}. 
The figure demonstrates that the mean values of the mixed reward for the game episodes per iteration are maximized and converge by iteration 30. 
In an experimental space this large, it is difficult to perfectly scale the rewards to be exactly maximized at unity. 
Therefore, we empirically tune the scaling so that the rewards are roughly in the range of [0,1]. 
The RL algorithm still discovers a complex path that leads to a maximum reward, and the calibrated parameters shown in Table~\ref{tab:hill_params} are deemed adequate.

\begin{figure}[ht!]
    \centering
    \includegraphics[width=0.85\textwidth]{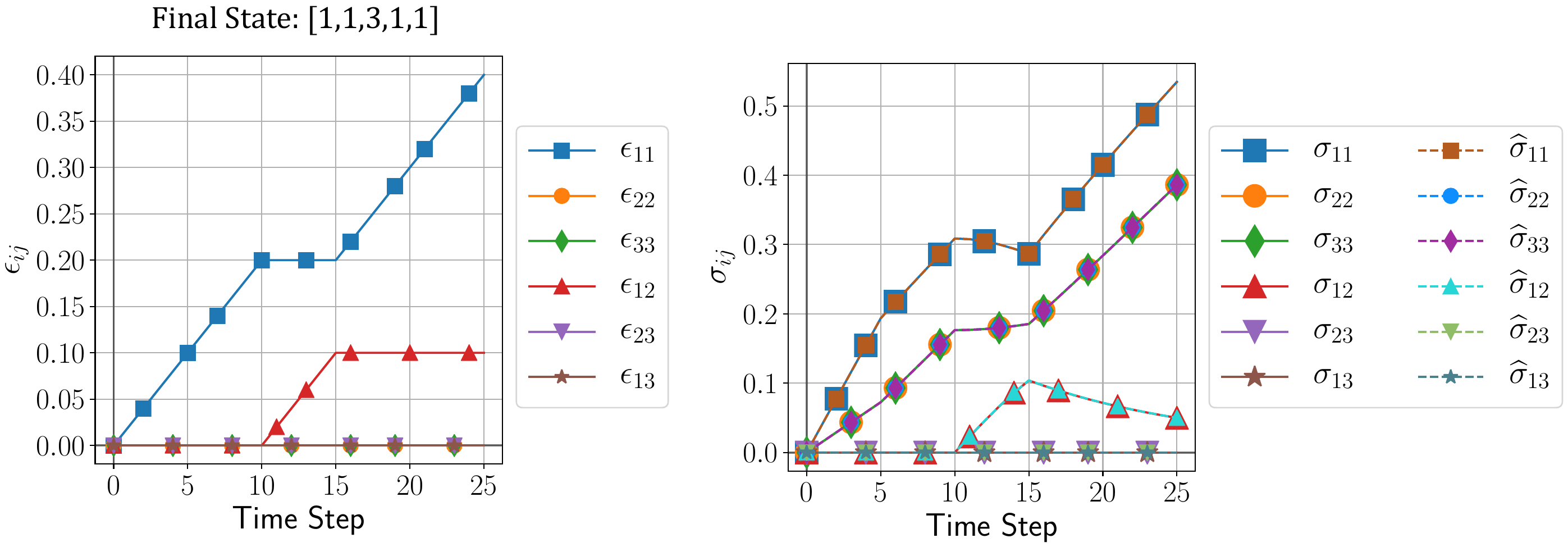}
    \caption{(a) Discovered experiment strain path for the Hill model dataset with $B=0.5$ that corresponds to the tree state $\left[1,1,3,1,1\right]$. (b) The corresponding data stress components $\sigma_{ij}$ and the KF model prediction $\widehat{\sigma}_{ij}$.}
    \label{fig:hill_game_calibration_curve}
\end{figure}

\begin{figure}[ht!]
    \centering
    \includegraphics[width=0.85\textwidth]{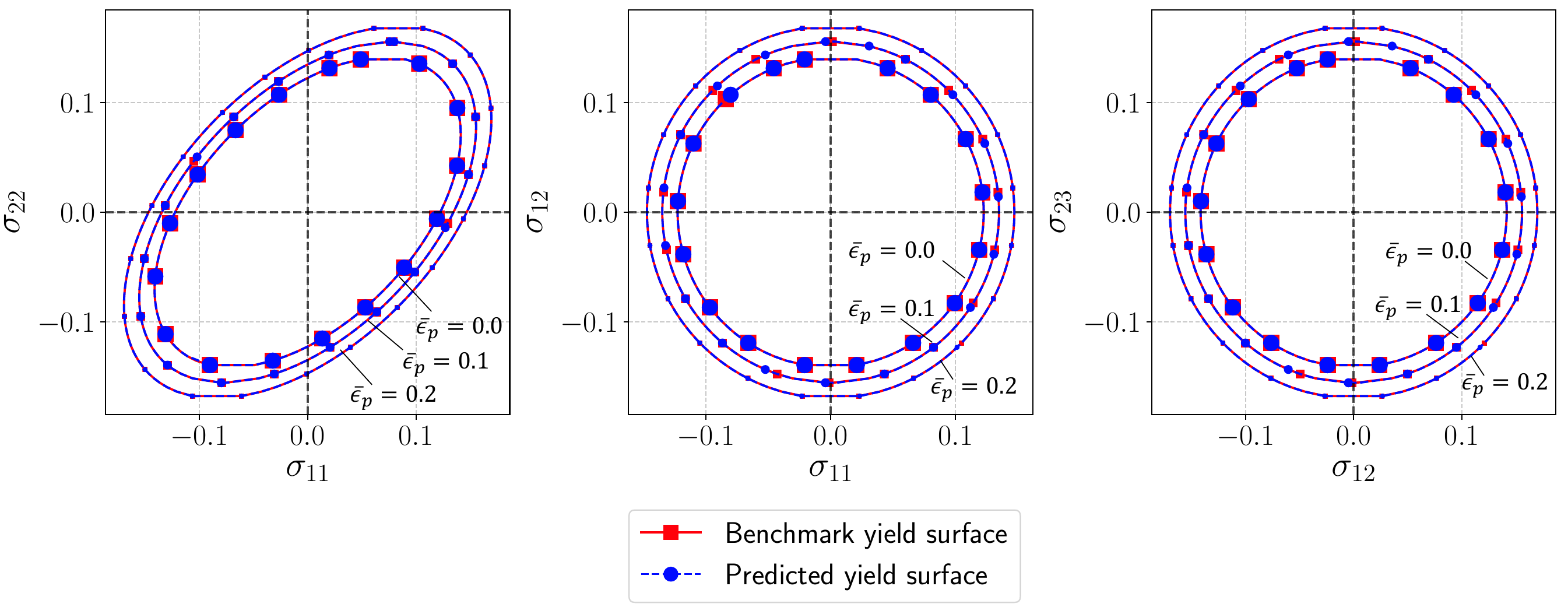}
    \caption{Predicted yield surface for the Hill model $B=0.5$ for the KF-calibrated model for accumulated plastic strain $\overline{\epsilon}^p = 0$, $0.1$ and $0.2$.}
    \label{fig:predicted_hill_surface}
\end{figure}

In \fref{fig:hill_game_calibration_curve}, we present the curve discovered by the RL algorithm for the Hill model dataset with parameter $B=0.5$. 
The figure shows the corresponding full strain tensor components $\epsilon_{ij}$ time history for the predicted converged final state $[1,1,3,1,1]$ at iteration 30. 
It is observed that the KF model calibrates very well to that data with the predicted stress component paths matching the benchmark data. 
The good approximation of the Hill model plasticity parameters is also observed in predicting the anisotropic yield surface and its evolution with hardening as shown in \fref{fig:predicted_hill_surface}.

\begin{figure}[ht!]
    \centering
    \includegraphics[width=0.45\textwidth]{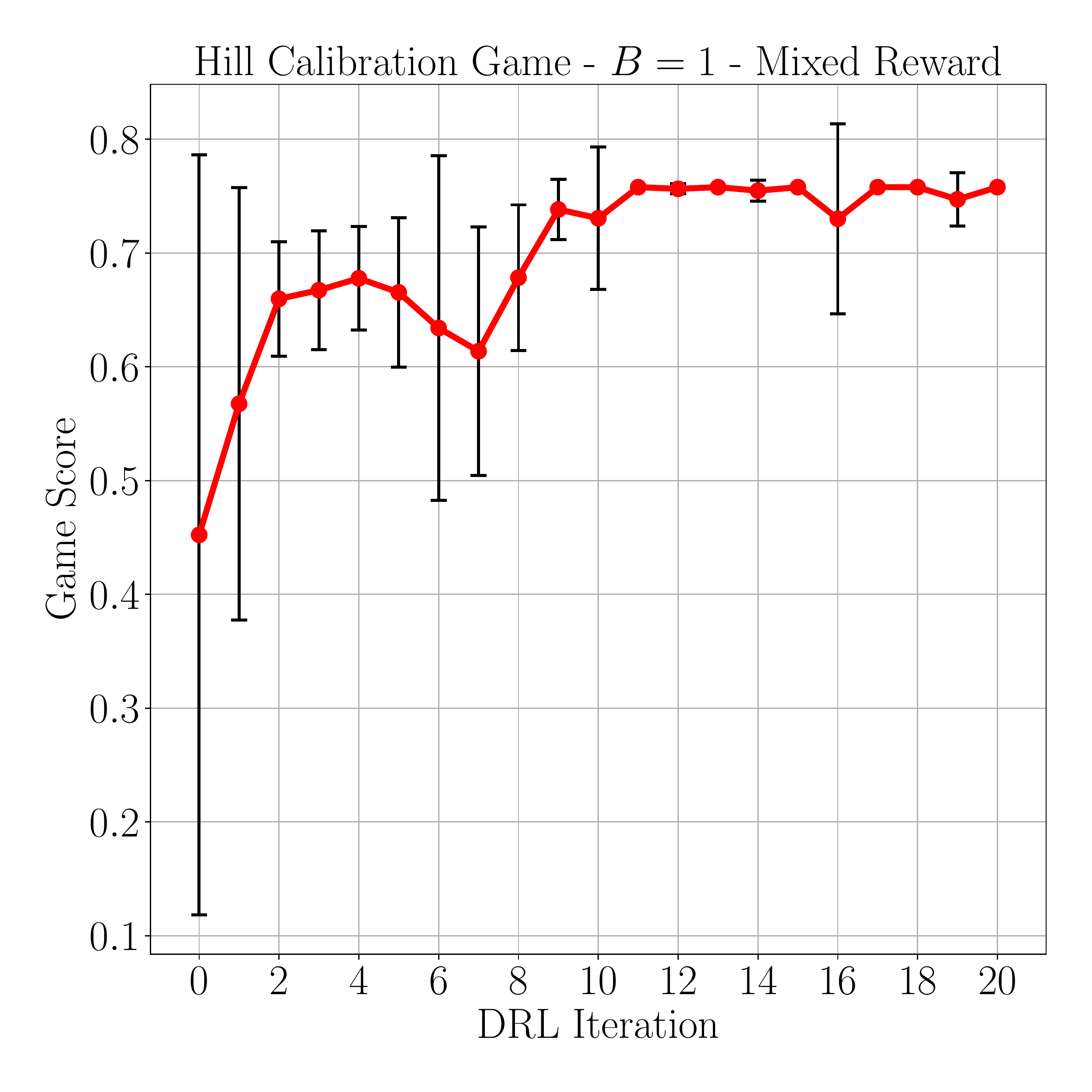}
    \caption{ Game score vs. iteration for the Hill plasticity calibration game model given a dataset for isotropic elasticity and von Mises yield surface ($B=1$). The red dots and the error bars are the mean and $\pm$ standard derivation of the rewards of over the episodes respectively.}
    \label{fig:hill_j2_game_convergence}
\end{figure}

In a third experiment, we test the capacity of the RL algorithm to design an experiment that can reduce the degree of anisotropy of the Hill model to a simpler model. 
Specifically, we allow the KF to calibrate all six anisotropic parameters $\thetab = \{ E, \nu, \nu_\perp, B, Y_0, H \}$ while providing data for an isotropic elastic, von Mises plasticity model. 
The decision tree, neural network setup and RL hyperparameters are identical to the previous two experiments. 
The RL algorithm is run for 20 training iterations and 20 game episodes each.
The mixed reward is utilized in this experiment as well with $w_\text{NSE}= w_\text{KL} = 0.5$. 
The efficiency index is calculated against the strain-stress path shown in \fref{fig:j2_blind_test}.
The convergence of the mixed reward is shown in \fref{fig:hill_j2_game_convergence}, in which the algorithm reaches a maximum score by the 20th iteration. 
The discovered parameters of the Kalman model calibration are shown in Table~\ref{tab:j2_hill_params} where the two calibrated Poisson ratios are equal, indicating isotropy, and the anisotropy parameter $B$ is close to unity, which coincides with the von Mises yield surface.
Note that, due to the usage of the upper confident bound to balance exploitation and exploration, the game score is not necessarily improving monotonically and may dip locally in between an iteration due to the exploration. Nevertheless, the trend of the game score is improving until the game equilibrium is reached. 

\begin{table}[ht!]
\centering
\begin{tabular}{|cc|}
\hline
\multicolumn{2}{|c|}{\textbf{Modified Hill model (B = 1.0 / von Mises)}}                                                                                         \\ \hline
\multicolumn{1}{|c|}{\textbf{Final State}}          & \textbf{Experimental Design}                                                                               \\ \hline
\multicolumn{1}{|c|}{$[1,1,1,12,1]$}                & $+\Delta\epsilon_{11},+\Delta\epsilon_{11},+\Delta\epsilon_{11},-\Delta\epsilon_{13},+\Delta\epsilon_{11}$ \\ \hline
\multicolumn{1}{|c|}{\textbf{Benchmark Parameters}} & \textbf{Calibrated Parameters}                                                                             \\ \hline
\multicolumn{1}{|c|}{$E = 1.5$}                     & $E = 1.5017$                                                                                               \\ \hline
\multicolumn{1}{|c|}{$\nu = 0.3$}                   & $\nu = 0.2996$                                                                                             \\ \hline
\multicolumn{1}{|c|}{$\nu_\perp =0.3 $}             & $\nu_\perp =0.3002 $                                                                                       \\ \hline
\multicolumn{1}{|c|}{$B=1.0$}                       & $B=0.9992$                                                                                                 \\ \hline
\multicolumn{1}{|c|}{$Y_0 = 0.1$}                   & $Y_0 = 0.0999$                                                                                             \\ \hline
\multicolumn{1}{|c|}{$H = 0.1$}                     & $H = 0.1000$                                                                                               \\ \hline
\end{tabular}
    \caption{The RL-discovered final state and corresponding experimental design for the Hill model calibration game along with the benchmark and KF-calibrated parameters calibrated on the linear elasticity/von Mises data.}
    \label{tab:j2_hill_params}
\end{table}

\section{Conclusion} \label{sec:conclusion}

In this paper, we introduced an integrated framework that combines the strengths of Kalman filters (KF) and model-based deep reinforcement learning (DRL) to design experiments for calibrating material models.
The enhanced Kalman filter provides the means to estimate the information gain corresponding to each individual action in an experiment without further sampling.
As shown in the numerical examples, this estimated information gain, quantified by the Kullback-Leibler (KL) divergence, can help us improve the efficiency of the DRL by either replacing the more expensive Nash-Sutcliffe efficiency (NSE) index (which can be expensive to estimate due to the additional cost for k-fold validation), or enabling us to reduce the sampling size for the NSE reward estimation.

In future work, we will enhance the algorithm to improve its performance.
In particular including a variable step size in the action space will allow the method to optimize the decisions near the onset of yield. 
This enhancement may entail reformulating the RL as a continuous time process \citep{doya2000reinforcement}, with explorations conducted by multiple agents and experiment replays to accelerate the training process \citep{vinyals2019grandmaster}.

Ultimately, the goal is to deploy the method as a real-time decision-making process for concurrent experiment data collection and model calibration.
What we have developed is an actor with a policy that is trained to synthetic data that can deployed on an actual experiment and will react to the real time rewards.
We rely on similarity between the model used to generate training data and the actual behavior of the material of interest with the assumption that similar behavior is sufficient to generate a good policy.
Due to the extensive interactions required to generate good policies, (a)  transfer learning between the simulated environmental and physical tests and (b) more effective state representation may both be necessary \citep{de2018integrating}. 
Transfer learning would entail retraining the DNN policy starting from the parameters optimized to the synthetic data with limited data from the actual experiment, while keeping as much of the problem the same, e.g., the action set.
As mentioned, selecting an ideal state representation is an open question, with downsides to adding too many state variables.
Combinatorial studies are brute force means of determining which state representation have advantages over others.
Latent encoding method may prove to have applications as well.
On a related thrust, the appropriate material model is generally not known in advance, so model selection and/or machine learning/data-driven models to augment the traditional models are likely needed. 
As we have shown, the SKF is an effective means to perform model selection among a finite set of models.

Real experiments also present the complications of measurement noise, imprecise or incomplete control of the sample, indirect measurements (e.g. observing forces and displacements instead of stresses and strains), multiple measurement methods (strain gauges as well as full-field surface measurements) and potential advantages of multiple protocols on equivalent samples. 
Each of these challenges will likely engender new developments to the proposed RL algorithm; however, the KF provides means of tackling many of them, such as noise, incomplete control and observation.

\section*{Credit Statement}
The extended Kalman filter was implemented by Dr. Ruben Villareal, whereas the incorporation of the extended Kalman filter and the framework of the deep reinforcement learning was implemented by Dr. Nikolaos Vlassis.
The rest of the authors contributed to developing ideas, writing the manuscript and discussions. 

\section*{Data Availability statement}
The code and data that support the findings of this paper will be posted in an open-source repositories upon the acceptance of the manuscript. 

\section*{Acknowledgements}
The authors are primarily supported by Sandia National Laboratories Computing and Information Sciences Laboratory Directed Research and Development program, with additional support from the Department of Defense SMART scholarship is provided to support Nhon N. Phan. NAT acknowledges support from the Department of Energy early career research program. This support is gratefully acknowledged. WCS would also like to thank Dr. Christine Anderson-Cook from Los Alamos National Laboratory for a fruitful discussion on the design of experiments in 2019.  

Sandia National Laboratories is a multimission laboratory managed and operated by National Technology \& Engineering Solutions of Sandia, LLC, a wholly owned subsidiary of Honeywell International Inc., for the U.S. Department of Energy’s National Nuclear Security Administration under contract DE-NA0003525. This paper describes objective technical results and analysis, which is also archived in the internal Sandia report SAND2022-13022. Any subjective views or opinions that might be expressed in the paper do not necessarily represent the views of the U.S. Department of Energy or the United States Government. This article has been co-authored by an employee of National Technology \& Engineering Solutions of Sandia, LLC under Contract No. DE-NA0003525 with the U.S. Department of Energy (DOE). The employee owns all right, title and interest in and to the article and is solely responsible for its contents. The United States Government retains and the publisher, by accepting the article for publication, acknowledges that the United States Government retains a non-exclusive, paid-up, irrevocable, world-wide license to publish or reproduce the published form of this article or allow others to do so, for United States Government purposes. The DOE will provide public access to these results of federally sponsored research in accordance with the DOE Public Access Plan https://www.energy.gov/downloads/doe-public-access-plan.

\bibliography{main}

\begin{thebibliography}{47}
\providecommand{\natexlab}[1]{#1}
\providecommand{\url}[1]{\texttt{#1}}
\expandafter\ifx\csname urlstyle\endcsname\relax
  \providecommand{\doi}[1]{doi: #1}\else
  \providecommand{\doi}{doi: \begingroup \urlstyle{rm}\Url}\fi

\bibitem[Ames et~al.(2009)Ames, Srivastava, Chester, and Anand]{ames2009thermo}
Nicoli~M Ames, Vikas Srivastava, Shawn~A Chester, and Lallit Anand.
\newblock A thermo-mechanically coupled theory for large deformations of
  amorphous polymers. part ii: Applications.
\newblock \emph{International Journal of Plasticity}, 25\penalty0 (8):\penalty0
  1495--1539, 2009.

\bibitem[Bower(2009)]{bower2009applied}
Allan~F Bower.
\newblock \emph{Applied mechanics of solids}.
\newblock CRC press, 2009.

\bibitem[Catanach(2017)]{catanach2017computational}
Thomas~Anthony Catanach.
\newblock \emph{Computational methods for Bayesian inference in complex
  systems}.
\newblock PhD thesis, California Institute of Technology, 2017.

\bibitem[Chaloner and Verdinelli(1995)]{chaloner1995bayesian}
Kathryn Chaloner and Isabella Verdinelli.
\newblock Bayesian experimental design: A review.
\newblock \emph{Statistical Science}, pages 273--304, 1995.

\bibitem[De~Bruin et~al.(2018)De~Bruin, Kober, Tuyls, and
  Babu{\v{s}}ka]{de2018integrating}
Tim De~Bruin, Jens Kober, Karl Tuyls, and Robert Babu{\v{s}}ka.
\newblock Integrating state representation learning into deep reinforcement
  learning.
\newblock \emph{IEEE Robotics and Automation Letters}, 3\penalty0 (3):\penalty0
  1394--1401, 2018.

\bibitem[Doya(2000)]{doya2000reinforcement}
Kenji Doya.
\newblock Reinforcement learning in continuous time and space.
\newblock \emph{Neural computation}, 12\penalty0 (1):\penalty0 219--245, 2000.

\bibitem[Evensen(2003)]{evensen2003ensemble}
Geir Evensen.
\newblock The ensemble kalman filter: Theoretical formulation and practical
  implementation.
\newblock \emph{Ocean dynamics}, 53\penalty0 (4):\penalty0 343--367, 2003.

\bibitem[Fisher et~al.(1937)]{fisher1937design}
Ronald~Aylmer Fisher et~al.
\newblock \emph{The design of experiments.}
\newblock Oliver \& Boyd, Edinburgh \& London., 1937.

\bibitem[Fuchs et~al.(2021)Fuchs, Heider, Wang, Sun, and
  Kaliske]{fuchs2021dnn2}
Alexander Fuchs, Yousef Heider, Kun Wang, WaiChing Sun, and Michael Kaliske.
\newblock Dnn2: A hyper-parameter reinforcement learning game for self-design
  of neural network based elasto-plastic constitutive descriptions.
\newblock \emph{Computers \& Structures}, 249:\penalty0 106505, 2021.

\bibitem[Gnecco et~al.(2008)Gnecco, Sanguineti,
  et~al.]{gnecco2008approximation}
Giorgio Gnecco, Marcello Sanguineti, et~al.
\newblock Approximation error bounds via rademacher complexity.
\newblock \emph{Applied Mathematical Sciences}, 2:\penalty0 153--176, 2008.

\bibitem[Graesser and Keng(2019)]{graesser2019foundations}
Laura Graesser and Wah~Loon Keng.
\newblock \emph{Foundations of deep reinforcement learning: theory and practice
  in Python}.
\newblock Addison-Wesley Professional, 2019.

\bibitem[Gu et~al.(2016)Gu, Lillicrap, Sutskever, and Levine]{gu2016continuous}
Shixiang Gu, Timothy Lillicrap, Ilya Sutskever, and Sergey Levine.
\newblock Continuous deep q-learning with model-based acceleration.
\newblock In \emph{International conference on machine learning}, pages
  2829--2838. PMLR, 2016.

\bibitem[Heider et~al.(2020)Heider, Wang, and Sun]{heider2020so}
Yousef Heider, Kun Wang, and WaiChing Sun.
\newblock So (3)-invariance of informed-graph-based deep neural network for
  anisotropic elastoplastic materials.
\newblock \emph{Computer Methods in Applied Mechanics and Engineering},
  363:\penalty0 112875, 2020.

\bibitem[Heider et~al.(2021)Heider, Suh, and Sun]{heider2021offline}
Yousef Heider, Hyoung~Suk Suh, and WaiChing Sun.
\newblock An offline multi-scale unsaturated poromechanics model enabled by
  self-designed/self-improved neural networks.
\newblock \emph{International Journal for Numerical and Analytical Methods in
  Geomechanics}, 45\penalty0 (9):\penalty0 1212--1237, 2021.

\bibitem[Jazwinski(2007)]{jazwinski2007stochastic}
Andrew~H Jazwinski.
\newblock \emph{Stochastic processes and filtering theory}.
\newblock Courier Corporation, 2007.

\bibitem[Jones et~al.(2022)Jones, Frankel, and Johnson]{jones2022neural}
Reese~E Jones, Ari~L Frankel, and KL~Johnson.
\newblock A neural ordinary differential equation framework for modeling
  inelastic stress response via internal state variables.
\newblock \emph{Journal of Machine Learning for Modeling and Computing},
  3\penalty0 (3), 2022.

\bibitem[Julier and Uhlmann(1997)]{julier1997new}
Simon~J Julier and Jeffrey~K Uhlmann.
\newblock New extension of the kalman filter to nonlinear systems.
\newblock In \emph{Signal processing, sensor fusion, and target recognition
  VI}, volume 3068, pages 182--193. Spie, 1997.

\bibitem[Kaelbling et~al.(1996)Kaelbling, Littman, and
  Moore]{kaelbling1996reinforcement}
Leslie~Pack Kaelbling, Michael~L Littman, and Andrew~W Moore.
\newblock Reinforcement learning: A survey.
\newblock \emph{Journal of artificial intelligence research}, 4:\penalty0
  237--285, 1996.

\bibitem[Kalman(1960)]{kalman1960new}
R.~E. Kalman.
\newblock {A New Approach to Linear Filtering and Prediction Problems}.
\newblock \emph{Journal of Basic Engineering}, 82\penalty0 (1):\penalty0
  35--45, 03 1960.

\bibitem[Kingma and Ba(2014)]{kingma2014adam}
Diederik~P Kingma and Jimmy Ba.
\newblock Adam: A method for stochastic optimization.
\newblock \emph{arXiv preprint arXiv:1412.6980}, 2014.

\bibitem[Landajuela et~al.(2021)Landajuela, Petersen, Kim, Santiago, Glatt,
  Mundhenk, Pettit, and Faissol]{landajuela2021discovering}
Mikel Landajuela, Brenden~K Petersen, Sookyung Kim, Claudio~P Santiago, Ruben
  Glatt, Nathan Mundhenk, Jacob~F Pettit, and Daniel Faissol.
\newblock Discovering symbolic policies with deep reinforcement learning.
\newblock In \emph{International Conference on Machine Learning}, pages
  5979--5989. PMLR, 2021.

\bibitem[LaViola(2003)]{laviola2003comparison}
Joseph~J LaViola.
\newblock A comparison of unscented and extended kalman filtering for
  estimating quaternion motion.
\newblock In \emph{Proceedings of the 2003 American Control Conference, 2003.},
  volume~3, pages 2435--2440. IEEE, 2003.

\bibitem[Lee and Ricker(1994)]{lee1994extended}
Jay~H Lee and N~Lawrence Ricker.
\newblock Extended kalman filter based nonlinear model predictive control.
\newblock \emph{Industrial \& Engineering Chemistry Research}, 33\penalty0
  (6):\penalty0 1530--1541, 1994.

\bibitem[Li(2017)]{li2017deep}
Yuxi Li.
\newblock Deep reinforcement learning: An overview.
\newblock \emph{arXiv preprint arXiv:1701.07274}, 2017.

\bibitem[Lubliner(2008)]{lubliner2008plasticity}
Jacob Lubliner.
\newblock \emph{Plasticity theory}.
\newblock Courier Corporation, 2008.

\bibitem[Ma and Sun(2020)]{ma2020computational}
Ran Ma and WaiChing Sun.
\newblock Computational thermomechanics for crystalline rock. part ii:
  Chemo-damage-plasticity and healing in strongly anisotropic polycrystals.
\newblock \emph{Computer Methods in Applied Mechanics and Engineering},
  369:\penalty0 113184, 2020.

\bibitem[McCuen et~al.(2006)McCuen, Knight, and Cutter]{mccuen2006evaluation}
Richard~H McCuen, Zachary Knight, and A~Gillian Cutter.
\newblock Evaluation of the nash--sutcliffe efficiency index.
\newblock \emph{Journal of hydrologic engineering}, 11\penalty0 (6):\penalty0
  597--602, 2006.

\bibitem[Moskovitz et~al.(2021)Moskovitz, Parker-Holder, Pacchiano, Arbel, and
  Jordan]{moskovitz2021tactical}
Ted Moskovitz, Jack Parker-Holder, Aldo Pacchiano, Michael Arbel, and Michael
  Jordan.
\newblock Tactical optimism and pessimism for deep reinforcement learning.
\newblock \emph{Advances in Neural Information Processing Systems},
  34:\penalty0 12849--12863, 2021.

\bibitem[Murphy(1998)]{Murphy98switchingkalman}
Kevin~P. Murphy.
\newblock Switching kalman filters.
\newblock Technical report, DEC/Compaq Cambridge Research Labs, 1998.

\bibitem[Niv(2009)]{niv2009reinforcement}
Yael Niv.
\newblock Reinforcement learning in the brain.
\newblock \emph{Journal of Mathematical Psychology}, 53\penalty0 (3):\penalty0
  139--154, 2009.

\bibitem[Pukelsheim(2006)]{pukelsheim2006optimal}
Friedrich Pukelsheim.
\newblock \emph{Optimal design of experiments}.
\newblock SIAM, 2006.

\bibitem[Reda et~al.(2020)Reda, Tao, and van~de Panne]{reda2020learning}
Daniele Reda, Tianxin Tao, and Michiel van~de Panne.
\newblock Learning to locomote: Understanding how environment design matters
  for deep reinforcement learning.
\newblock In \emph{Motion, Interaction and Games}, pages 1--10. ACM, 2020.

\bibitem[Scherzinger(2017)]{scherzinger2017return}
William~M Scherzinger.
\newblock A return mapping algorithm for isotropic and anisotropic plasticity
  models using a line search method.
\newblock \emph{Computer Methods in Applied Mechanics and Engineering},
  317:\penalty0 526--553, 2017.

\bibitem[Schrittwieser et~al.(2021)Schrittwieser, Hubert, Mandhane, Barekatain,
  Antonoglou, and Silver]{schrittwieser2021online}
Julian Schrittwieser, Thomas Hubert, Amol Mandhane, Mohammadamin Barekatain,
  Ioannis Antonoglou, and David Silver.
\newblock Online and offline reinforcement learning by planning with a learned
  model.
\newblock \emph{Advances in Neural Information Processing Systems},
  34:\penalty0 27580--27591, 2021.

\bibitem[Silver et~al.(2017{\natexlab{a}})Silver, Hubert, Schrittwieser,
  Antonoglou, Lai, Guez, Lanctot, Sifre, Kumaran, Graepel,
  et~al.]{silver2017mastering}
David Silver, Thomas Hubert, Julian Schrittwieser, Ioannis Antonoglou, Matthew
  Lai, Arthur Guez, Marc Lanctot, Laurent Sifre, Dharshan Kumaran, Thore
  Graepel, et~al.
\newblock Mastering chess and shogi by self-play with a general reinforcement
  learning algorithm.
\newblock \emph{arXiv preprint arXiv:1712.01815}, 2017{\natexlab{a}}.

\bibitem[Silver et~al.(2017{\natexlab{b}})Silver, Hubert, Schrittwieser,
  Antonoglou, Lai, Guez, Lanctot, Sifre, Kumaran, Graepel,
  et~al.]{silver2017masteringchess}
David Silver, Thomas Hubert, Julian Schrittwieser, Ioannis Antonoglou, Matthew
  Lai, Arthur Guez, Marc Lanctot, Laurent Sifre, Dharshan Kumaran, Thore
  Graepel, et~al.
\newblock Mastering chess and shogi by self-play with a general reinforcement
  learning algorithm.
\newblock \emph{arXiv preprint arXiv:1712.01815}, 2017{\natexlab{b}}.

\bibitem[Silver et~al.(2017{\natexlab{c}})Silver, Schrittwieser, Simonyan,
  Antonoglou, Huang, Guez, Hubert, Baker, Lai, Bolton,
  et~al.]{silver2017masteringb}
David Silver, Julian Schrittwieser, Karen Simonyan, Ioannis Antonoglou, Aja
  Huang, Arthur Guez, Thomas Hubert, Lucas Baker, Matthew Lai, Adrian Bolton,
  et~al.
\newblock Mastering the game of go without human knowledge.
\newblock \emph{Nature}, 550\penalty0 (7676):\penalty0 354, 2017{\natexlab{c}}.

\bibitem[Simo and Hughes(2006)]{simo2006computational}
Juan~C Simo and Thomas~JR Hughes.
\newblock \emph{Computational inelasticity}, volume~7.
\newblock Springer Science \& Business Media, 2006.

\bibitem[Sun and Sun(2015)]{sun2015model}
Ne-Zheng Sun and Alexander Sun.
\newblock \emph{Model calibration and parameter estimation: for environmental
  and water resource systems}.
\newblock Springer, 2015.

\bibitem[Sutton and Barto(2018)]{sutton2018reinforcement}
Richard~S Sutton and Andrew~G Barto.
\newblock \emph{Reinforcement learning: An introduction}.
\newblock MIT press, 2018.

\bibitem[Vinyals et~al.(2019)Vinyals, Babuschkin, Czarnecki, Mathieu, Dudzik,
  Chung, Choi, Powell, Ewalds, Georgiev, et~al.]{vinyals2019grandmaster}
Oriol Vinyals, Igor Babuschkin, Wojciech~M Czarnecki, Micha{\"e}l Mathieu,
  Andrew Dudzik, Junyoung Chung, David~H Choi, Richard Powell, Timo Ewalds,
  Petko Georgiev, et~al.
\newblock Grandmaster level in starcraft ii using multi-agent reinforcement
  learning.
\newblock \emph{Nature}, 575\penalty0 (7782):\penalty0 350--354, 2019.

\bibitem[Vlassis and Sun(2022)]{vlassis2022component}
Nikolaos~N Vlassis and WaiChing Sun.
\newblock Component-based machine learning paradigm for discovering
  rate-dependent and pressure-sensitive level-set plasticity models.
\newblock \emph{Journal of Applied Mechanics}, 89\penalty0 (2), 2022.

\bibitem[Wang and Sun(2019)]{wang2019meta}
Kun Wang and WaiChing Sun.
\newblock Meta-modeling game for deriving theory-consistent,
  microstructure-based traction--separation laws via deep reinforcement
  learning.
\newblock \emph{Computer Methods in Applied Mechanics and Engineering},
  346:\penalty0 216--241, 2019.

\bibitem[Wang et~al.(2019)Wang, Sun, and Du]{wang2019cooperative}
Kun Wang, WaiChing Sun, and Qiang Du.
\newblock A cooperative game for automated learning of elasto-plasticity
  knowledge graphs and models with ai-guided experimentation.
\newblock \emph{Computational Mechanics}, pages 1--33, 2019.

\bibitem[Wang et~al.(2021)Wang, Sun, and Du]{wang2021non}
Kun Wang, WaiChing Sun, and Qiang Du.
\newblock A non-cooperative meta-modeling game for automated third-party
  calibrating, validating and falsifying constitutive laws with parallelized
  adversarial attacks.
\newblock \emph{Computer Methods in Applied Mechanics and Engineering},
  373:\penalty0 113514, 2021.

\bibitem[West et~al.(2001)]{west2001introduction}
Douglas~Brent West et~al.
\newblock \emph{Introduction to graph theory}, volume~2.
\newblock Prentice hall Upper Saddle River, 2001.

\bibitem[Williams(1992)]{williams1992training}
Ronald~J Williams.
\newblock Training recurrent networks using the extended kalman filter.
\newblock In \emph{[Proceedings 1992] IJCNN International Joint Conference on
  Neural Networks}, volume~4, pages 241--246. IEEE, 1992.

\end{thebibliography}

\section*{Appendix}
\appendix
\section{EKF for state and parameter estimation of DAEs} \label{app:DAE_KF}
\setcounter{equation}{0}
\renewcommand{\theequation}{\thesection.\arabic{equation}}

In this appendix we will derive and discuss the extended Kalman filter (EKF) in the context of semi-explicit index-1 differential algebraic equations (DAEs) for joint state and parameter estimation. 
The plasticity model described in \sref{sec:exemplar} is an example of a DAE system with an algebraic stress rule \eref{eq:stress} and an ordinary differential equation (ODE) prescribing the flow of the hidden state variables \eref{eq:flow} subjected to the algebraic yield constraint \eref{eq:yield}.
In general, these DAEs have the form
\begin{align}
\dot{\zb} &= \fb\left(\zb, \sigmab, \thetab, \xb, t\right )\\
\mathbf{0} &= \gb \left(\zb, \sigmab, \thetab, \xb, t\right) \ .
\end{align}
Here $\zb$ are unobserved dynamic states, $\sigmab$ are unobserved algebraic states, $\thetab$ are model parameters, $\xb$ are known inputs and $t$ is time. 
Since we are dealing with index-1 DAEs, we can assume that $g\left(\zb, \sigmab, \xb, t\right) = 0$ is solvable for $\sigmab$. 
In addition to the DAE process model, we assume that there is an observation model for measurement $\ds$, given by
\begin{equation}
\ds = \mb\left(\zb, \sigmab, \thetab, \xb, t\right) + \varepsilonb \ ,
\end{equation}
where $\varepsilonb$ is noise which we assume follows a Gaussian distribution.

Considering that these dynamics are specified using a continuous DAE system, we need to discretize them in time. 
Further we will assume that the models are not time dependent for simplicity but extending this method to the time-dependent case is straight forward. 
While there are no closed-form solutions to the DAEs explicitly, for convenience we can define the solution as the function $\mathbcal{f}$ for the dynamic state update and $\mathbcal{m}$ for the explicit measurement function when a closed-form solution does not exist (e.g., if it depends on $\sigmab$). 
As a result,
\begin{align}
\zb_k &= \mathbcal{f} \left(\zb_{k-1}, \thetab, \xb_k\right) + \etab_k\\
\ds_k &= \mathbcal{m} \left(\zb_k, \thetab, \xb_k \right) + \varepsilonb_k \ .
\end{align}
for time $t_k$.
Here the addition of a Gaussian process noise term $\etab$ reflects modeling errors due to the discretization in addition to any intrinsic noise. 
It is important to note that there are many choices of $\mathbcal{f}$ depending on the discretization and numerical integration scheme used to solve the DAEs. 
This explicit construction, though not implementable in a closed form, defines the functions that we need to linearize in order to construct the EKF. 
We can also augment the state to include fictitious dynamics of the model parameters to aid in model parameter identification:
\begin{equation}
\thetab_k = \thetab_{k-1} + \deltab_k \ ,
\end{equation}
where $\deltab$ is again additive Gaussian noise. 
For exact parameter estimation, $\deltab_k=0$ because the parameters are fixed; however, in some cases for stability adding small amounts of noise can reduce bias in the estimated parameters at the cost of increased variance and slower convergence of the estimation.

Under this construction, the EKF prediction step has the form
\begin{align}
\zb_{k\mid k-1} &= \mathbcal{f} \left (\zb_{k-1 \mid k-1}, \thetab_{k-1 \mid k-1}, \xb_k\right)\\
\thetab_{k\mid k-1} &= \thetab_{k-1 \mid k-1}\\
\ds_{k\mid k-1} &= \mathbcal{m} \left (\zb_{k\mid k-1} , \thetab_{k\mid k-1}, \xb_k \right) \ ,
\end{align}
while the uncertainty propagation on the prediction has the form
\begin{align}
\Sigmab_{k \mid k-1} &= \Fs_k \Sigmab_{k-1 \mid k-1} \Fs_k^T + \Qs_k\\
\Ss_{k \mid k-1} &= \As_k \Sigmab_{k \mid k-1}\As_k^T + \Rs \ ,
\end{align}
where $\Sigmab$ is the covariance for the joint state $\left[\zb, \thetab \right]^T$, 
$\Qs$ is the process and parameter additive uncertainty assumed to be independent
and has covariances $\Qs_\etab$ and $\Qs_\deltab$, respectively, and $\Rs$ is the measurement noise covariance. 
We also must construct the linearizations of the dynamics $\Fs$ and the measurement $\As$:
\begin{align}
\Qs_{k}&=\begin{bmatrix} \Qs_\etab & \mathbf{0} \\
\mathbf{0}  &  \Qs_\deltab
\end{bmatrix}\\
\Fs_{k}&=\begin{bmatrix} \partialb_\zb \mathbcal{f}  &  \partialb_\thetab  \mathbcal{f} \\
\Ib & \mathbf{0}
\end{bmatrix}_{\zb_{k\mid k-1}, \thetab_{k\mid k-1}, \xb_k}\\
\As_{k}&=\begin{bmatrix} \partialb_\zb \mathbcal{m}  &  \partialb_\thetab \mathbcal{m}
\end{bmatrix}_{\zb_{k-1\mid k-1}, \thetab_{k-1\mid k-1}, \xb_k} \ .
\end{align}

Once all these variables have been defined and a measurement $\ds_k$ has been made, the EKF update is straight forward. 
The EKF update is given by:
\begin{align}
\rs_k &= \ds_k - \ds_{k\mid k-1}\\
\Ks_k &= \Sigmab_{k \mid k-1} \As_{k}^T \Ss_{k \mid k-1}^{-1}\\
\zb_{k\mid k} &= \zb_{k\mid k-1} + \Ks_k \rs_k\\
\Sigmab_{k \mid k} &= \left ( \Ib - \Ks_k \As_k \right )  \Sigmab_{k \mid k-1} \ .
\end{align}
Therefore, in order to apply the EKF to the DAEs, we must compute the derivatives: $\partialb_\zb \mathbcal{f}$, $\partialb_\thetab \mathbcal{f}$, $\partial_\zb \mathbcal{m}$ and $\partial_\thetab \mathbcal{m}$.

We will compute these derivatives for the case where the dynamics are implicitly solved using the backward Euler method (as is common for plasticity updates). 
Thus, returning to the discrete time DAEs, the model becomes
\begin{align}
\zb_{k} &= \zb_{k-1} + \Delta_t  \fb\left(\zb_k, \sigmab_k, \thetab_k, \xb_k \right)\\
\thetab_k &= \thetab_{k-1}\\
\mathbf{0} &= \gb\left(\zb_k, \sigmab_k, \thetab_k, \xb_k \right)\\
\ds_{k} &= \mb\left (\zb_k , \sigmab_k, \thetab_k, \zb_k \right ) \ .
\end{align}
Before we begin with the derivation of the derivatives needed for the EKF, we will derive the following useful derivatives: $\partialb_{\zb_{k-1}} \thetab_k$, 
$\partialb_{\thetab_{k-1}} \zb_{k-1}$, 
$\partialb_{\zb_{k-1}} \sigmab_k$, 
$\partialb_{\zb_k} \sigmab_k$ and 
$\partialb_{\thetab_{k-1}} \sigmab_k$. 
By inspection, we see that $\partialb_{\zb_{k-1}} \thetab_k = \mathbf{0}$ and  $\partialb_{\thetab_{k-1}} \zb_{k-1} = \mathbf{0}$. 
This realization might seem counter-intuitive because obviously there is a relationship between $\zb_{k-1}$ and $\thetab_{k-1}$; however, that relationship is already being accounted for via $\Sigmab$. 
To find $\partialb_{\zb_{k-1}} \sigmab_k$ we must implicitly compute the derivative of the algebraic constraint
\begin{alignat}{2}
&&\partialb_{\zb_{k-1}} \mathbf{0} \equiv \mathbf{0} &= \partialb_{\zb_{k-1}} \gb \left(\zb_k, \sigmab_k, \thetab_k, \xb_k \right)\\
& &&=
\partialb_{\zb_{k}} \gb \,
\partialb_{\zb_{k-1}} \zb_k  + 
\partialb_{\thetab_{k}} \gb \,
\partialb_{\zb_{k-1}} \thetab_k + 
\partialb_{\sigmab_{k}} \gb \,
\partialb_{\zb_{k-1}} \sigmab_k \nonumber\\
& &&=
\partialb_{\zb_{k}} \gb \,
\partialb_{\zb_{k-1}} \zb_k + 
\partialb_{\sigmab_{k}} \gb \,
\partialb_{\zb_{k-1}} \sigmab_k \nonumber\\
& \implies &
\partialb_{\zb_{k-1}} \sigmab_k &= - \left(
\partialb_{\sigmab_{k}} \gb\right)^{-1} \,
\partialb_{\zb_{k}} \gb \,
\partialb_{\zb_{k-1}} \zb_k \ .
\end{alignat}
Here we used the fact that $\partialb_{\zb_{k-1}} \thetab_k = \mathbf{0}$. 
Also, we know that $\partialb_{\sigmab_{k}} g$ is invertible because it is an index-1 DAE. 
Following the same argument, we also find that $\partialb_{\zb_k} \sigmab_k = -\left(\partialb_{\sigmab_{k}} \gb \right)^{-1} \partialb_{\zb_{k}} \gb$.

Similarly, to find $\partialb_{\thetab_{k-1}} \sigmab_k$ we must implicitly compute the derivative of the algebraic constraint
\begin{alignat}{2}
&& \partialb_{\thetab_{k-1}} \mathbf{0} \equiv \mathbf{0} &= 
\partialb_{\thetab_{k-1}} \mathbf{g} \left(\zb_k, \sigmab_k, \thetab_k, \xb_k \right) \\
& &&= \partialb_{\zb_{k}} \mathbf{g} \,
\partialb_{\thetab_{k-1}} \zb_k  + 
\partialb_{\thetab_{k}} \mathbf{g} \, 
\partialb_{\thetab_{k-1}} \thetab_k + 
\partialb_{\sigmab_{k}} \mathbf{g} \, 
\partialb_{\thetab_{k-1}} \sigmab_k \nonumber\\
& \implies &
\partialb_{\thetab_{k-1}} \sigmab_k &= - \left(
\partialb_{\sigmab_{k}} \gb  \right)^{-1} \left(
\partialb_{ \zb_{k}} \gb \,
\partialb_{ \thetab_{k-1}} \zb_k + 
\partialb_{ \thetab_k}  \gb \right)\ .
\end{alignat}

Now, using the previous definitions, we can use the same implicit strategy to solve for $\partialb_{\zb} \mathbcal{f}$. 
We can derive it as
\begin{alignat}{2}
&& \partialb_{\zb} \mathbcal{f} = 
\partialb_{\zb_{k-1}} \zb_k &= 
\partialb_{\zb_{k-1}} \left(\zb_{k-1} + \Delta_t \mathbcal{f} \left (\zb_k, \sigmab_k, \thetab_k, \xb_k \right) \right)\\
& &&= \Ib + \Delta_t \left( 
\partialb_{\zb_k} \mathbcal{f} \,
\partialb_{\zb_{k-1}} \zb_k + 
\partialb_{\thetab_k} \mathbcal{f} \,
\partialb_{\zb_{k-1}} \thetab_k + 
\partialb_{\sigmab_k} \mathbcal{f} \,
\partialb_{\zb_{k-1}} \sigmab_k \right) \nonumber\\
& &&= \Ib + \Delta_t \left( 
\partialb_{\zb_k} \mathbcal{f} \,
\partialb_{\zb_{k-1}} \zb_k - 
\partialb_{\sigmab_k} \mathbcal{f} \left(
\partialb_{\sigmab_{k}} \gb \right)^{-1} 
\partialb_{\zb_{k}} \gb \,
\partialb_{\zb_{k-1}} \zb_k \right ) \nonumber\\
&\implies &
\partialb_{\zb} \mathbcal{f} &= \left(\Ib - \Delta_t 
\partialb_{\zb_k} \mathbcal{f} + \Delta_t 
\partialb_{\sigmab_k} \mathbcal{f} \left(
\partialb_{\sigmab_{k}} \gb \right)^{-1} 
\partialb_{\zb_{k}} \gb \right )^{-1} \ .
\end{alignat}
Using a similar strategy, we can compute $\partialb_{\thetab} \mathbcal{f}$:
\begin{alignat}{2}
&&
\partialb_{\theta} \mathbcal{f} = 
\partialb_{\thetab_{k-1}} \zb_k &= 
\partialb_{\thetab_{k-1}} \left(\zb_{k-1} + \Delta_t \mathbcal{f} \left(\zb_k, \sigmab_k, \thetab_k, \xb_k \right) \right )\\
& &&= 
\partialb_{\thetab_{k-1}} \zb_{k-1} + \Delta_t \left( 
\partialb_{\zb_k} \mathbcal{f} \, 
\partialb_{\thetab_{k-1}} \zb_k + 
\partialb_{\thetab_k} \mathbcal{f} \,
\partialb_{\thetab_{k-1}} \thetab_k + 
\partialb_{\sigmab_k} \mathbcal{f} \,
\partialb_{\thetab_{k-1}} \sigmab_k \right) \nonumber \\
& &&= \Delta_t \left ( 
\partialb_{\zb_k} \mathbcal{f} \,
\partialb_{\thetab_{k-1}} \zb_k + 
\partialb_{\thetab_k} \mathbcal{f} - 
\partialb_{\sigmab_k} \mathbcal{f} \left(
\partialb_{\sigmab_{k}} \gb\right)^{-1} \left(
\partialb_{\zb_{k}} \gb \, 
\partialb_{\thetab_{k-1}} \zb_k + 
\partialb_{\thetab_k} \gb \right) \right) \nonumber\\
&\implies &
\partialb_{\thetab} \mathbcal{f} &= \left(\Ib - \Delta_t 
\partialb_{\zb_k} \mathbcal{f} +\Delta_t 
\partialb_{\sigmab_k} \mathbcal{f} \left(
\partialb_{\sigmab_k} \gb \right )^{-1} 
\partialb_{\zb_k} \gb \right )^{-1} \left( \Delta_t 
\partialb_{\thetab_k} \mathbcal{f} - \Delta_t 
\partialb_{\sigmab_k} \mathbcal{f} \left (
\partialb_{\sigmab_k} \gb \right )^{-1} 
\partialb_{\thetab_k} \gb \right )\ .
\end{alignat}

Computing the derivatives for the measurement function is more straight forward because we are now no longer using the implicit integrator, and all the key derivatives have already been defined. 
We find that
\begin{align}
\partialb_{\zb_k} \mathbcal{m} &= 
\partialb_{\zb_k} \mathbcal{m} + 
\partialb_{\sigmab_k} \mathbcal{m} \,
\partialb_{\zb_k} \sigmab_k \\
\partialb_{\thetab_k} \mathbcal{m} &= 
\partialb_{\zb_k} \mathbcal{m} \,
\partialb_{\theta} \mathbcal{f} + 
\partialb_{\thetab_k} \mathbcal{m}+ 
\partialb_{\sigmab_k} \mathbcal{m} \,
\partialb_{\thetab_k} \sigmab_k \ .
\end{align}

For the special case when $m \left(\zb_k , \sigmab_k, \thetab_k, \xb_k \right) = \sigmab_k$, as in our models, we can significantly simplify these equations as
\begin{align}
\partialb_{\zb_k} \mathbcal{m} &= 
\partialb_{\zb_k} \sigmab_k = - \left (
\partialb_{\sigmab_{k}} \gb \right)^{-1} 
\partialb_{\zb_{k}} \gb \\
\partialb_{\thetab_k} \mathbcal{m} &= 
\partialb_{\thetab_k} \sigmab_k = - \left(
\partialb_{\sigmab_{k}} \gb \right)^{-1} \left(
\partialb_{\zb_{k}} \gb \,
\partialb_{\theta} \mathbcal{f} + 
\partialb_{\thetab_k} \gb \right) \ .
\end{align}

\end{document}